\documentclass[10pt]{article} 
\usepackage[accepted]{tmlr}

\usepackage{amsmath,amsfonts,bm}

\def\eqref#1{equation~\ref{#1}}

\def\1{\bm{1}}

\DeclareMathAlphabet{\mathsfit}{\encodingdefault}{\sfdefault}{m}{sl}
\SetMathAlphabet{\mathsfit}{bold}{\encodingdefault}{\sfdefault}{bx}{n}

\usepackage[T1]{fontenc}
\usepackage[utf8]{inputenc}
\usepackage{url}
\usepackage{hyperref}

\usepackage{amsmath}
\usepackage{amssymb}

\usepackage{graphicx}
\usepackage{subcaption}
\usepackage{booktabs}
\usepackage{tabularx}
\usepackage{array}
\usepackage{caption}
\usepackage{makecell}
\usepackage{multirow}
\usepackage{colortbl}
\usepackage{fontawesome5}  

\usepackage{xcolor}
\usepackage[most]{tcolorbox}
\usepackage{pgfplots}
\pgfplotsset{compat=1.17}

\usepackage{pifont}
\usepackage{xspace}

\usepackage[capitalize,noabbrev]{cleveref}

\definecolor{royalblue(web)}{rgb}{0.25, 0.41, 0.88}
\definecolor{blue-violet}{rgb}{0.54, 0.17, 0.89}
\definecolor{brightmaroon}{rgb}{0.76, 0.13, 0.28}
\definecolor{darkmagenta}{rgb}{0.55, 0.0, 0.55}
\definecolor{bleudefrance}{rgb}{0.19, 0.55, 0.91}
\definecolor{palatinateblue}{rgb}{0.15, 0.23, 0.89}
\definecolor{whitesmoke}{rgb}{0.96, 0.96, 0.96}
\definecolor{thulianpink}{rgb}{0.87, 0.44, 0.63}
\definecolor{amber(sae/ece)}{rgb}{1.0, 0.49, 0.0}
\definecolor{darkblue}{rgb}{0.0, 0.0, 0.55}
\definecolor{alizarin}{rgb}{0.82, 0.1, 0.26}
\definecolor{asparagus}{rgb}{0.53, 0.66, 0.42}
\definecolor{darkspringgreen}{rgb}{0.09, 0.45, 0.27}
\definecolor{columbiablue}{rgb}{0.61, 0.87, 1.0}
\definecolor{wildblueyonder}{rgb}{0.64, 0.68, 0.82}
\definecolor{trolleygrey}{rgb}{0.5, 0.5, 0.5}
\definecolor{paleaqua}{rgb}{0.74, 0.83, 0.9}
\definecolor{bubblegum}{rgb}{0.99, 0.76, 0.8}
\definecolor{coralred}{rgb}{1.0, 0.25, 0.25}
\definecolor{green(ryb)}{rgb}{0.4, 0.69, 0.2}
\definecolor{flame}{rgb}{0.89, 0.35, 0.13}
\definecolor{bittersweet}{rgb}{1.0, 0.44, 0.37}
\definecolor{darksalmon}{rgb}{0.91, 0.59, 0.48}
\definecolor{emerald}{rgb}{0.31, 0.78, 0.47}
\definecolor{green(pigment)}{rgb}{0.0, 0.65, 0.31}
\definecolor{codegreen}{rgb}{0,0.6,0}
\definecolor{codegray}{rgb}{0.5,0.5,0.5}
\definecolor{codepurple}{rgb}{0.58,0,0.82}
\definecolor{backcolour}{rgb}{0.96,0.96,0.94}
\definecolor{forestgreen}{RGB}{34, 139, 34}
\definecolor{custompurple}{HTML}{800080}

\tcbuselibrary{listingsutf8}
\newtcblisting{promptbox}{
  colback=gray!5!white,
  colframe=gray!75!black,
  listing only,
  listing options={
    basicstyle=\ttfamily\footnotesize,
    breaklines=true,
    columns=fullflexible,
    keepspaces=true,
    showstringspaces=false
  },
  left=5pt, right=5pt, top=5pt, bottom=5pt,
  sharp corners,
  boxrule=0.5pt
}

\newcommand{\xhdr}[1]{{\noindent\bfseries #1}.}
\newcommand{\bench}{\texttt{LLMFusionBench}\xspace}
\newcommand{\method}{\texttt{FusionFactory}\xspace}

\title{FusionFactory: Fusing LLM Capabilities with Multi-LLM Log Data}

\author{%
  \name Tao Feng\textsuperscript{*} \email taofeng2@illinois.edu \\
  \addr University of Illinois Urbana-Champaign
  \AND
  \name Haozhen Zhang\textsuperscript{*} \email wazhz14@gmail.com \\
  \addr Nanyang Technological University
  \AND
  \name Zijie Lei\textsuperscript{*} \email zijielei303@meta.com \\
  \addr Meta Monetization AI
  \AND
  \name Pengrui Han \email phan12@illinois.edu \\
  \addr University of Illinois Urbana-Champaign
  \AND
  \name Mostofa Patwary \email mpatwary@nvidia.com \\
  \addr NVIDIA
  \AND
  \name Mohammad Shoeybi \email mshoeybi@nvidia.com \\
  \addr NVIDIA
  \AND
  \name Bryan Catanzaro \email bcatanzaro@nvidia.com \\
  \addr NVIDIA
  \AND
  \name Jiaxuan You \email jiaxuan@illinois.edu \\
  \addr University of Illinois Urbana-Champaign
}

\def\month{07}
\def\year{2026}
\def\openreview{\url{https://openreview.net/forum?id=N951scS3yE}}
\usepackage{xcolor}
\newcommand{\revise}[1]{\textcolor{black}{#1}}
\begin{document}
\let\thefootnote\relax\footnotetext{\textsuperscript{*}\,Equal contribution.}
\maketitle

\vspace{-5mm}
\begin{abstract}
\vspace{-2mm}

The rapid advancement of large language models (LLMs) has created a diverse landscape of models, each excelling at different tasks. 
This diversity drives researchers to employ multiple LLMs in practice, leaving behind valuable multi-LLM log data. 
This naturally leads to the question of whether such logs can be fully leveraged to fuse LLMs’ complementary capabilities.
Although prior work has explored various strategies for integrating multiple LLMs, we argue that practical fusion must meet two essential requirements: (1) \textit{compatibility with real-world serving scenarios} (\textit{e.g.}, local and API-based serving), and (2) \textit{flexibility to operate at different stages of the LLM pipeline to meet varied user needs} (\textit{e.g.}, fine-tuning and inference stages).
To this end, we introduce \bench, a large-scale benchmark for LLM fusion that spans 14 tasks across \revise{six} domains, with responses from 20 open-source LLMs (8B–671B) totaling 103M tokens. Building on \bench, we propose \method, a systematic framework with three elaborated levels: \textbf{(1) query-level fusion} via tailored LLM routers, \textbf{(2) thought-level fusion} leveraging retrieved abstract reasoning templates, and \textbf{(3) model-level fusion} via distillation from top-ranked responses. 
Experiments show that \method consistently outperforms the best individual LLM across all 14 benchmarks, with the optimal fusion configuration varying across benchmarks, highlighting the promise of multi-LLM log data as a practical foundation for fusing diverse LLM capabilities.

\end{abstract}
\begin{center}
\faGithub\ \href{https://github.com/ulab-uiuc/FusionFactory}{\texttt{ulab-uiuc/FusionFactory}} \quad
\includegraphics[height=0.9em]{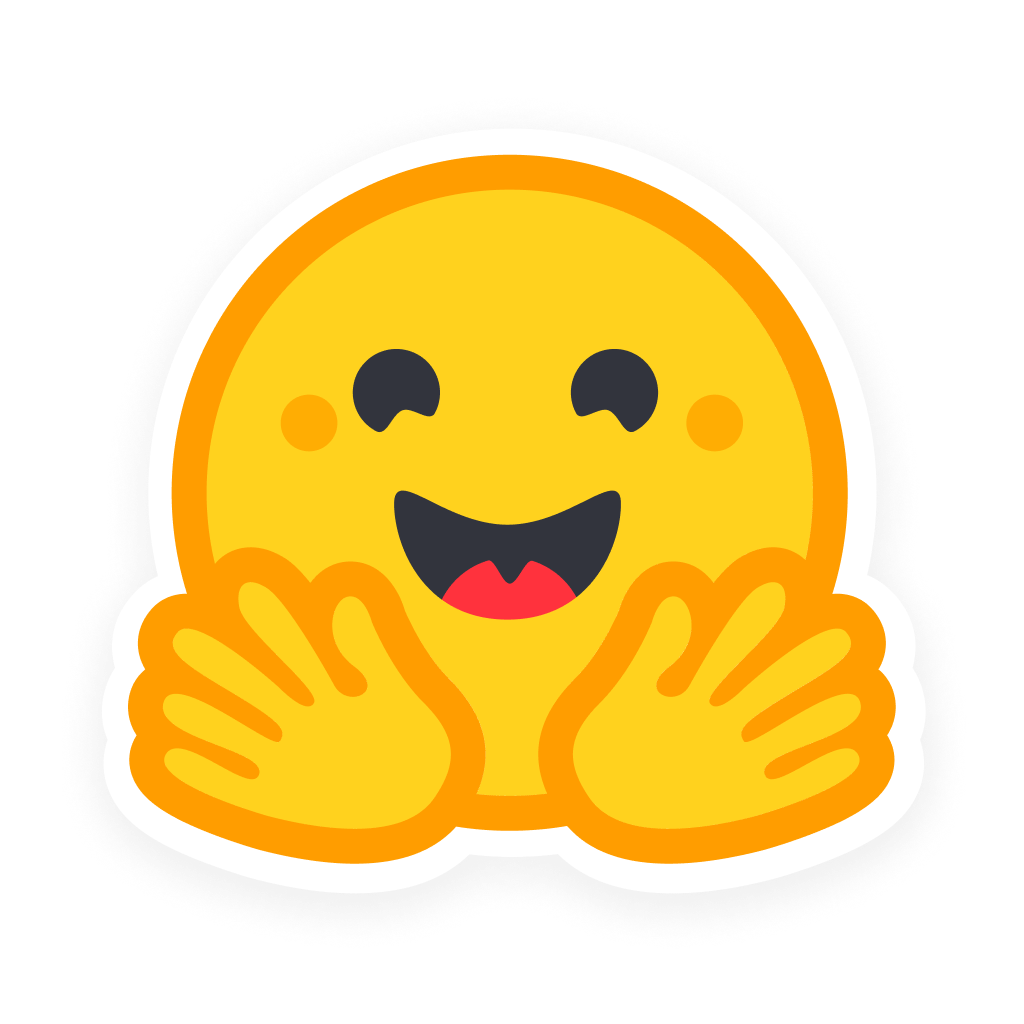}~\href{https://huggingface.co/datasets/ulab-ai/FusionBench}{\texttt{Hugging Face Collection}}
\end{center}


\vspace{-2mm}
\section{Introduction}
\label{sec:intro}
\vspace{-2mm}

Large language models (LLMs), with differences in architecture design, training data, and optimization objectives, often excel at different tasks \citep{ahmed2024studying,zhang2024pybench,singhal2022large,luo2022biogpt}. For example, DeepSeek-R1 achieves strong results in code generation \citep{guo2025deepseek}, while Claude 3.7 Sonnet is particularly effective in factual retrieval and knowledge coverage \citep{anthropic2025claude37}; our work also verifies this diversity of LLM performance (Figure \ref{fig:2.2} and \ref{fig:Histogram}).
Due to this diversity, researchers tend to utilize multiple LLMs in their workflow, notably in LLM API server platforms and agentic workflows~\citep{zhang2024aflow}; such usage pattern leaves us with valuable multi-LLM log data.
\revise{Here, multi-LLM log data refers to structured serving records produced when multiple LLMs are invoked for user queries, rather than an ordinary collection of independent model responses. Such records preserve not only query-response pairs, but also model identity, task or domain information, response mode, inference cost, and quality feedback, which together characterize the behavior of different LLMs under aligned query settings.}
We point out that these multi-LLM log data are a valuable vault, leading to an important research question: \textit{can we fully leverage multi-LLM log data to fuse LLMs' complementary capabilities?}

To address this challenge, prior work has explored several different ways of integrating multiple LLMs. Some studies investigate multi-agent systems, where LLMs collaborate through division of labor \citep{talebirad2023multi,han2024llm,chen2023agentverse}. Others develop LLM routers that dynamically assign queries to the most suitable model \citep{ding2024hybrid,dai2024cost,chen2023frugalgpt,ong2024routellm,feng2024graphrouter}, while some existing model fusion approaches, including naive ensembles \citep{haber2013model,li2023deep} and model merging~\citep{yang2024modelmerging}, aim to leverage multiple LLMs at the output and parameter levels, respectively. 
Despite their different formulations, these approaches share the same underlying goal of harnessing complementary strengths among LLMs. 
\revise{However, most existing methods study only one specific fusion stage in isolation. For example, routing methods mainly use multi-LLM records for query-level model selection, while distillation or model-fusion methods mainly focus on transferring responses or knowledge into a target model. As a result, they do not fully answer a practical question faced by users: given the same source of multi-LLM log data, should it be used for routing, thought-level prompting, or model-level distillation?}
While these explorations mark an important step forward, fundamental challenges remain in building LLM fusion approaches that are both broadly effective and practically feasible.
These challenges motivate a closer examination of what properties an effective fusion framework should possess.

In this work, we argue that effective fusion methods should ideally satisfy two key requirements. 
\textit{First, effective fusion should be compatible with different LLM serving scenarios}: 
Real-world LLM servicing, broadly speaking, can be categorized into local serving (when model weights are available) and API-based serving (when models are closed-source or expensive to run).
While LLM distillation~\citep{xu2024distillation-survey} has emerged as a popular paradigm for fusion, existing methods typically rely on logits or internal states, which are inaccessible in the generic LLM serving scenarios (\textit{e.g.}, API-based serving). 
In contrast, multi-LLM log data, which is stored in text format, offers the most practical, cost-effective, and scalable foundation for studying LLM fusion, and thus forms the focus of this work.
\revise{Different from generic multi-response collections that mainly contain several outputs for the same input, the log data considered in this work contains structured metadata and feedback signals associated with LLM serving, including model identity, response mode, cost, task-specific performance, and LLM-based quality scores. These signals are directly useful for learning routing policies, retrieving reusable reasoning patterns, and selecting high-quality responses for distillation.}
\textit{Second, effective fusion should happen at different stages in the LLM pipeline to accommodate different users' needs}.
In practice, users may or may not have the privilege of finetuning an LLM (due to cost and model availability considerations).
Moreover, different tasks may require an LLM to fuse at different stages; for example, a RAG-based LLM system works more favorably with retrieval-based thought-level LLM fusion.
In this work, we aim to systematically understand the trade-offs of LLM fusion at different stages.
\revise{This stage-aware perspective is central to our motivation. API-only compatibility is not the sole novelty of our work; rather, we aim to provide a unified benchmark and framework for studying how the same structured multi-LLM logs can be exploited at different intervention points in the LLM pipeline. This motivates the fusion-stage matrix studied in this paper, which compares query-level, thought-level, and model-level fusion under a shared log source and evaluation protocol.}

\begin{table}[t]
    \vspace{-6mm}
    \caption{\textbf{Comparison of \method with existing works leveraging multi-LLM log data.} 
    We evaluate along three fusion stages, API serving compatibility, and task domain coverage. \method supports all three fusion stages, is compatible with API-based serving, and covers six task domains, demonstrating broader applicability and flexibility compared to prior work.}
    \label{Tab:FusionBench_intro}
    \centering
    \setlength{\tabcolsep}{12pt}
    \resizebox{\textwidth}{!}{%
    \begin{tabular}{lccccc}
        \toprule
        \multirow{2}{*}{\textbf{Method}} & \multicolumn{3}{c}{\textbf{LLM Fusion Stage}} & \multirow{2}{*}{\shortstack{\textbf{API Serving}\\\textbf{Compatibility}}} & \multirow{2}{*}{\shortstack{\textbf{Task Domain}\\\textbf{Coverage}}} \\
        \cmidrule(lr){2-4}
         & \textbf{Query-level} & \textbf{Thought-level} & \textbf{Model-level} &  &  \\
        \midrule
        RouteLLM \citep{ong2024routellm} & \textcolor{forestgreen}{\textbf{\ding{51}}} & \textcolor{red}{\textbf{\ding{55}}} & \textcolor{red}{\textbf{\ding{55}}} & \textcolor{forestgreen}{\textbf{\ding{51}}} & 3 \\
        GraphRouter \citep{feng2024graphrouter} & \textcolor{forestgreen}{\textbf{\ding{51}}} & \textcolor{red}{\textbf{\ding{55}}} & \textcolor{red}{\textbf{\ding{55}}} & \textcolor{forestgreen}{\textbf{\ding{51}}} & 4 \\
        RouterBench \citep{hu2024routerbench} & \textcolor{forestgreen}{\textbf{\ding{51}}} & \textcolor{red}{\textbf{\ding{55}}} & \textcolor{red}{\textbf{\ding{55}}} & \textcolor{forestgreen}{\textbf{\ding{51}}} & 5 \\
        LLM-Blender \citep{jiang2023Llm-blender} & \textcolor{forestgreen}{\textbf{\ding{51}}} & \textcolor{red}{\textbf{\ding{55}}} & \textcolor{red}{\textbf{\ding{55}}} & \textcolor{forestgreen}{\textbf{\ding{51}}} & 4 \\
        FuseLLM \citep{wan2024fusellm} & \textcolor{red}{\textbf{\ding{55}}} & \textcolor{red}{\textbf{\ding{55}}} & \textcolor{forestgreen}{\textbf{\ding{51}}} & \textcolor{red}{\textbf{\ding{55}}} & 3 \\
        \midrule
        \rowcolor{cyan!10} \method & \textcolor{forestgreen}{\textbf{\ding{51}}} & \textcolor{forestgreen}{\textbf{\ding{51}}} & \textcolor{forestgreen}{\textbf{\ding{51}}} & \textcolor{forestgreen}{\textbf{\ding{51}}} & \textbf{6} \\
        \bottomrule
    \end{tabular}}
    \vspace{-4mm}
\end{table}

\revise{As illustrated in Table~\ref{Tab:FusionBench_intro}, the unique value of \method is not merely API-serving compatibility, since several prior routing methods also support API-only usage. Instead, \method differs by systematically studying how multi-LLM log data can support multiple fusion stages under a unified framework. Prior routing methods such as RouteLLM, GraphRouter, RouterBench, and LLM-Blender mainly focus on query-level fusion, while FuseLLM mainly studies model-level fusion. In contrast, \method jointly covers query-level, thought-level, and model-level fusion, which enables a cross-stage comparison of when and how the same log data should be used.}


Based on the above observations, we propose \bench, a large-scale benchmark designed to support comprehensive studies on LLM fusion. \bench provides rich supervision by including both direct and reasoning-augmented responses from multiple LLMs, as well as reusable thought templates summarized from the top-performing models. Covering 14 tasks across 6 domains, \bench is built from 20 open-source LLMs ranging from 8B to 671B, making it one of the most diverse resources for studying multi-LLM integration (Table~\ref{tab:2.1} and~\ref{tab:2.2}).
\revise{In this sense, \bench operationalizes the above notion of multi-LLM log data: for each query, it stores aligned responses from multiple LLMs together with their response modes, performance scores, inference costs, and LLM-judge scores. This structure distinguishes \bench from ordinary multi-response datasets and provides the supervision signals needed for flexible capability fusion.}

Building on \bench, we introduce \method, a systematic framework for LLM fusion across three representative levels (comparison shown in Table \ref{Tab:FusionBench_intro}).
\revise{\method leverages different components of the structured log records at different fusion stages. Query-level fusion uses query and task information, model identity, performance, cost, and judge scores to learn routing decisions. Thought-level fusion uses historical queries, response modes, and top-ranked reasoning traces to construct reusable thought templates. Model-level fusion uses quality-ranked logged responses as distillation data.}
\textbf{(1) Query-level fusion} adapts to the varying strengths of different models and the diverse user preferences regarding token generation costs by constructing routers tailored to each query, enabling finer-grained capability fusion. Unlike traditional methods, we incorporate both direct outputs and reasoning-augmented responses~\citep{yuan2024advancing,besta2024graph,yao2023tree,wei2022chain} to better address complex user queries.
\textbf{(2) Thought-level fusion} improves new responses by retrieving abstract thought templates (\textit{i.e.}, concise summaries distilled from top-performing model outputs) based on embedding similarity to past queries. This design avoids the noise of full past responses while still capturing useful reasoning patterns, which are then used as few-shot demonstrations for new queries~\citep{song2023llm,kang2023large,li2024flexkbqa,zhao2021calibrate}.
\textbf{(3) Model-level fusion} takes a distillation view~\citep{sreenivas2024llm,stichlmair2021distillation}, transferring complementary capabilities into a single base model by conducting supervised fine-tuning on high-quality responses aggregated across LLMs. Inspired by imitation learning in the LLM distillation domain~\citep{hussein2017imitation,osa2018algorithmic}, for each query, we select the LLM responses with the top-k performance or the highest LLM judge scores as training data. This enables downstream deployment of a fused model without relying on multiple LLMs at inference time.

We systematically evaluate \method on \bench and find that \textit{fusing LLMs with \method yields consistent benchmark performance improvement over the best individual LLM across 14 popular LLM benchmarks}, where the optimal \method configuration varies across different benchmarks.
Specifically, we find that: (1) \textit{thought-level fusion achieves the best overall performance}, especially when using hybrid selection and large summaries; (2) \textit{model-level fusion performs the worst among the three levels}, likely due to overfitting and difficulty generalizing across tasks; (3) \textit{query-level fusion offers a good balance between performance and efficiency}, achieving decent results while requiring minimal computational overhead; (4) \textit{gains from fusing World Knowledge and Math domains are moderate}, where these domains demand high factual accuracy or strict logical consistency, which are difficult to maintain through multi-LLM fusion.
These results highlight the promise of multi-LLM log data as a practical foundation for developing fusion methods that flexibly adapt to diverse serving scenarios and fusion stages.
\revise{Moreover, the cross-level comparison enabled by our fusion-stage matrix provides insights that cannot be obtained by studying routing or distillation alone. It reveals that different ways of using the same multi-LLM log data lead to different performance, cost, and generalization behaviors, thereby providing a practical decision framework for choosing among routing-based, thought-based, and training-based fusion under different deployment constraints.}



\section{\bench: Benchmarking LLM Capability Fusion with Multi-LLM Log Data}
\label{sec:bench}

\begin{figure}[t]
    \centering
    \vspace{-6mm}
    \includegraphics[width=\linewidth]{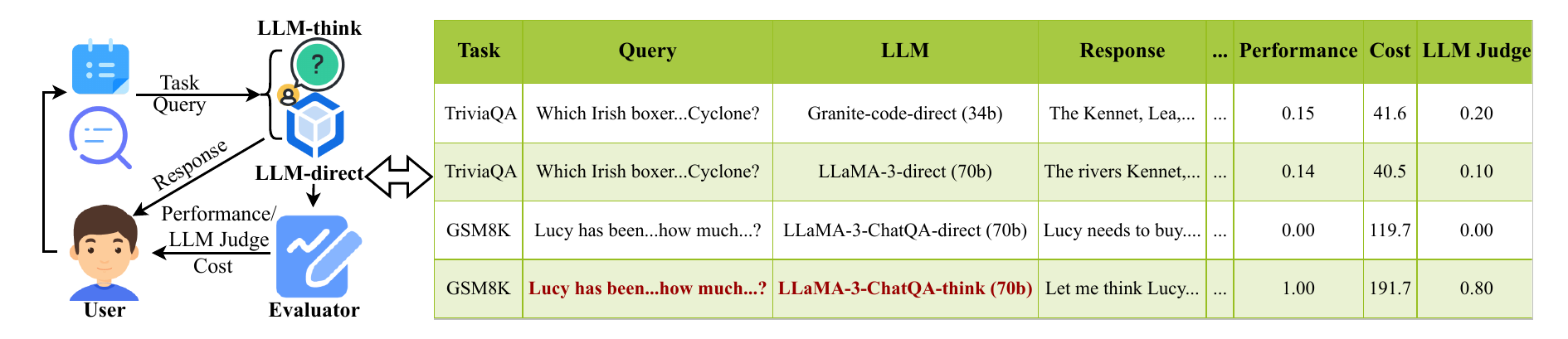}
    \caption{\textbf{Overview of \bench's construction process.} For each task and user query, multiple LLMs generate responses in two modes: direct and reason (\textit{i.e.}, chain-of-thoughts (CoT)~\citep{wei2022chain}). These responses are then evaluated for performance, cost, and judge scores.}
    \vspace{-3mm}
\label{fig:2.1}
\end{figure}

\begin{table}[t]
  \centering
  \begin{minipage}[t]{0.46\textwidth}
    \centering
    \captionof{table}{\textbf{\bench covers 14 tasks across 6 representative domains}.}
    \label{tab:2.1}
    \begin{scriptsize}
      \setlength{\tabcolsep}{1pt}
    \resizebox{\linewidth}{!}{        
      \begin{tabularx}{\linewidth}{@{}p{3.5cm} X@{}}
        \toprule
        \textbf{Domain} & \textbf{Tasks} \\
        \midrule
        Math                  & GSM8K, MATH \\
        \midrule
        Code                  & MBPP, HumanEval \\
        \midrule
        Commonsense Reasoning &   \makecell[l]{CommonsenseQA, ARC,\\ HellaSwag, OpenBookQA} \\
        \midrule
        World Knowledge       & NaturalQuestions, TriviaQA \\
        \midrule
        Reading Comprehension  & SQuAD,  BoolQ \\
        
        \midrule
        Popular               & MMLU, GPQA \\
        \bottomrule
      \end{tabularx}
      }
    \end{scriptsize}
  \end{minipage}\hfill
  \begin{minipage}[t]{0.5\textwidth}
    \centering
    \captionof{table}{\textbf{\bench includes 20 LLMs grouped into three size-based scales.}}
    \label{tab:2.2}
    \begin{scriptsize}
    \begin{tabularx}{\linewidth}{@{} >{\centering\arraybackslash}m{1cm} X @{}}
      \toprule
      \textbf{Scale} & \textbf{LLMs} \\
      \midrule
      \multirow{2}{*}{Small} & 
      Qwen2 (7b), Mistral (7b), Qwen2.5 (7b), Gemma (7b), CodeGemma (7b), Gemma-2 (9b), LLaMA-3.1 (8b), Granite (8b), LLaMA-3 ChatQA (8b), Mistral-Nemo (12b) \\
      \midrule
      \multirow{2}{*}{Medium} & 
      LLaMA-3.3 Nemotron Super (49b), Granite Code (34b), LLaMA-3.1 Nemotron (51b), LLaMA-3 ChatQA (70b), LLaMA-3.1 (70b), LLaMA-3 (70b), Mixtral (8x7b) \\
      \midrule
      Large & 
      Palmyra Creative (122b), Mixtral (8x22b), DeepSeek-R1 (671b) \\
      \bottomrule
    \end{tabularx}
    \end{scriptsize}
  \end{minipage}
\end{table}

In this section, we detail the construction of \bench (Section~\ref{sec:2.1}) and its support for LLM capability fusion (Section~\ref{sec:2.2}), as illustrated in Figure~\ref{fig:2.1}.

\subsection{\bench Construction} \label{sec:2.1} 


As shown in Figure~\ref{fig:2.1}, we construct \bench by sampling up to 500 training and 50 partial test queries (10:1 ratio) from each of 14 tasks across six domains (Table~\ref{tab:2.1}). For each query, responses from multiple LLMs are collected to support comparison and exploration of query-, thought-, and model-level fusion (Sections~\ref{sec:4},~\ref{sec:thoughtfusion}, and~\ref{sec:model}). For the final cross-level analysis (Section~\ref{sec:cross}), we also use each task’s full test set to comprehensively evaluate fusion levels. Table~\ref{app_tab:2} in the Appendix summarizes sample counts for training, partial test, and full test sets.

\revise{We formalize the resulting multi-LLM log data as structured serving records rather than ordinary response collections. Specifically, each log entry is represented as $(q,t,m,s,r,c,y,j)$, where $q$ denotes the user query, $t$ denotes the task or domain, $m$ denotes the LLM identity, $s$ denotes the response mode such as direct or reasoning-augmented prompting, $r$ denotes the generated response, $c$ denotes the inference cost, $y$ denotes the task-specific evaluation result, and $j$ denotes the LLM-judge score. Under this definition, \bench is not simply a set of $(q,r)$ pairs, but an aligned multi-LLM serving log that records model behavior, response style, cost, and quality feedback for the same query.}


As summarized in Table~\ref{tab:2.2}, we select 20 LLMs across varying scales. \revise{To elicit diverse responses, we apply distinct system prompts for direct and reasoning-based outputs. The 20-model pool contains both LLMs with native trained think-token behavior (e.g., DeepSeek-R1 and the Nemotron \texttt{-think} variants) and LLMs without it. We apply the same chain-of-thought elicitation procedure uniformly across all 20 models to enable apples-to-apples comparison across the pool; for models with trained think-token behavior the model's native reasoning style is preserved when responding to our prompt, while for models without it the \texttt{-think} responses are produced via prompted CoT.} Responses are evaluated with task-specific metrics (Table~\ref{app_tab:1}), but as shown in Figure~\ref{tab:performance_and_LLM_judge}, such metrics may favor short, low-information answers that hinder effective fusion. To address this, we introduce an LLM judge that scores the insightfulness of correct answers (prompt in Table~\ref{prompt-template:llm judge}). As shown in Table~\ref{tab:performance_and_LLM_judge}, the LLM judge better reflects informative value, offering greater utility for fusion.


To assess cost, we follow~\citep{feng2024graphrouter} by estimating total token usage via GPT-2 \citep{radford2019language} and converting it into monetary cost using Together AI’s API pricing (Table~\ref{tab:llm-costs}). As shown in the right of Figure~\ref{fig:2.1}, collected data is stored in a DataFrame, with each row capturing task, query, LLM, response, performance, cost, and LLM judge score (details in Table~\ref{tab:data_collection_info}).

\revise{This construction focuses on a practical subset of log signals that can be consistently collected in both local and API-based LLM serving scenarios. Richer production logs may additionally contain temporal order, explicit user feedback, error patterns, and multi-turn interaction history. These signals are valuable, but they are not always available across different serving settings, especially when using closed-source or API-based models. Therefore, \bench emphasizes broadly accessible log attributes, including model identity, response mode, cost, task-specific performance, and LLM-based quality feedback, while leaving interaction-level signals such as temporal dynamics and user feedback as important extensions.}

\begin{figure}[t]
    \centering
    \vspace{0mm}
    \includegraphics[width=1\linewidth]{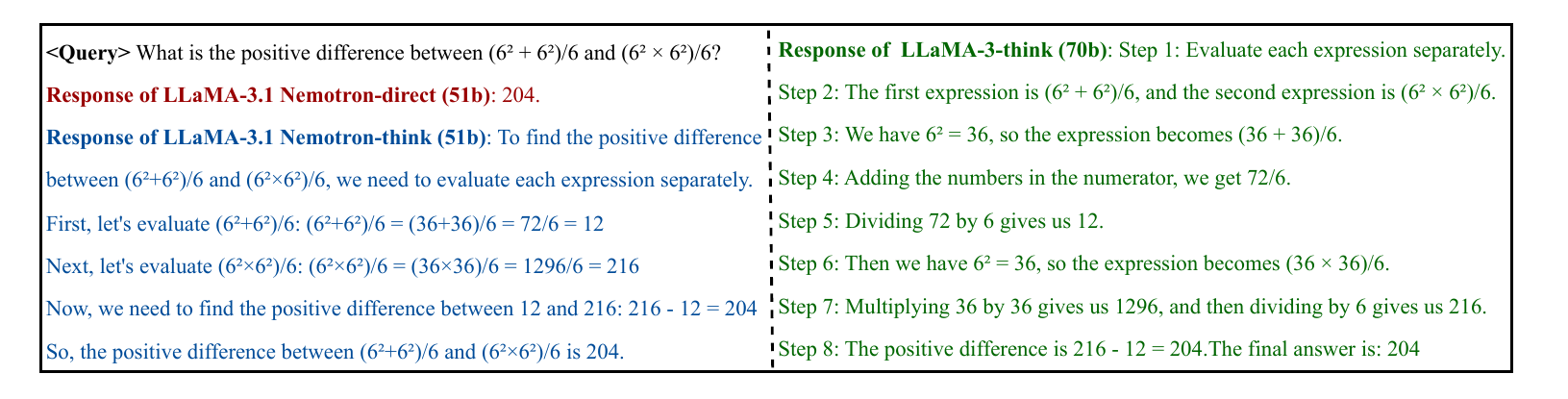}
    \vspace{0mm}
    \caption{\textbf{LLMs exhibit varying reasoning styles when answering the same query.} This example from the Math task compares responses from LLaMA-3.1 Nemotron-direct (51B), Nemotron-think (51B), and LLaMA-3-think (70B), where -think denotes CoT prompting. The responses show a progression from concise answers to increasingly detailed reasoning.}  
\label{fig:2.2}
\end{figure}

\begin{figure*}[t]
    \centering
    \begin{subfigure}[t]{0.32\textwidth}
        \includegraphics[height=4.2cm]{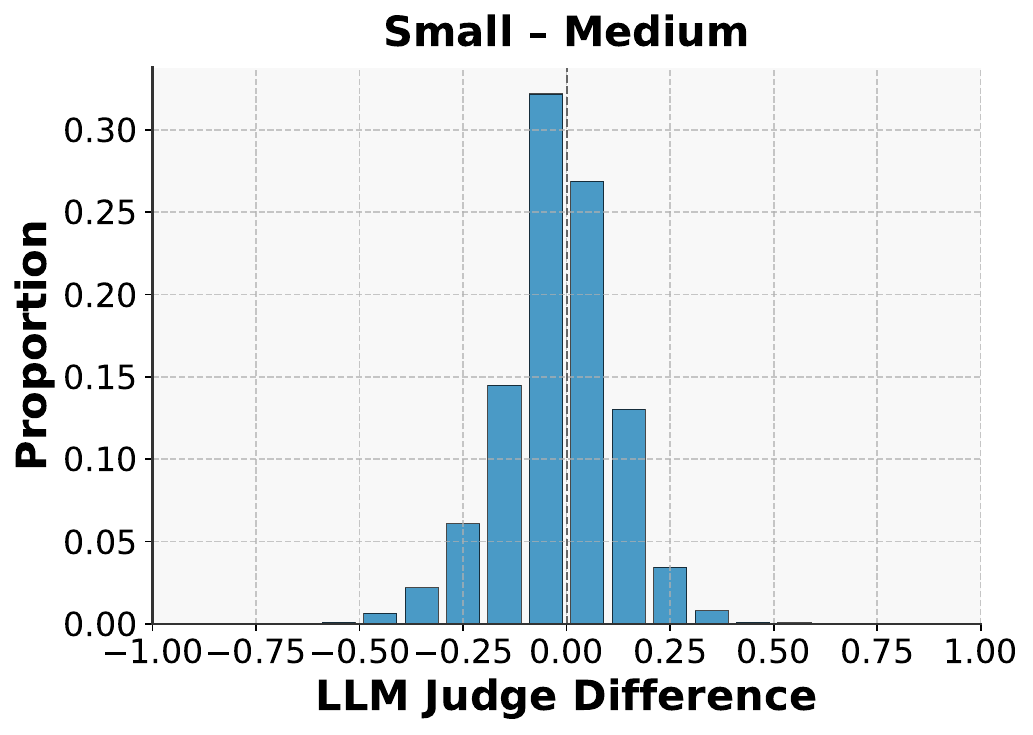}
        \label{fig:step1}
    \end{subfigure}
    \hspace{1pt}
    \begin{subfigure}[t]{0.32\textwidth}
        \includegraphics[height=4.2cm]{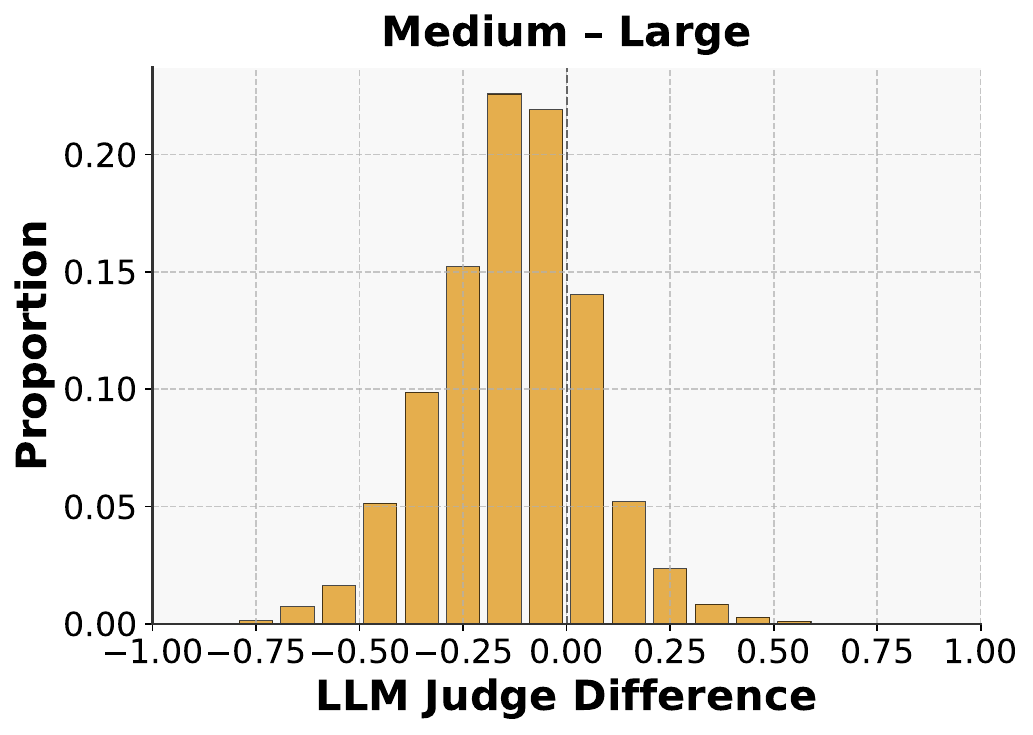}
        \label{fig:step2}
    \end{subfigure}
    \hspace{1pt}
    \begin{subfigure}[t]{0.32\textwidth}
        \includegraphics[height=4.2cm]{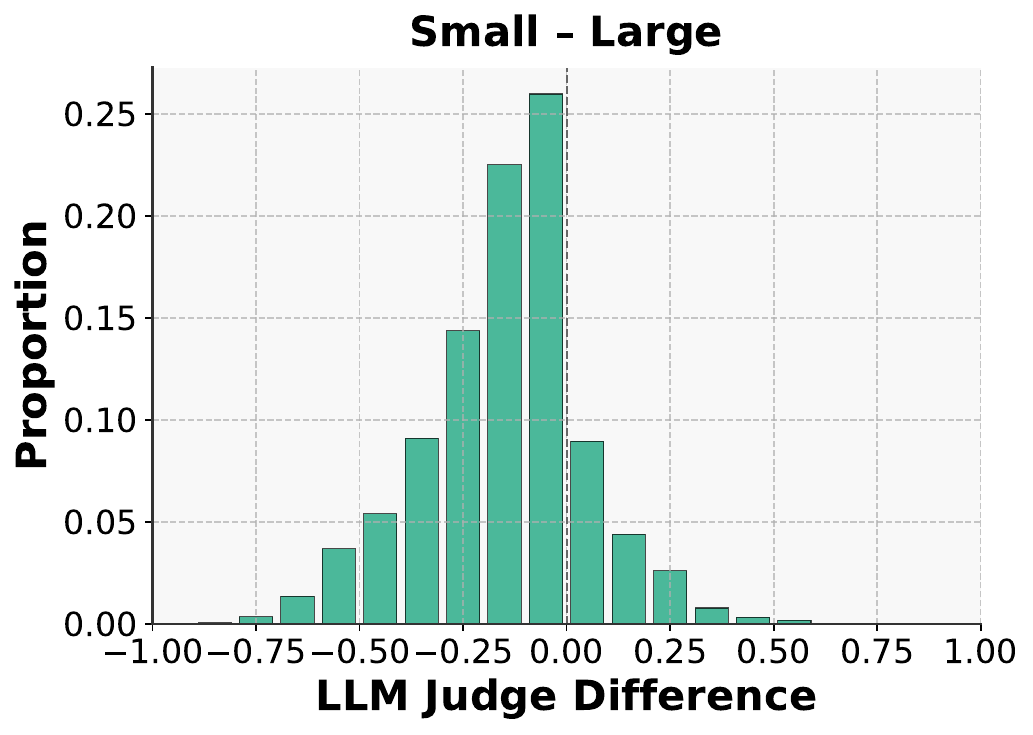}
        \label{fig:step3}
    \end{subfigure}
    \vspace{0mm}
    \caption{\textbf{LLMs of different scales show distinct strengths across queries.} For each query, we compute average LLM judge scores for small, medium, and large models (Table~\ref{tab:2.2}), and subtract these scores pairwise (small--medium, medium--large, small--large) to measure relative advantages. We then visualize the resulting score differences as three histograms.}
    \label{fig:Histogram}
\end{figure*}

\subsection{Support for LLM Capabilities Fusion} \label{sec:2.2}


As introduced earlier, \bench captures diverse LLM responses to the same query, enabling potential for LLM capability fusion. To illustrate this, we conduct a qualitative case study using a Math query (Figure~\ref{fig:2.2}). We compare responses from LLaMA-3.1 Nemotron-direct (51B), LLaMA-3.1 Nemotron-think (51B), and LLaMA-3-think (70B), where -think denotes CoT prompting. These responses exhibit distinct styles: direct and concise, analytical reasoning, and detailed explanation, respectively. These variations highlight opportunities to learn diverse reasoning patterns and enhance fusion effectiveness.

\revise{The structured log format is crucial for this capability-fusion analysis. Because different LLMs are evaluated on the same query with shared task-specific metrics, cost estimates, and judge scores, \bench allows us to compare aligned model behaviors rather than unrelated responses. Model identity helps identify which LLM is suitable for which type of query, response mode distinguishes direct answering from reasoning-augmented generation, and performance or judge feedback provides supervision for selecting useful responses. These log properties are later used by query-level fusion for routing, thought-level fusion for retrieving and summarizing reasoning templates, and model-level fusion for selecting distillation data.}


To complement our analysis, we conduct quantitative evaluations using LLM judge scores. As shown in Figure~\ref{fig:Histogram}, we compute the average judge score for each model scale per query (based on Table~\ref{tab:2.2}), and then calculate pairwise differences: small-medium, medium-large, and small-large. The resulting histograms reveal distinct performance gaps across scales, suggesting that models of different sizes exhibit complementary strengths and further motivating fusion via \bench.
\section{\method: Flexible Fusion of LLM Capabilities}
\label{sec:method}

\begin{figure}[t]
    \centering
    \vspace{0mm}
    \includegraphics[width=\linewidth]{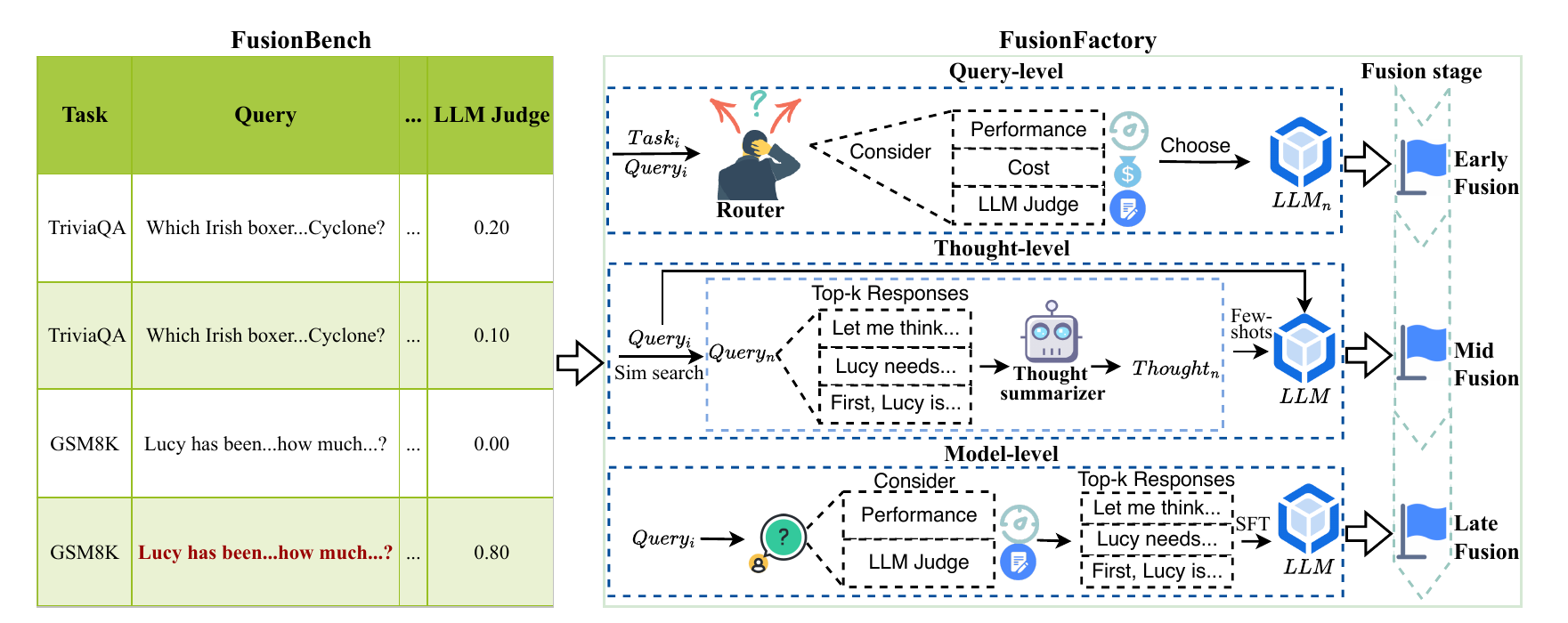}
    \vspace{0mm}
\caption{\textbf{Overview of \method via \bench across three fusion stages:} \textit{query-}, \textit{thought-}, and \textit{model-level}, corresponding to early, mid, and late fusion. At the query level, a router selects the best LLM per query using performance, cost, and judge scores. At the thought level, a summarizer extracts thought templates from top-k responses for few-shot prompting. At the model level, we fine-tune a base LLM on high-quality responses to encode fused capabilities.}
\label{fig:Overview method}
\vspace{0em}
\end{figure}

As shown in Figure~\ref{fig:Overview method}, we present \method across three representative levels of LLM capabilities fusion: query-, thought-, and model-level, corresponding to early, mid, and late fusion stages. Each level targets a distinct phase in the reasoning process.

\revise{This stage-aware organization provides the conceptual taxonomy used in this empirical study. We do not aim to provide formal theoretical guarantees or closed-form complexity bounds; instead, our goal is to systematically compare practical fusion strategies according to where they intervene in the LLM pipeline. Query-level fusion intervenes before generation by selecting a model or response mode for each query. Thought-level fusion intervenes during generation by retrieving and injecting reusable reasoning templates. Model-level fusion intervenes after response collection by consolidating selected logged responses into model parameters through supervised fine-tuning. This taxonomy clarifies that the three levels are not arbitrary engineering categories, but distinct intervention points in the LLM inference and learning process.}

\revise{This stage-aware organization is also the basis of the fusion-stage matrix studied in this paper. Rather than treating API-only compatibility as the main novelty, \method aims to answer a broader question: given the same structured multi-LLM log data, should it be used before generation for routing, during generation for reasoning augmentation, or after response collection for model training? As illustrated in Figure~\ref{fig:Overview method}, query-level fusion is an early-stage intervention before response generation, thought-level fusion is a mid-stage intervention during reasoning, and model-level fusion is a late-stage intervention that consolidates logged responses into model parameters. This matrix therefore provides a unified way to compare different uses of multi-LLM logs under the same benchmark and evaluation protocol.}

\revise{Importantly, the three levels leverage different components of the structured multi-LLM log records introduced in Section~\ref{sec:2.1}. Query-level fusion uses query and task information, model identity, response mode, performance, cost, and LLM-judge scores to learn routing decisions. Thought-level fusion uses historical queries, response modes, generated responses, and quality-ranked reasoning traces to construct reusable thought templates. Model-level fusion uses task-specific performance or LLM-judge feedback to select high-quality logged responses as supervised fine-tuning data. Therefore, \method does not treat multi-LLM logs as ordinary text-only response collections; instead, it exploits their structured metadata and feedback signals to support different forms of capability fusion.}

\textbf{Query-level Fusion (Early Fusion)} (Section~\ref{sec:Query-level fusion}) operates at the problem interpretation stage. A router analyzes the query and task characteristics to select the best LLM configuration, considering performance, cost, and LLM-based evaluations. The selected model then processes the query independently, enabling efficient task-specialized execution.
\textbf{Thought-level Fusion (Mid Fusion)} (Section~\ref{sec:Thought-level fusion}) intervenes during reasoning. An LLM summarizer distills abstract thought templates from top-k responses (ranked by performance or LLM judgment). For new queries, we retrieve similar historical queries and apply their thought templates as few-shot prompts, allowing dynamic coordination across reasoning styles.
\textbf{Model-level Fusion (Late Fusion)} (Section~\ref{sec:Model-level fusion}) occurs after full responses are generated. A base LLM is fine-tuned using top-k high-quality outputs (selected via performance or LLM judgment) to internalize fused capabilities. This stage reflects post-hoc fusion via training rather than online collaboration.
\textbf{These three fusion stages offer complementary strengths at different points in the reasoning pipeline, allowing users to flexibly choose the most suitable level of integration based on their specific serving constraints, performance goals, and resource availability.}

\revise{From a practical computational and deployment perspective, the three stages expose different trade-offs. Query-level fusion has the lowest inference overhead because it only requires a lightweight routing step before calling one selected LLM. Thought-level fusion introduces additional retrieval and prompt-token overhead, but it avoids model fine-tuning and remains compatible with API-based LLMs. Model-level fusion requires offline training and access to a trainable base model, but after training it only requires a single fused model at inference time. We present these as practical deployment trade-offs rather than formal complexity bounds, because exact theoretical costs are difficult to define uniformly across heterogeneous LLM APIs, model sizes, serving infrastructures, and pricing schemes.}

\revise{This comparison is necessary because existing routing and distillation methods typically focus on one stage in isolation. Routing-based methods mainly study query-level model selection, while distillation-based methods mainly study model-level capability transfer. In contrast, \method uses the same multi-LLM log source to study all three stages together. This enables cross-stage analysis of performance, cost, and generalization behavior, and provides a practical decision framework for users who face different deployment constraints.}

\section{Query-level \method Fusion} 
\label{sec:Query-level fusion} 

\subsection{Methodology: Query-level Fusion}


The goal of query-level fusion is to assign the most suitable LLM for each user query, fusing LLMs' capabilities through a router. This router selects models based on the query, task info, and LLM features, jointly considering performance, cost, and LLM judge scores. \revise{Under the structured log formulation in Section~\ref{sec:2.1}, this level directly uses the log fields related to the input and serving decision, including the query $q$, task or domain $t$, model identity $m$, response mode $s$, cost $c$, task-specific performance $y$, and LLM-judge score $j$. To clarify the routing notation, we define each router action as $a=(m,s)\in\mathcal{A}$, where $m$ denotes the selected LLM and $s$ denotes the selected response mode, such as direct or reasoning-augmented prompting. Thus, the router chooses an LLM-response-mode configuration rather than an unspecified model feature.} As shown in the first panel of Figure~\ref{fig:Overview method}, we follow \citet{feng2024graphrouter} and define the reward for selecting action $a$ for query-task pair $(q,t)$ as:
\begin{equation}
\label{eq:Reward}
\text{Reward}(q,t,a) = \alpha \cdot P(q,t,a) - \beta \cdot C(q,t,a) + \gamma \cdot J(q,t,a),
\end{equation}
\revise{where $P(q,t,a)$ denotes the task-specific performance score obtained by applying the evaluation metric of task $t$ to the response generated under action $a$; $C(q,t,a)$ denotes the inference cost, estimated from token consumption and API pricing; and $J(q,t,a)$ denotes the LLM-judge score, which measures the informativeness or usefulness of the generated response beyond the task-specific metric. The coefficients $\alpha$, $\beta$, and $\gamma$ control the relative importance of task performance, inference cost, and LLM-judge quality, respectively. The negative sign before $C(q,t,a)$ reflects that lower inference cost is preferred, while higher performance and judge scores are preferred.}
Based on this, for a query $q$ and its corresponding task $t$, our goal is to learn a router model $f_{\phi}$ such as a neural network that maps $(q,t)$ to an action $a\in\mathcal{A}$ and maximizes the expected reward:
\begin{equation}
\label{eq:router_obj_joint}
\phi^{*} = \arg\max_{\phi} \; \mathbb{E}_{(q,t) \sim \mathcal{D}} 
\left[ 
\text{Reward}\left(q,t,f_{\phi}(q,t)\right) 
\right],
\end{equation}
\revise{where \(\mathcal{D}\) denotes the distribution of query-task pairs and $\mathcal{A}$ denotes the set of candidate LLM-response-mode configurations. After obtaining $\phi^{*}$, $f_{\phi^{*}}$ selects the action $a=f_{\phi^{*}}(q,t)$ for each new query-task pair, and the selected LLM configuration generates the final response. This revised formulation removes the ambiguous dot notation in the original reward function and explicitly connects the routing objective to the structured log fields used in query-level fusion.}

\revise{From the perspective of the fusion-stage matrix, query-level fusion offers a lightweight way to exploit multi-LLM logs in API-based serving settings. It does not require access to model weights, logits, or internal states; instead, it only requires logged performance, cost, and judge signals to learn a routing policy. Thus, this stage is suitable when users want to improve the accuracy--cost trade-off while still calling only one selected LLM at inference time.}

\subsection{Experimental Settings: Query-level Fusion} \label{sec:4}

\xhdr{Scenario}
To explore query-level fusion, we define four routing scenarios aligned with Section~\ref{sec:Query-level fusion}: \textit{Performance First}, \textit{Balance}, \textit{Cost First}, and \textit{LLM Judge}. The first three examine trade-offs between performance and cost, while the last assesses routing under LLM-as-judge evaluation. These scenarios differ in their reward weights $(\alpha, \beta, \gamma)$ in Equation~\ref{eq:Reward}, which control the emphasis on performance, cost, and LLM-judge scores and are set to $(1.0, 0.0, 0.0)$, $(0.5, 0.5, 0.0)$, $(0.2, 0.8, 0.0)$, and $(0.0, 0.0, 1.0)$, respectively.

\xhdr{Comparison Routers} We evaluate five representative routing methods for LLM fusion: \textit{RouterKNN} \citep{shnitzer2023large}, \textit{RouterSVM} \citep{hu2024routerbench}, \textit{RouterMLP} \citep{shnitzer2023large}, \textit{RouterBERT} \citep{ong2024routellm}, and \textit{GraphRouter} \citep{feng2024graphrouter}. 
We also include several static baselines: \textit{Best LLM}, which achieves the highest performance on the training set; \textit{LLM-Mean}, which yields the average test performance across all LLMs; and \textit{Largest LLM}, which denotes the test performance of the largest model. 
Detailed baseline description can be found in Appendix~\ref{app:query_level_fusion}.


\xhdr{Evaluation Metric} Following \citep{feng2024graphrouter, ong2024routellm}, we adopt three metrics: \textit{Performance}, the average task accuracy over queries; \textit{Cost}, the average inference expense of selected LLMs; and \textit{LLMScore}, a learned evaluator's quality rating over \(\langle\text{query},\;\text{ground truth},\;\text{ response}\rangle\) triples. Full LLMScore details are in Appendix~\ref{app:LLM-As-Judge_prompt}.

\begin{table*}[t]
    \centering
    \vspace{-6mm}
    \caption{
    \textbf{Multiple query-level fusion methods surpass the best single LLM across four scenarios.}
    \textbf{Bold} and \underline{underline} indicate the best and second-best scores. In the Performance First, Balance, and Cost First settings, we report Performance, Cost, and Reward to evaluate how well each router balances performance and cost. The LLM Judge scenario evaluates router ability to integrate informative responses, using LLMScore and Performance.
    }
    \vspace{0mm}
    \setlength\tabcolsep{2pt}
    \resizebox{\textwidth}{!}{%
    \begin{tabular}{cccccccccccc}
        \toprule
        \textbf{Scenario} 
        & \multicolumn{3}{c}{\textbf{Performance First}} 
        & \multicolumn{3}{c}{\textbf{Balance}} 
        & \multicolumn{3}{c}{\textbf{Cost First}} 
        & \multicolumn{2}{c}{\textbf{LLM Judge}} \\
        \cmidrule(lr){2-4}\cmidrule(lr){5-7}\cmidrule(lr){8-10}\cmidrule(lr){11-12}
        \textbf{Method} & \textbf{Performance} & \textbf{Cost} & \textbf{Reward} 
                       & \textbf{Performance} & \textbf{Cost} & \textbf{Reward}  
                       & \textbf{Performance} & \textbf{Cost} & \textbf{Reward}  
                       & \textbf{LLMScore} & \textbf{Performance} \\
        \midrule
        Best LLM & 0.546 & 0.1850 & 0.546 & 0.556 & 0.0681 & \underline{0.244} & 0.431 & 0.0047 & 0.0823 & 0.823 & 0.431 \\
        \midrule
        RouterKNN & 0.558 & 0.0666 & 0.558 & 0.461 & 0.0068 & 0.227 & 0.463 & 0.0067 & 0.0872 & 0.835 & \underline{0.470} \\
        RouterSVM  & 0.546 & 0.0673 & 0.546 & 0.448 & 0.0084 & 0.220 & 0.442 & 0.0075 & 0.0825 & 0.790 & 0.449 \\
        RouterMLP  & 0.561 & 0.0652 & \underline{0.561} & 0.482 & 0.0095 & 0.236 & 0.477 & 0.0078 & \underline{0.0884} & \underline{0.836} & 0.464 \\
        RouterBERT & 0.528 & 0.1030  & 0.528 & 0.387 & 0.0262 & 0.180 & 0.368 & 0.0531 & 0.0310 & 0.805 & 0.414 \\
        GraphRouter & 0.602 & 0.0184 & \textbf{0.602} & 0.464 & 0.0184 & \textbf{0.269} & 0.496 & 0.0047 & \textbf{0.0955} & \textbf{0.870} & \textbf{0.538} \\
        \bottomrule
    \end{tabular}
    }
    \label{tab:query-level-results}
\end{table*}

\begin{figure}[t]
    \centering
    \includegraphics[width=0.85\linewidth]{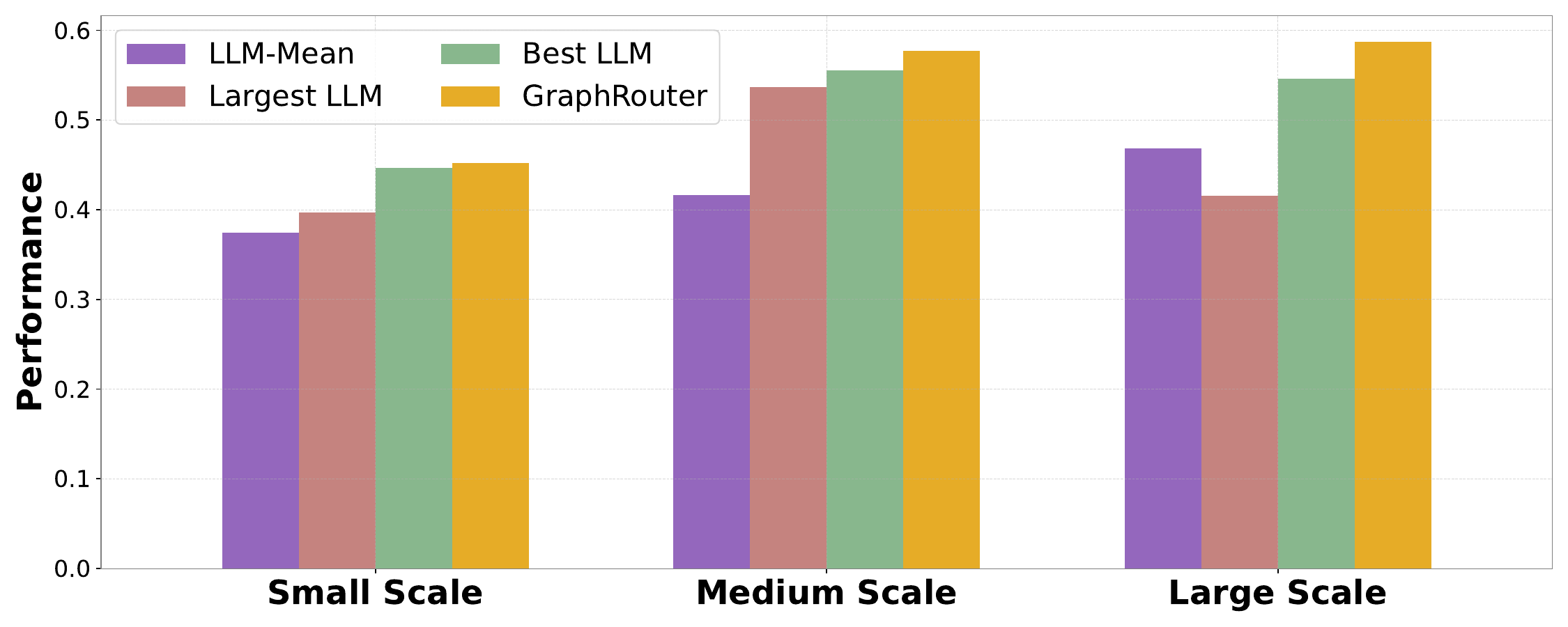}
    \caption{
    \textbf{GraphRouter fuses smaller LLMs to beat both top single and largest models.}
    Comparison across scales shows larger LLMs don’t always perform best under the 1024-token constraint.
    }
\label{fig:level-1 fig}
\end{figure}

\subsection{Result Analysis: Query-level Fusion}

We report results in Table~\ref{tab:query-level-results} with three key findings:
\textbf{(1) Query-level fusion outperforms the best single LLM across all four scenarios.}
As shown in Table~\ref{tab:query-level-results}, methods such as RouterMLP, RouterKNN, and RouterSVM consistently surpass the best individual LLM, with gains of at least 2\% and up to 16\% in performance or reward, depending on the scenario and metric.
\textbf{(2) GraphRouter achieves the strongest performance consistently.}
It brings over 10\% relative reward gains in the performance first and balance settings, 16\%+ in cost first, and a 5.7\% LLMScore improvement under the LLM Judge scenario. Other routers are less stable, sometimes even underperforming the Best LLM baseline, showing the challenge of effective multi-task multi-LLM fusion.
\textbf{(3) GraphRouter effectively fuses smaller LLMs to outperform both the best and largest single models.}
We further assess GraphRouter’s ability across LLM scales (Table~\ref{tab:2.2}). Under the performance-first setting, it consistently beats static baselines across small, medium, and large LLMs (Figure~\ref{fig:level-1 fig}), highlighting both the strength of the method and the versatility of \bench in LLM fusion.

\section{Thought-level \method Fusion} \label{sec:Thought-level fusion}

\subsection{Methodology: Thought-level Fusion}

At the thought level, the core insight is to summarize the responses of multiple LLMs for each query into a thought template, thereby enabling the fusion of different LLMs' capabilities to enhance few-shot responses. \revise{This level relies on the log fields that describe historical reasoning behavior, including historical query $q_i$, response mode $s$, generated response $r$, task-specific performance $y$, and LLM-judge score $j$. These fields allow \method to select informative reasoning traces from aligned multi-LLM responses and abstract them into reusable templates, rather than directly retrieving arbitrary raw responses.} Specifically, as shown in the second part on the right side of Figure \ref{fig:Overview method}, 
an LLM is employed to serve as a thought summarizer $L_s$, distilling a generalized reasoning template $T_i$ from the top-k responses $R_i$ of query $q_i$ selected according to performance or LLM judge:
\begin{equation}
\label{eq:topk_selection}
R_i = \{ r_1^{i}, r_2^{i}, \ldots, r_k^{i} \} = \text{Top-}k\left(\{r_j^i\}_{j=1}^{n} \text{Score}(r_j^i)\right), \hspace{5mm} T_i = L_s(R_i)
\end{equation}
where \( \{r_j^i\}_{j=1}^{n} \) denotes the set of responses generated by \(n\) different LLMs for query \(q^i\),  
and \(\text{Score}(r_j^i)\) is given by task-specific metrics or an LLM judge.  

Based on this, for new queries $q^l$, we retrieve the top-d similar queries $\mathcal{Q}^l=(q_1,q_2,...,q_d)$ via similarity search and use their corresponding thought
templates $\mathcal{T}^l=(T_1,T_2,...,T_d)$  to enable few-shot prompting of the LLM response $R^l$:
\begin{equation}
\label{eq:query_retrieval}
\mathcal{Q}^l = \text{Top-}d\left( \mathcal{Q}, \text{sim}(q^l, q_i) \right), \hspace{5mm} \mathcal{T}^l = (T_1, T_2, \ldots, T_d), \hspace{5mm} R^l = \text{LLM}\left(q^l, \mathcal{T}^l \right)
\end{equation}

\revise{Within the fusion-stage matrix, thought-level fusion occupies a distinct middle stage between routing and distillation. Unlike query-level fusion, it does not merely choose one model before generation. Unlike model-level fusion, it does not require fine-tuning a base model. Instead, it reuses high-quality reasoning traces from the logs as non-parametric guidance at inference time. This makes thought-level fusion useful when users want to exploit historical multi-LLM reasoning patterns while remaining compatible with API-based LLMs and avoiding offline model training.}

\subsection{Experimental Settings: Thought-level Fusion}
\label{sec:thoughtfusion}

\begin{figure}[t]
    \centering
    \vspace{-6mm}
    \includegraphics[width=\linewidth]{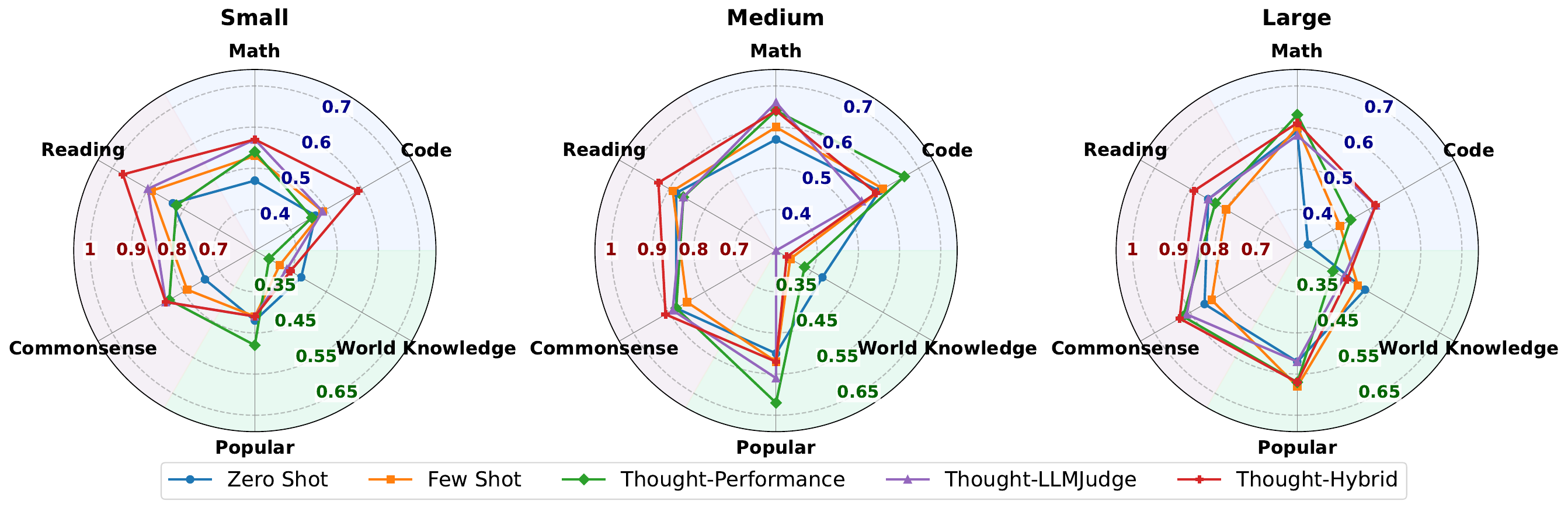}
    \caption{\textbf{Thought-level fusion improves performance across model scales and task domains.} Radar plots show accuracy per domain for small (LLaMA-3.1 8B), medium (LLaMA-3.1 Nemotron 51B), and large (Palmyra Creative 122B) models under different paradigms. The thought template approach yields the largest gains, especially on reasoning-heavy tasks like Math and Code. The hybrid strategy balances performance and response quality, often achieving the best results.}
\label{fig:5.1}
\end{figure}

\xhdr{Model and Implementation}
We evaluate thought-level fusion across three model sizes and six task domains in \bench. For each training query, \texttt{LLaMA-3 (70B)} generates a high-level \emph{thought template} by summarizing the top-3 responses, selected by \textit{Performance} and \textit{LLMScore}, with cost as tiebreaker (prompt in Table~\ref{table:thoughtTcreation}).
To assess effectiveness, we compare three paradigms: (1) \textit{Zero-Shot}, using only the query; (2) \textit{Few-Shot}, retrieving three similar queries (by cosine similarity) each with three top responses; (3) \textit{Thought Template}, using the same three similar queries, but with distilled templates.
For \textit{Thought Template}, we try three strategies: (a) \textit{Performance-based}, using top-3 responses by \textit{Performance}; (b) \textit{LLMScore-based}, by \textit{LLMScore}; (c) \textit{Hybrid}, selecting top-5 by \textit{Performance}, then top-3 by \textit{LLMScore}.
We experiment with \textit{llama3-8b} (small), \textit{llama-3.1-nemotron-51b-instruct} (medium), and \textit{palmyra-creative-122b} (large), chosen as top performers in their scale. Accuracy is averaged by domain and shown as radar plots in Figure~\ref{fig:5.1}.

\subsection{Result Analysis: Thought-level Fusion}

As shown in Figure~\ref{fig:5.1}, thought-level fusion consistently improves performance across all model sizes. Averaged over six task domains, small, medium, and large models improve by 12.7\%, 4.8\%, and 8.3\%, respectively, compared to zero-shot. 
These improvements manifest differently across reasoning types and model sizes, as detailed below.
\revise{\textbf{(1) Strong gains on reasoning-intensive tasks (relative to zero-shot).} Thought fusion is most effective on tasks like math and code. Measured against the zero-shot baseline, Hybrid fusion improves math by 21.3\% (small) and 15.8\% (medium); the large model (Palmyra Creative 122B) gains 57.6\% in coding (from 0.33 to 0.52). Commonsense and reading comprehension also improve (14.9\% and 16.9\% for small models). In contrast, performance drops on world knowledge (-7.9\% for small models) and varies on popular knowledge, likely due to these tasks emphasizing factual recall over reasoning. Relative to the stronger few-shot baseline (also shown in Figure~\ref{fig:5.1}), the best thought-fusion variant remains positive on math and code but the margin is smaller (within $\sim$5\% on several scale--strategy combinations); we therefore present thought-template fusion as a robust alternative to few-shot rather than a uniform improvement over it.}
\textbf{(2) Large models still benefit.} Thought-level fusion yields notable improvements even for large models: +57.6\% in code and +6.8\% in math. Medium models gain 11.9\% (code) and 15.8\% (math), showing the value of distilling diverse reasoning even at large model scales.
\revise{\textbf{(3) Hybrid strategy offers best trade-off.} Here, \textit{accuracy} refers to task-specific Performance (e.g., exact-match for QA, pass@1 for code) and \textit{quality} refers to the LLM-judge score (LLMScore) over $\langle$query, ground-truth, response$\rangle$ triples (Appendix~\ref{app:LLM-As-Judge_prompt}); Hybrid selection picks top-5 by Performance, then top-3 by LLMScore as tie-breaker (Section~\ref{sec:thoughtfusion}). Averaged across the six domains for small models, Hybrid achieves the largest mean gain over zero-shot (+12.7\%, vs Performance-based +3.6\% and LLMScore-based +7.3\%); at task granularity, however, no single strategy strictly dominates across all tasks in Figure~\ref{fig:5.1}, and the three are best understood as competitive alternatives whose relative ranking depends on the task. The corresponding overall (across-strategy) thought-fusion means for medium and large scales are +4.8\% and +8.3\% as reported in the opening paragraph of this section, with the per-strategy breakdown for those scales shown visually in Figure~\ref{fig:5.1}.}
See Appendix~\ref{app:inference_time_thought_level} and~\ref{app:reasoning_data_help} for analysis of inference time and benefits of reasoning-augmented responses in thought-level fusion.

\section{Model-level \method Fusion} \label{sec:Model-level fusion}

\subsection{Methodology: Model-level Fusion}

The model-level aims to leverage different LLMs' high-quality responses as training data to fuse  LLM capabilities. \revise{In this level, the structured logs provide supervised distillation data by pairing each query $q_i$ with quality-ranked responses $R_i$, where the ranking is determined by task-specific performance $y$ or LLM-judge score $j$. Thus, model-level fusion uses the feedback signals stored in the logs to select which responses should be internalized by the base model.} Specifically, as shown in the last part on the right side of Figure \ref{fig:Overview method}, we obtain the top-k responses $R_i$ of query $q_i$ selected according to performance or LLM judge as equation (\ref{eq:topk_selection}). Based on this, we can obtain a training data containing $n$ samples $Z=\{q_i,R_i\}_{i=1}^{n}$. Finally, we leverage $Z$ to train an LLM $f_{\theta}$ based on supervised fine-tuning:
\begin{equation}
\label{eq:loss_minimization}
\theta^{*} = \arg\min_{\theta} \frac{1}{n} \sum_{i=1}^{n} \mathcal{L}\left(f_{\theta}(q_i), R_i\right)
\end{equation}
Once  $\theta^{*}$ is obtained, the fine-tuned $f_{\theta^{*}}$  can generate responses for different queries.

\revise{In the fusion-stage matrix, model-level fusion represents late-stage parametric consolidation. Compared with query-level and thought-level fusion, it requires access to a trainable base model and introduces offline fine-tuning cost. However, once training is completed, inference only requires a single fused model, which can be desirable when users need deployment simplicity or cannot call multiple LLMs during inference. Studying this stage together with query- and thought-level fusion allows us to compare whether multi-LLM logs are more effectively used as routing supervision, prompting guidance, or training data.}

\subsection{Experimental Settings: Model-level Fusion}
\label{sec:model}

\xhdr{Model and Implementation}
To evaluate model-level fusion, we fine-tune the best-performing small LLM (LLaMA-3.1 (8B)) on \bench. For each query, we rank and select LLM-generated responses using three strategies from Section~\ref{sec:2.1}:
(1) \textbf{Task Performance}: top-$K$ by performance;
(2) \textbf{LLM Judge}: top-$K$ by judge score;
(3) \textbf{Hybrid}: first by performance, then tie-break with judge score. 
The selected responses from each strategy serve as training data.

Following knowledge distillation practices \citep{hinton2015distilling,xie2020self}, we augment training set by including each query’s ground truth along with the top-$K$ LLM responses. This forms the \textbf{Top-$K$ SFT} method. We compare it with two baselines:
\textbf{Zero-Shot:} directly uses LLaMA-3.1 (8B) without fine-tuning;
\textbf{Top-$K$ Label-only SFT:} trains only on repeated ground truths (replicated $K+1$ times to match training data volume).
This setup tests whether model-level fusion effectively distills multi-LLM knowledge beyond supervised fine-tuning on ground truth alone.

\begin{table}[t]
\centering
\vspace{-6mm}
\caption{
\textbf{Model-level fusion performance of LLaMA-3.1 (8B) on \bench under different fine-tuning and selection strategies.}
We compare Zero-Shot, Top-$K$ Label-only SFT, and Top-$K$ SFT across three response selection criteria: Task Performance, LLM Judge, and Hybrid. Results span six tasks, with \textbf{bold} and \underline{underline} indicating the best and second-best. CS, WK, and Read denote Commonsense, World Knowledge, and Reading Comprehension.
}
\setlength\tabcolsep{2pt}
\resizebox{1\textwidth}{!}{%
\begin{tabular}{lcccccccccccccccccc}
\toprule
\textbf{Sort Criterion} & \multicolumn{6}{c}{\textbf{Task Performance}} & \multicolumn{6}{c}{\textbf{LLM Judge}} & \multicolumn{6}{c}{\textbf{Hybrid}} \\
\cmidrule(r){1-1} \cmidrule(r){2-7} \cmidrule(r){8-13} \cmidrule(r){14-19}
\textbf{Scenarios} & Math & Code & CS & WK & Read & Popular & Math & Code & CS & WK & Read & Popular & Math & Code & CS & WK & Read & Popular \\
\midrule
Zero-Shot & 0.473 & \textbf{0.564} & 0.725 & \underline{0.300} & 0.703 & 0.415 & \underline{0.473} & \underline{0.564} & 0.725 & \textbf{0.300} & \textbf{0.703} & 0.415 & \underline{0.473} & \underline{0.564} & 0.725 & \textbf{0.300} & 0.703 & 0.415 \\
Top-5 Label-only SFT & 0.217 & 0.510 & \underline{0.785} & 0.150 & 0.646 & 0.372 & 0.217 & 0.510 & 0.785 & 0.150 & 0.646 & 0.372 & 0.217 & 0.510 & 0.785 & 0.150 & 0.646 & 0.372 \\
Top-5 SFT & \textbf{0.547} & 0.503 & 0.755 & 0.290 & \textbf{0.731} & \underline{0.468} & \textbf{0.528} & \textbf{0.584} & 0.780 & \underline{0.280} & 0.662 & \textbf{0.489} & 0.443 & \textbf{0.578} & 0.790 & \textbf{0.300} & \textbf{0.728} & 0.457 \\
Top-10 Label-only SFT & 0.340 & 0.498 & \textbf{0.830} & 0.150 & 0.667 & \textbf{0.489} & 0.340 & 0.498 & \textbf{0.830} & 0.150 & \underline{0.667} & \textbf{0.489} & 0.340 & 0.498 & \textbf{0.830} & 0.150 & 0.667 & \underline{0.489} \\
Top-10 SFT & \underline{0.509} & \underline{0.535} & 0.765 & \textbf{0.310} & \underline{0.707} & 0.425 & \textbf{0.528} & 0.515 & \underline{0.815} & \textbf{0.300} & 0.647 & \underline{0.468} & \textbf{0.491} & 0.526 & \underline{0.815} & \underline{0.250} & \underline{0.706} & \textbf{0.500} \\
\bottomrule
\end{tabular}}
\label{tab:comparison_results_model_fusion}
\end{table}

\subsection{Result Analysis: Model-level Fusion}

We summarize results in Table~\ref{tab:comparison_results_model_fusion} with three key observations:
\revise{\textbf{(1) Model-level fusion yields modest gains that are consistent across our reported configurations.}
Computing the cross-domain mean of Table~\ref{tab:comparison_results_model_fusion} for each Top-$K$ SFT configuration against the zero-shot baseline of $0.530$, every Top-$K$ SFT configuration improves the cross-domain mean by $+0.012$ to $+0.024$ (range: Performance-based $+0.012$ at $K{=}10$ to LLM-Judge $+0.024$ at $K{=}5$). Per-domain strict-win counts vary more: counting wins across all six Top-$K$ SFT configurations, model-level fusion beats zero-shot on \textbf{Commonsense (6/6)}, \textbf{Popular (6/6)}, \textbf{Math (5/6)}, and \textbf{Reading (4/6)}; the two domains where it usually loses are \textbf{Code (2/6)} and \textbf{World Knowledge (1/6, with 2 additional ties at 0.300)}. Top-$K$ Label-only SFT (rows 2 and 4 of Table~\ref{tab:comparison_results_model_fusion}) is reported as an ablation isolating the contribution of multi-LLM responses: at $K{=}5$, multi-LLM Top-$K$ SFT outperforms Label-only by $+0.102$ on the cross-domain mean ($0.549$ vs $0.447$); at $K{=}10$, by $+0.05$ ($0.546$--$0.548$ vs $0.496$). The gap is largest on World Knowledge, where Label-only collapses to $0.150$ at both $K$ values, below the zero-shot baseline of $0.300$. All model-level results are reported under a single SFT seed and a single response-sample draw; multi-seed variance estimation is left to future work.}
\revise{\textbf{(2) Fusion may hurt on code; we conjecture data scarcity and task mismatch as contributing factors.}
All fusion strategies perform worse on code. We conjecture that data scarcity (MBPP and HumanEval together provide the smallest training pool of any domain --- 494 samples / 6.97M tokens vs $\geq 900$ samples / $\geq 8.5$M tokens for every other domain, Table~\ref{app_tab:2}) and the divergence between code generation and QA-style supervision contribute to this result; verifying this hypothesis is left to future work.}
\revise{\textbf{(3) On the cross-domain mean, LLM-Judge or Hybrid selection dominates metric-based selection.}
Computing the cross-domain mean of Table~\ref{tab:comparison_results_model_fusion}, LLM-Judge selection achieves the highest mean at Top-$5$ SFT ($0.554$, driven primarily by Code), and Hybrid selection is best at Top-$10$ SFT ($0.548$); Performance-based selection is dominated by at least one alternative at both $K$ values. We therefore recommend LLM-Judge or Hybrid selection over rigid metric-based selection, particularly in domains such as Code where exact-match metrics undervalue informative responses.}

\begin{table*}[t]
    \centering
    \caption{\textbf{Comparison across three fusion levels and six domains.} Commonsense, World, and Read denote Commonsense Reasoning, World Knowledge, and Reading Comprehension. Average indicates mean task performance. \textbf{Bold} and \underline{underline} highlight the best and second-best.}
    \vspace{-2mm}
    \setlength\tabcolsep{3pt}
    \footnotesize
    \resizebox{\textwidth}{!}{
    \begin{tabular}{lcc|cccccc|c}
        \toprule
        \textbf{Fusion Level} 
        & \textbf{Fusion LLM Size} & \textbf{Fusion Criteria} 
        & \textbf{Math} & \textbf{Code} & \textbf{Commonsense} & \textbf{World} & \textbf{Read} & \textbf{Popular} & \textbf{Average} \\
        \midrule
        \revise{Best LLM} & \revise{Full} & \revise{N/A} & \revise{\textbf{0.612}} & \revise{0.545} & \revise{0.786} & \revise{\textbf{0.345}} & \revise{0.872} & \revise{0.441} & \revise{0.600} \\
        Zero-shot & Small & N/A & 0.602 & 0.517 & 0.715 & 0.334 & 0.834 & 0.398 & 0.567 \\
        \midrule
        \revise{Self-Consistency} & \revise{Full} & \revise{Response-level} & \revise{\underline{0.610}} & \revise{0.535} & \revise{0.758} & \revise{0.326} & \revise{0.884} & \revise{0.420} & \revise{0.589} \\
        \revise{LLM-Blender} & \revise{Full} & \revise{Response-level} & \revise{0.604} & \revise{0.548} & \revise{0.775} & \revise{0.332} & \revise{0.918} & \revise{0.438} & \revise{0.603} \\
        \revise{FuseLLM-adapted} & \revise{Full} & \revise{Model-level} & \revise{0.542} & \revise{0.455} & \revise{0.792} & \revise{0.286} & \revise{0.846} & \revise{0.418} & \revise{0.557} \\
        \midrule
        Query & Small & Performance & 0.599 & 0.522 & 0.792 & \revise{\underline{0.338}} & 0.862 & 0.395 & 0.585 \\
        \midrule
        Thought & Small & Hybrid-8b & \underline{0.610} & 0.423 & 0.778 & 0.301 & 0.956 & 0.443 & 0.585 \\
        Thought & Full & Hybrid-8b & 0.544 & \textbf{0.564} & \underline{0.799} & 0.304 & 0.965 & \revise{0.452} & 0.605 \\
        Thought & Small & Hybrid-70b & \textbf{0.612} & 0.550 & 0.778 & 0.312 & \underline{0.966} & 0.447 & \underline{0.611} \\
        Thought & Full & Hybrid-70b & 0.588 & \underline{0.562} & 0.796 & 0.303 & \textbf{0.970} & \textbf{0.473} & \textbf{0.615} \\
        \revise{Thought} & \revise{Full} & \revise{Hybrid-70b w/o near-dup} & \revise{0.576} & \revise{0.560} & \revise{0.774} & \revise{0.297} & \revise{0.958} & \revise{\underline{0.461}} & \revise{0.604} \\
        \midrule
        Model & Small & LLM Judge & 0.530 & 0.510 & 0.708 & 0.286 & 0.851 & 0.410 & 0.549 \\
        Model & Full & LLM Judge & 0.528 & 0.436 & \textbf{0.802} & 0.290 & 0.851 & 0.425 & 0.555 \\
        \bottomrule
    \end{tabular}
    }
    \label{tab:level_4_results}
\end{table*}

\section{Comparisons Across Three \method Fusion Levels} \label{sec:cross}

In this section, we compare the three \method levels on the full test set of 14 tasks (Table~\ref{tab:level_4_results}), selecting the best-performing configuration for each level:
Query-level: GraphRouter using task performance as the routing criterion;
Thought-level: Hybrid strategy with thought summarization;
Model-level: Fine-tuning with top-$K$ responses selected by the LLM Judge.
\revise{
For the original cross-level comparison, we use a unified generation setting to compare query-, thought-, and model-level fusion. In the revised table, we further add the strongest individual LLM among all 20 models, representative response/model-level fusion baselines, and a near-duplicate-controlled thought-level setting to provide a more complete comparison.
}
Table~\ref{tab:level_4_results} further breaks down fusion settings by two factors: Fusion LLM Size (whether responses are fused from all or only small models) and Fusion Criteria (how responses are selected, \textit{e.g.}, task performance vs. LLMScore).

\revise{
This cross-level comparison is central to the motivation of our fusion-stage matrix. Existing routing and distillation methods often study a single stage of the LLM pipeline in isolation, such as query-level model selection or model-level capability transfer. In contrast, our comparison asks how the same structured multi-LLM log data should be used across different intervention points: before generation for routing, during generation for thought-level reasoning augmentation, or after response collection for supervised fine-tuning. Therefore, Table~\ref{tab:level_4_results} is not merely an engineering comparison, but a decision-oriented matrix for understanding the performance, cost, and generalization trade-offs among different ways of exploiting multi-LLM logs.
}

\revise{
Consistent with the empirical nature of this work, our goal here is not to derive formal theoretical guarantees or closed-form complexity bounds. Instead, we use the unified benchmark setting to provide an empirical trade-off analysis of the three fusion stages. Query-level fusion tests whether logs are best used as routing supervision, thought-level fusion tests whether logs are best used as reusable reasoning guidance, and model-level fusion tests whether logs are best used as distillation data. This comparison helps characterize the practical behavior of different fusion stages under the same data source and evaluation protocol.
}

Using unified experimental settings, we now summarize and compare the results across all three fusion levels.
\textbf{(1) Thought-level fusion achieves the best overall performance.}
Thought-Full-Hybrid-70B outperforms all settings, demonstrating its ability to integrate reasoning from diverse LLMs. However, performance gains plateau when increasing model size or response count, suggesting diminishing returns relative to cost.
\revise{
After adding the strongest individual LLM and representative fusion baselines, Thought-Full-Hybrid-70B still achieves the best average performance, outperforming Best LLM (0.615 vs. 0.600), Self-Consistency (0.615 vs. 0.589), LLM-Blender (0.615 vs. 0.603), and FuseLLM-adapted (0.615 vs. 0.557). This indicates that abstract thought-template retrieval provides benefits beyond selecting the strongest single model or aggregating final responses.
}
\revise{
This result suggests that multi-LLM logs are especially useful when their high-quality reasoning traces are reused as non-parametric guidance at inference time. Compared with pure routing, thought-level fusion can expose the responding model to diverse reasoning patterns rather than selecting only one model. Compared with model-level fusion, it avoids compressing heterogeneous task behaviors into a single set of parameters. Empirically, this gives thought-level fusion a favorable balance between flexibility and effectiveness, especially for tasks where reusable reasoning patterns are more valuable than simply selecting a model.
}
\revise{
The near-duplicate-controlled setting further supports this interpretation. After removing retrieved training queries with embedding similarity above 0.9, thought-level fusion still achieves 0.604 average performance, remaining above zero-shot (0.567) and close to the original thought-level result (0.615). This suggests that the gain is not solely driven by retrieving near-duplicate training queries, but also reflects reusable reasoning patterns captured by the thought templates.
}

\revise{\textbf{(2) Model-level fusion underperforms despite access to high-quality responses.}
Fine-tuned models outperform zero-shot in only 3 of 6 domains, and even fall behind on average. We hypothesize two contributing factors --- overfitting to multi-response inputs and difficulty abstracting across heterogeneous task styles during SFT --- and view rigorous verification (e.g., per-task SFT ablations) as future work.}
\revise{
This observation highlights an important limitation of treating multi-LLM logs purely as distillation data. Although model-level fusion has deployment advantages after training, the offline consolidation process may introduce training bias, reduce adaptability, or overfit to the particular response styles selected from the logs. In an empirical bias-generalization interpretation, model-level fusion can reduce inference-time variability by using a single fine-tuned model, but it may increase training-induced bias when heterogeneous responses from different tasks and LLMs are compressed into one base model. This explains why API-compatible distillation alone does not fully address the broader question of how multi-LLM logs should be used.
}
\revise{
The adapted FuseLLM baseline shows a similar trend, achieving 0.557 average performance, which is close to our model-level fusion result but still below thought-level fusion. This further suggests that directly compressing heterogeneous multi-LLM logs into model parameters may be less effective than using them as retrieval-time reasoning guidance.
}

\revise{\textbf{(3) Query-level fusion achieves moderate accuracy with the lowest cost.}
Query-level fusion via GraphRouter avoids running the backbone on most queries and therefore achieves the lowest per-query cost: under cost-aware routing, GraphRouter cost is $0.0184$ vs $0.1850$ for Best-LLM under Performance First, a roughly $10\times$ reduction (Table~\ref{tab:query-level-results}). Thought- and model-level fusion in this section's unified setting incur full LLaMA-3 (8B) inference cost per query, computed as the LLaMA-3 (8B) per-token price (Table~\ref{tab:llm-costs}) applied to the per-query token counts available in \bench (Table~\ref{tab:data_collection_info}). Thought-level fusion additionally incurs $\sim$5\% latency overhead and $0.03$s retrieval overhead over zero-shot (Appendix~\ref{app:inference_time_thought_level}, Table~\ref{tab:inference_time_thought}). The cost ordering across fusion levels follows directly from this routing-vs-full-inference distinction.}
\revise{
This result confirms the value of query-level fusion for cost-sensitive API-serving scenarios. Since query-level fusion only requires a lightweight router and calls one selected LLM at inference time, it provides a practical option when users prioritize efficiency. However, its performance ceiling can be limited because it exploits multi-LLM logs only through model selection, without reusing richer reasoning traces or consolidating capabilities into parameters. Empirically, query-level fusion therefore has a low-overhead profile, but may be biased toward the capability boundary of the selected LLM.
}
\revise{
Compared with response-level baselines, query-level fusion achieves lower average performance than LLM-Blender (0.585 vs. 0.603), but it requires only one selected model call rather than aggregating multiple candidate responses. This highlights its distinct role as an efficiency-oriented fusion strategy rather than a purely performance-oriented one.
}

\revise{\textbf{(4) Fusion provides only marginal improvements on World Knowledge and Math.}
On these domains, the best fusion result exceeds zero-shot by at most 2 points absolute (Table~\ref{tab:level_4_results}: Math thought-level $0.612$ vs zero-shot $0.602$; World Knowledge query-level $0.338$ vs zero-shot $0.334$), suggesting these domains --- which depend heavily on factual recall and strict logical sequencing --- are less amenable to multi-LLM fusion under our current framework.}
\revise{
These domain-level results further support the need for a fusion-stage matrix. Different tasks benefit from different ways of using multi-LLM logs. Knowledge-intensive tasks may require strict factual grounding and are sensitive to noisy fused information, while mathematical tasks require coherent step-by-step reasoning and may be harmed by inconsistent reasoning traces. Therefore, no single fusion level uniformly dominates across all settings, and a systematic comparison is needed to determine which fusion stage is appropriate under different domain requirements.
}

\revise{
Taken together, these observations provide an empirical interpretation of the trade-offs among the three fusion levels. Query-level fusion is efficient and easy to deploy, but it mainly benefits from better model selection. Thought-level fusion is more flexible because it reuses reasoning traces without modifying model parameters, but it introduces retrieval and prompt-token overhead. Model-level fusion is attractive when a single deployable model is desired, but it requires offline training and may generalize poorly when the logs contain heterogeneous task patterns. The additional comparisons with Best LLM, response-level fusion baselines, and near-duplicate-controlled retrieval further show that the gains of thought-level fusion are not simply due to a stronger single model, final-response aggregation, or near-duplicate retrieval. This empirical trade-off analysis complements the stage-aware taxonomy in Section~\ref{sec:method}.
}

\revise{Overall, these findings validate our central claim: multi-LLM log data, when systematically structured by query / task / response triples and ranked by performance and LLM-judge scores, supports fusion strategies that outperform single-LLM baselines. On the full test set (Table~\ref{tab:level_4_results}), thought-level fusion improves over zero-shot by $4.8$ points on average across the six domains ($0.567 \rightarrow 0.615$); under cost-aware routing, query-level fusion matches the strongest single LLM at an order of magnitude lower cost (Table~\ref{tab:query-level-results}: GraphRouter $0.602$ reward / $0.0184$ cost vs Best-LLM $0.546$ / $0.1850$ under Performance First). The framework supports both local and API-served LLMs and applies across all three fusion stages, and \bench provides a demanding benchmark for real-world multi-LLM integration.}
\revise{
More importantly, the comparison demonstrates the unique value of our work beyond existing API-only routing or distillation methods. The main contribution is not simply that a method can operate with API-accessible models, but that the same multi-LLM log source can be systematically evaluated across query-level, thought-level, and model-level fusion. This matrix reveals when logs are better used as routing supervision, when they are better used as reasoning templates, and when they are less effective as fine-tuning data. Such cross-stage evidence would be difficult to obtain from studies that focus only on routing, response-level fusion, or distillation.
}

\section{Additional Related Work}

\xhdr{LLM fusion}
Recent work explores fusing multiple LLMs or reasoning traces to enhance performance, robustness, and generalization.
Query-level methods (\textit{e.g.}, LLMRouter \citep{li2023llmrouter}, STaR \citep{zhou2022star}) treat routing as policy learning to select the best model per query.
Thought-level methods (\textit{e.g.}, Thought Propagation \citep{li2023thoughtprop}, Tree of Thoughts \citep{yao2023tree}) aggregate reasoning paths for more consistent answers.
Model-level methods (\textit{e.g.}, FrugalGPT \citep{jiang2023frugalgpt}, Self-Consistency \citep{wang2022selfconsistency}) combine multiple model outputs or samples to approximate stronger reasoning while controlling cost.
LLM-as-judge approaches (\textit{e.g.}, LLM-as-Judge \citep{zhang2023llmjudge}, RAG Fusion \citep{lewis2021ragfusion}) evaluate and select responses via a secondary LLM.
However, most prior studies focus on a single level of fusion. We instead introduce \bench, a large-scale benchmark for LLM fusion, and conduct a systematic comparison across three levels via \method, enabling more holistic and integrated insights.


\xhdr{LLM Router}
Recent work on LLM routing focuses on improving efficiency and response quality by directing queries to the most suitable model.
Early methods like RouteLLM \citep{ding2024routellm} optimize preference-aligned model selection via learned policies, while RouterBench \citep{hu2024routerbench} benchmarks routing decisions but does not study routing logs for capability fusion.
TO-Router \citep{stripelis2024torouter} and Expert Router \citep{anonymous2024expert} emphasize parallel dispatch for efficiency, and Glider \citep{li2024glider} integrates global and local signals to refine expert selection.
In contrast, our work focuses on capability-oriented analysis. We leverage large-scale multi-LLM log data not only for model selection, but as a form of weak supervision to systematically study and fuse LLM capabilities across three flexible levels.


\section{Conclusion}



In this work, we revisit LLM fusion through the lens of API-based multi-LLM log data, arguing that practical fusion must (1) be compatible with real-world serving (local and API) and (2) support integration at different pipeline stages. We introduce \bench, a large-scale benchmark spanning 14 tasks and 20 LLMs with diverse response modes and reusable thought templates, and propose \method, a three-level framework for stage-aware fusion. Experiments show that thought-level fusion delivers the strongest gains, query-level fusion offers the best accuracy–cost trade-off, and model-level fusion lags due to generalization limits. \revise{Overall, these results validate multi-LLM log data as a practical supervision source and position \bench as a demanding, real-world testbed for flexible, serving-compatible LLM fusion methods. \method is model-agnostic by construction: extending the LLM pool to newer frontier models (e.g., GPT-5, Claude 4, Gemini 2) requires only re-running response collection on the existing query set, with no architectural change to the fusion pipeline.}



\bibliography{Ref}

@article{shnitzer2023large,
  title={Large language model routing with benchmark datasets},
  author={Shnitzer, Tal and Ou, Anthony and Silva, M{\'\i}rian and Soule, Kate and Sun, Yuekai and Solomon, Justin and Thompson, Neil and Yurochkin, Mikhail},
  journal={arXiv preprint arXiv:2309.15789},
  year={2023}
}

@article{ding2024hybrid,
  title={Hybrid LLM: Cost-efficient and quality-aware query routing},
  author={Ding, Dujian and Mallick, Ankur and Wang, Chi and Sim, Robert and Mukherjee, Subhabrata and Ruhle, Victor and Lakshmanan, Laks VS and Awadallah, Ahmed Hassan},
  journal={arXiv preprint arXiv:2404.14618},
  year={2024}
}

@article{chen2023frugalgpt,
  title={Frugalgpt: How to use large language models while reducing cost and improving performance},
  author={Chen, Lingjiao and Zaharia, Matei and Zou, James},
  journal={arXiv preprint arXiv:2305.05176},
  year={2023}
}

@article{dai2024cost,
  title={Cost-Effective Online Multi-LLM Selection with Versatile Reward Models},
  author={Dai, Xiangxiang and Li, Jin and Liu, Xutong and Yu, Anqi and Lui, John},
  journal={arXiv preprint arXiv:2405.16587},
  year={2024}
}

@article{ahmed2024studying,
  title={Studying LLM Performance on Closed-and Open-source Data},
  author={Ahmed, Toufique and Bird, Christian and Devanbu, Premkumar and Chakraborty, Saikat},
  journal={arXiv preprint arXiv:2402.15100},
  year={2024}
}

@article{zhang2024pybench,
  title={PyBench: Evaluating LLM Agent on various real-world coding tasks},
  author={Zhang, Yaolun and Pan, Yinxu and Wang, Yudong and Cai, Jie and Zheng, Zhi and Zeng, Guoyang and Liu, Zhiyuan},
  journal={arXiv preprint arXiv:2407.16732},
  year={2024}
}

@String { aaai        = {AAAI Conference on Artificial Intelligence (AAAI)} }

@String { iclr        = {International Conference on Learning Representations (ICLR)} }

@article{wei2022chain,
  title={Chain-of-thought prompting elicits reasoning in large language models},
  author={Wei, Jason and Wang, Xuezhi and Schuurmans, Dale and Bosma, Maarten and Xia, Fei and Chi, Ed and Le, Quoc V and Zhou, Denny and others},
  journal={Advances in neural information processing systems},
  volume={35},
  pages={24824--24837},
  year={2022}
}

@article{jiang2024mixtral,
  title={Mixtral of experts},
  author={Jiang, Albert Q and Sablayrolles, Alexandre and Roux, Antoine and Mensch, Arthur and Savary, Blanche and Bamford, Chris and Chaplot, Devendra Singh and Casas, Diego de las and Hanna, Emma Bou and Bressand, Florian and others},
  journal={arXiv preprint arXiv:2401.04088},
  year={2024}
}

@article{singhal2022large,
  title={Large language models encode clinical knowledge},
  author={Singhal, Karan and Azizi, Shekoofeh and Tu, Tao and Mahdavi, S Sara and Wei, Jason and Chung, Hyung Won and Scales, Nathan and Tanwani, Ajay and Cole-Lewis, Heather and Pfohl, Stephen and others},
  journal={arXiv preprint arXiv:2212.13138},
  year={2022}
}

@article{luo2022biogpt,
  title={BioGPT: generative pre-trained transformer for biomedical text generation and mining},
  author={Luo, Renqian and Sun, Liai and Xia, Yingce and Qin, Tao and Zhang, Sheng and Poon, Hoifung and Liu, Tie-Yan},
  journal={Briefings in bioinformatics},
  volume={23},
  number={6},
  pages={bbac409},
  year={2022},
  publisher={Oxford University Press}
}

@article{feng2024graphrouter,
  title={Graphrouter: A graph-based router for llm selections},
  author={Feng, Tao and Shen, Yanzhen and You, Jiaxuan},
  journal={arXiv preprint arXiv:2410.03834},
  year={2024}
}

@inproceedings{ong2024routellm,
  title={RouteLLM: Learning to Route LLMs from Preference Data},
  author={Ong, Isaac and Almahairi, Amjad and Wu, Vincent and Chiang, Wei-Lin and Wu, Tianhao and Gonzalez, Joseph E and Kadous, M Waleed and Stoica, Ion},
  booktitle={The Thirteenth International Conference on Learning Representations},
  year={2024}
}

@article{hu2024routerbench,
  title={Routerbench: A benchmark for multi-llm routing system},
  author={Hu, Qitian Jason and Bieker, Jacob and Li, Xiuyu and Jiang, Nan and Keigwin, Benjamin and Ranganath, Gaurav and Keutzer, Kurt and Upadhyay, Shriyash Kaustubh},
  journal={arXiv preprint arXiv:2403.12031},
  year={2024}
}

@article{guo2025deepseek,
  title={Deepseek-r1: Incentivizing reasoning capability in llms via reinforcement learning},
  author={Guo, Daya and Yang, Dejian and Zhang, Haowei and Song, Junxiao and Zhang, Ruoyu and Xu, Runxin and Zhu, Qihao and Ma, Shirong and Wang, Peiyi and Bi, Xiao and others},
  journal={arXiv preprint arXiv:2501.12948},
  year={2025}
}

@online{anthropic2025claude37,
  author = {{Anthropic}},
  title = {Claude 3.7 Sonnet and Claude Code},
  year = {2025},
  month = {2},
  url = {https://www.anthropic.com/news/claude-3-7-sonnet},
  urldate = {2025-04-30},
  organization = {Anthropic}
}

@article{talebirad2023multi,
  title={Multi-agent collaboration: Harnessing the power of intelligent llm agents},
  author={Talebirad, Yashar and Nadiri, Amirhossein},
  journal={arXiv preprint arXiv:2306.03314},
  year={2023}
}

@article{han2024llm,
  title={LLM multi-agent systems: Challenges and open problems},
  author={Han, Shanshan and Zhang, Qifan and Yao, Yuhang and Jin, Weizhao and Xu, Zhaozhuo and He, Chaoyang},
  journal={arXiv preprint arXiv:2402.03578},
  year={2024}
}

@article{chen2023agentverse,
  title={Agentverse: Facilitating multi-agent collaboration and exploring emergent behaviors in agents},
  author={Chen, Weize and Su, Yusheng and Zuo, Jingwei and Yang, Cheng and Yuan, Chenfei and Qian, Chen and Chan, Chi-Min and Qin, Yujia and Lu, Yaxi and Xie, Ruobing and others},
  journal={arXiv preprint arXiv:2308.10848},
  volume={2},
  number={4},
  pages={6},
  year={2023}
}

@article{haber2013model,
  title={Model fusion and joint inversion},
  author={Haber, Eldad and Holtzman Gazit, Michal},
  journal={Surveys in Geophysics},
  volume={34},
  pages={675--695},
  year={2013},
  publisher={Springer}
}

@article{li2023deep,
  title={Deep model fusion: A survey},
  author={Li, Weishi and Peng, Yong and Zhang, Miao and Ding, Liang and Hu, Han and Shen, Li},
  journal={arXiv preprint arXiv:2309.15698},
  year={2023}
}

@article{yuan2024advancing,
  title={Advancing llm reasoning generalists with preference trees},
  author={Yuan, Lifan and Cui, Ganqu and Wang, Hanbin and Ding, Ning and Wang, Xingyao and Deng, Jia and Shan, Boji and Chen, Huimin and Xie, Ruobing and Lin, Yankai and others},
  journal={arXiv preprint arXiv:2404.02078},
  year={2024}
}

@inproceedings{besta2024graph,
  title={Graph of thoughts: Solving elaborate problems with large language models},
  author={Besta, Maciej and Blach, Nils and Kubicek, Ales and Gerstenberger, Robert and Podstawski, Michal and Gianinazzi, Lukas and Gajda, Joanna and Lehmann, Tomasz and Niewiadomski, Hubert and Nyczyk, Piotr and others},
  booktitle={Proceedings of the AAAI Conference on Artificial Intelligence},
  volume={38},
  number={16},
  pages={17682--17690},
  year={2024}
}

@article{han2024context,
  title={In-Context Learning May Not Elicit Trustworthy Reasoning: A-Not-B Errors in Pretrained Language Models},
  author={Han, Pengrui and Song, Peiyang and Yu, Haofei and You, Jiaxuan},
  journal={arXiv preprint arXiv:2409.15454},
  year={2024}
}

@article{feng2024thought,
  title={Thought-retriever: Don’t just retrieve raw data, retrieve thoughts},
  author={Feng, Tao and Han, Pengrui and Lin, Guanyu and Liu, Ge and You, Jiaxuan},
  journal={ICLR 2024 How Far Are We From AGI},
  year={2024}
}

@article{yang2024buffer,
  title={Buffer of thoughts: Thought-augmented reasoning with large language models},
  author={Yang, Ling and Yu, Zhaochen and Zhang, Tianjun and Cao, Shiyi and Xu, Minkai and Zhang, Wentao and Gonzalez, Joseph E and Cui, Bin},
  journal={Advances in Neural Information Processing Systems},
  volume={37},
  pages={113519--113544},
  year={2024}
}

@article{liu2024chatqa,
  title={ChatQA: Surpassing GPT-4 on Conversational QA and RAG},
  author={Liu, Zihan and Ping, Wei and Roy, Rajarshi and Xu, Peng and Lee, Chankyu and Shoeybi, Mohammad and Catanzaro, Bryan},
  journal={arXiv preprint arXiv:2401.10225},
  year={2024}}

@misc{grattafiori2024llama3herdmodels,
      title={The Llama 3 Herd of Models}, 
      author={Aaron Grattafiori and Abhimanyu Dubey and Abhinav Jauhri and Abhinav Pandey and Abhishek Kadian and Ahmad Al-Dahle and Aiesha Letman and Akhil Mathur and Alan Schelten and Alex Vaughan and Amy Yang and Angela Fan and Anirudh Goyal and Anthony Hartshorn and Aobo Yang and Archi Mitra and Archie Sravankumar and Artem Korenev and Arthur Hinsvark and Arun Rao and Aston Zhang and Aurelien Rodriguez and Austen Gregerson and Ava Spataru and Baptiste Roziere and Bethany Biron and Binh Tang and Bobbie Chern and Charlotte Caucheteux and Chaya Nayak and Chloe Bi and Chris Marra and Chris McConnell and Christian Keller and Christophe Touret and Chunyang Wu and Corinne Wong and Cristian Canton Ferrer and Cyrus Nikolaidis and Damien Allonsius and Daniel Song and Danielle Pintz and Danny Livshits and Danny Wyatt and David Esiobu and Dhruv Choudhary and Dhruv Mahajan and Diego Garcia-Olano and Diego Perino and Dieuwke Hupkes and Egor Lakomkin and Ehab AlBadawy and Elina Lobanova and Emily Dinan and Eric Michael Smith and Filip Radenovic and Francisco Guzmán and Frank Zhang and Gabriel Synnaeve and Gabrielle Lee and Georgia Lewis Anderson and Govind Thattai and Graeme Nail and Gregoire Mialon and Guan Pang and Guillem Cucurell and Hailey Nguyen and Hannah Korevaar and Hu Xu and Hugo Touvron and Iliyan Zarov and Imanol Arrieta Ibarra and Isabel Kloumann and Ishan Misra and Ivan Evtimov and Jack Zhang and Jade Copet and Jaewon Lee and Jan Geffert and Jana Vranes and Jason Park and Jay Mahadeokar and Jeet Shah and Jelmer van der Linde and Jennifer Billock and Jenny Hong and Jenya Lee and Jeremy Fu and Jianfeng Chi and Jianyu Huang and Jiawen Liu and Jie Wang and Jiecao Yu and Joanna Bitton and Joe Spisak and Jongsoo Park and Joseph Rocca and Joshua Johnstun and Joshua Saxe and Junteng Jia and Kalyan Vasuden Alwala and Karthik Prasad and Kartikeya Upasani and Kate Plawiak and Ke Li and Kenneth Heafield and Kevin Stone and Khalid El-Arini and Krithika Iyer and Kshitiz Malik and Kuenley Chiu and Kunal Bhalla and Kushal Lakhotia and Lauren Rantala-Yeary and Laurens van der Maaten and Lawrence Chen and Liang Tan and Liz Jenkins and Louis Martin and Lovish Madaan and Lubo Malo and Lukas Blecher and Lukas Landzaat and Luke de Oliveira and Madeline Muzzi and Mahesh Pasupuleti and Mannat Singh and Manohar Paluri and Marcin Kardas and Maria Tsimpoukelli and Mathew Oldham and Mathieu Rita and Maya Pavlova and Melanie Kambadur and Mike Lewis and Min Si and Mitesh Kumar Singh and Mona Hassan and Naman Goyal and Narjes Torabi and Nikolay Bashlykov and Nikolay Bogoychev and Niladri Chatterji and Ning Zhang and Olivier Duchenne and Onur Çelebi and Patrick Alrassy and Pengchuan Zhang and Pengwei Li and Petar Vasic and Peter Weng and Prajjwal Bhargava and Pratik Dubal and Praveen Krishnan and Punit Singh Koura and Puxin Xu and Qing He and Qingxiao Dong and Ragavan Srinivasan and Raj Ganapathy and Ramon Calderer and Ricardo Silveira Cabral and Robert Stojnic and Roberta Raileanu and Rohan Maheswari and Rohit Girdhar and Rohit Patel and Romain Sauvestre and Ronnie Polidoro and Roshan Sumbaly and Ross Taylor and Ruan Silva and Rui Hou and Rui Wang and Saghar Hosseini and Sahana Chennabasappa and Sanjay Singh and Sean Bell and Seohyun Sonia Kim and Sergey Edunov and Shaoliang Nie and Sharan Narang and Sharath Raparthy and Sheng Shen and Shengye Wan and Shruti Bhosale and Shun Zhang and Simon Vandenhende and Soumya Batra and Spencer Whitman and Sten Sootla and Stephane Collot and Suchin Gururangan and Sydney Borodinsky and Tamar Herman and Tara Fowler and Tarek Sheasha and Thomas Georgiou and Thomas Scialom and Tobias Speckbacher and Todor Mihaylov and Tong Xiao and Ujjwal Karn and Vedanuj Goswami and Vibhor Gupta and Vignesh Ramanathan and Viktor Kerkez and Vincent Gonguet and Virginie Do and Vish Vogeti and Vítor Albiero and Vladan Petrovic and Weiwei Chu and Wenhan Xiong and Wenyin Fu and Whitney Meers and Xavier Martinet and Xiaodong Wang and Xiaofang Wang and Xiaoqing Ellen Tan and Xide Xia and Xinfeng Xie and Xuchao Jia and Xuewei Wang and Yaelle Goldschlag and Yashesh Gaur and Yasmine Babaei and Yi Wen and Yiwen Song and Yuchen Zhang and Yue Li and Yuning Mao and Zacharie Delpierre Coudert and Zheng Yan and Zhengxing Chen and Zoe Papakipos and Aaditya Singh and Aayushi Srivastava and Abha Jain and Adam Kelsey and Adam Shajnfeld and Adithya Gangidi and Adolfo Victoria and Ahuva Goldstand and Ajay Menon and Ajay Sharma and Alex Boesenberg and Alexei Baevski and Allie Feinstein and Amanda Kallet and Amit Sangani and Amos Teo and Anam Yunus and Andrei Lupu and Andres Alvarado and Andrew Caples and Andrew Gu and Andrew Ho and Andrew Poulton and Andrew Ryan and Ankit Ramchandani and Annie Dong and Annie Franco and Anuj Goyal and Aparajita Saraf and Arkabandhu Chowdhury and Ashley Gabriel and Ashwin Bharambe and Assaf Eisenman and Azadeh Yazdan and Beau James and Ben Maurer and Benjamin Leonhardi and Bernie Huang and Beth Loyd and Beto De Paola and Bhargavi Paranjape and Bing Liu and Bo Wu and Boyu Ni and Braden Hancock and Bram Wasti and Brandon Spence and Brani Stojkovic and Brian Gamido and Britt Montalvo and Carl Parker and Carly Burton and Catalina Mejia and Ce Liu and Changhan Wang and Changkyu Kim and Chao Zhou and Chester Hu and Ching-Hsiang Chu and Chris Cai and Chris Tindal and Christoph Feichtenhofer and Cynthia Gao and Damon Civin and Dana Beaty and Daniel Kreymer and Daniel Li and David Adkins and David Xu and Davide Testuggine and Delia David and Devi Parikh and Diana Liskovich and Didem Foss and Dingkang Wang and Duc Le and Dustin Holland and Edward Dowling and Eissa Jamil and Elaine Montgomery and Eleonora Presani and Emily Hahn and Emily Wood and Eric-Tuan Le and Erik Brinkman and Esteban Arcaute and Evan Dunbar and Evan Smothers and Fei Sun and Felix Kreuk and Feng Tian and Filippos Kokkinos and Firat Ozgenel and Francesco Caggioni and Frank Kanayet and Frank Seide and Gabriela Medina Florez and Gabriella Schwarz and Gada Badeer and Georgia Swee and Gil Halpern and Grant Herman and Grigory Sizov and Guangyi and Zhang and Guna Lakshminarayanan and Hakan Inan and Hamid Shojanazeri and Han Zou and Hannah Wang and Hanwen Zha and Haroun Habeeb and Harrison Rudolph and Helen Suk and Henry Aspegren and Hunter Goldman and Hongyuan Zhan and Ibrahim Damlaj and Igor Molybog and Igor Tufanov and Ilias Leontiadis and Irina-Elena Veliche and Itai Gat and Jake Weissman and James Geboski and James Kohli and Janice Lam and Japhet Asher and Jean-Baptiste Gaya and Jeff Marcus and Jeff Tang and Jennifer Chan and Jenny Zhen and Jeremy Reizenstein and Jeremy Teboul and Jessica Zhong and Jian Jin and Jingyi Yang and Joe Cummings and Jon Carvill and Jon Shepard and Jonathan McPhie and Jonathan Torres and Josh Ginsburg and Junjie Wang and Kai Wu and Kam Hou U and Karan Saxena and Kartikay Khandelwal and Katayoun Zand and Kathy Matosich and Kaushik Veeraraghavan and Kelly Michelena and Keqian Li and Kiran Jagadeesh and Kun Huang and Kunal Chawla and Kyle Huang and Lailin Chen and Lakshya Garg and Lavender A and Leandro Silva and Lee Bell and Lei Zhang and Liangpeng Guo and Licheng Yu and Liron Moshkovich and Luca Wehrstedt and Madian Khabsa and Manav Avalani and Manish Bhatt and Martynas Mankus and Matan Hasson and Matthew Lennie and Matthias Reso and Maxim Groshev and Maxim Naumov and Maya Lathi and Meghan Keneally and Miao Liu and Michael L. Seltzer and Michal Valko and Michelle Restrepo and Mihir Patel and Mik Vyatskov and Mikayel Samvelyan and Mike Clark and Mike Macey and Mike Wang and Miquel Jubert Hermoso and Mo Metanat and Mohammad Rastegari and Munish Bansal and Nandhini Santhanam and Natascha Parks and Natasha White and Navyata Bawa and Nayan Singhal and Nick Egebo and Nicolas Usunier and Nikhil Mehta and Nikolay Pavlovich Laptev and Ning Dong and Norman Cheng and Oleg Chernoguz and Olivia Hart and Omkar Salpekar and Ozlem Kalinli and Parkin Kent and Parth Parekh and Paul Saab and Pavan Balaji and Pedro Rittner and Philip Bontrager and Pierre Roux and Piotr Dollar and Polina Zvyagina and Prashant Ratanchandani and Pritish Yuvraj and Qian Liang and Rachad Alao and Rachel Rodriguez and Rafi Ayub and Raghotham Murthy and Raghu Nayani and Rahul Mitra and Rangaprabhu Parthasarathy and Raymond Li and Rebekkah Hogan and Robin Battey and Rocky Wang and Russ Howes and Ruty Rinott and Sachin Mehta and Sachin Siby and Sai Jayesh Bondu and Samyak Datta and Sara Chugh and Sara Hunt and Sargun Dhillon and Sasha Sidorov and Satadru Pan and Saurabh Mahajan and Saurabh Verma and Seiji Yamamoto and Sharadh Ramaswamy and Shaun Lindsay and Shaun Lindsay and Sheng Feng and Shenghao Lin and Shengxin Cindy Zha and Shishir Patil and Shiva Shankar and Shuqiang Zhang and Shuqiang Zhang and Sinong Wang and Sneha Agarwal and Soji Sajuyigbe and Soumith Chintala and Stephanie Max and Stephen Chen and Steve Kehoe and Steve Satterfield and Sudarshan Govindaprasad and Sumit Gupta and Summer Deng and Sungmin Cho and Sunny Virk and Suraj Subramanian and Sy Choudhury and Sydney Goldman and Tal Remez and Tamar Glaser and Tamara Best and Thilo Koehler and Thomas Robinson and Tianhe Li and Tianjun Zhang and Tim Matthews and Timothy Chou and Tzook Shaked and Varun Vontimitta and Victoria Ajayi and Victoria Montanez and Vijai Mohan and Vinay Satish Kumar and Vishal Mangla and Vlad Ionescu and Vlad Poenaru and Vlad Tiberiu Mihailescu and Vladimir Ivanov and Wei Li and Wenchen Wang and Wenwen Jiang and Wes Bouaziz and Will Constable and Xiaocheng Tang and Xiaojian Wu and Xiaolan Wang and Xilun Wu and Xinbo Gao and Yaniv Kleinman and Yanjun Chen and Ye Hu and Ye Jia and Ye Qi and Yenda Li and Yilin Zhang and Ying Zhang and Yossi Adi and Youngjin Nam and Yu and Wang and Yu Zhao and Yuchen Hao and Yundi Qian and Yunlu Li and Yuzi He and Zach Rait and Zachary DeVito and Zef Rosnbrick and Zhaoduo Wen and Zhenyu Yang and Zhiwei Zhao and Zhiyu Ma},
      year={2024},
      eprint={2407.21783},
      archivePrefix={arXiv},
      primaryClass={cs.AI},
}

@misc{mistralnemo2024,
  author = {{Mistral AI}},
  title = {Introducing Mistral NeMo},
  year = {2024},
  month = {May},
  url = {https://mistral.ai/news/mistral-nemo},
  note = {Accessed: 2025-05-16}
}

@misc{mihaylov2018suitarmorconductelectricity,
      title={Can a Suit of Armor Conduct Electricity? A New Dataset for Open Book Question Answering}, 
      author={Todor Mihaylov and Peter Clark and Tushar Khot and Ashish Sabharwal},
      year={2018},
      eprint={1809.02789},
      archivePrefix={arXiv},
      primaryClass={cs.CL},
      url={https://arxiv.org/abs/1809.02789}, 
}

@misc{rein2023gpqagraduatelevelgoogleproofqa,
      title={GPQA: A Graduate-Level Google-Proof Q \& A Benchmark}, 
      author={David Rein and Betty Li Hou and Asa Cooper Stickland and Jackson Petty and Richard Yuanzhe Pang and Julien Dirani and Julian Michael and Samuel R. Bowman},
      year={2023},
      eprint={2311.12022},
      archivePrefix={arXiv},
      primaryClass={cs.AI},
      url={https://arxiv.org/abs/2311.12022}, 
}

@misc{hendrycks2021measuringmassivemultitasklanguage,
      title={Measuring Massive Multitask Language Understanding}, 
      author={Dan Hendrycks and Collin Burns and Steven Basart and Andy Zou and Mantas Mazeika and Dawn Song and Jacob Steinhardt},
      year={2021},
      eprint={2009.03300},
      archivePrefix={arXiv},
      primaryClass={cs.CY},
      url={https://arxiv.org/abs/2009.03300}, 
}

@misc{clark2019boolqexploringsurprisingdifficulty,
      title={BoolQ: Exploring the Surprising Difficulty of Natural Yes/No Questions}, 
      author={Christopher Clark and Kenton Lee and Ming-Wei Chang and Tom Kwiatkowski and Michael Collins and Kristina Toutanova},
      year={2019},
      eprint={1905.10044},
      archivePrefix={arXiv},
      primaryClass={cs.CL},
      url={https://arxiv.org/abs/1905.10044}, 
}

@misc{rajpurkar2016squad100000questionsmachine,
      title={SQuAD: 100,000+ Questions for Machine Comprehension of Text}, 
      author={Pranav Rajpurkar and Jian Zhang and Konstantin Lopyrev and Percy Liang},
      year={2016},
      eprint={1606.05250},
      archivePrefix={arXiv},
      primaryClass={cs.CL},
      url={https://arxiv.org/abs/1606.05250}, 
}

@inproceedings{joshi-etal-2017-triviaqa,
    title = "{T}rivia{QA}: A Large Scale Distantly Supervised Challenge Dataset for Reading Comprehension",
    author = "Joshi, Mandar  and
      Choi, Eunsol  and
      Weld, Daniel  and
      Zettlemoyer, Luke",
    editor = "Barzilay, Regina  and
      Kan, Min-Yen",
    booktitle = "Proceedings of the 55th Annual Meeting of the Association for Computational Linguistics (Volume 1: Long Papers)",
    month = jul,
    year = "2017",
    address = "Vancouver, Canada",
    publisher = "Association for Computational Linguistics",
    url = "https://aclanthology.org/P17-1147/",
    doi = "10.18653/v1/P17-1147",
    pages = "1601--1611"
}

@misc{hendrycks2021measuringmathematicalproblemsolving,
      title={Measuring Mathematical Problem Solving With the MATH Dataset}, 
      author={Dan Hendrycks and Collin Burns and Saurav Kadavath and Akul Arora and Steven Basart and Eric Tang and Dawn Song and Jacob Steinhardt},
      year={2021},
      eprint={2103.03874},
      archivePrefix={arXiv},
      primaryClass={cs.LG},
      url={https://arxiv.org/abs/2103.03874}, 
}

@article{kwiatkowski-etal-2019-natural,
    title = "Natural Questions: A Benchmark for Question Answering Research",
    author = "Kwiatkowski, Tom  and
      Palomaki, Jennimaria  and
      Redfield, Olivia  and
      Collins, Michael  and
      Parikh, Ankur  and
      Alberti, Chris  and
      Epstein, Danielle  and
      Polosukhin, Illia  and
      Devlin, Jacob  and
      Lee, Kenton  and
      Toutanova, Kristina  and
      Jones, Llion  and
      Kelcey, Matthew  and
      Chang, Ming-Wei  and
      Dai, Andrew M.  and
      Uszkoreit, Jakob  and
      Le, Quoc  and
      Petrov, Slav",
    editor = "Lee, Lillian  and
      Johnson, Mark  and
      Roark, Brian  and
      Nenkova, Ani",
    journal = "Transactions of the Association for Computational Linguistics",
    volume = "7",
    year = "2019",
    address = "Cambridge, MA",
    publisher = "MIT Press",
    url = "https://aclanthology.org/Q19-1026/",
    doi = "10.1162/tacl_a_00276",
    pages = "452--466"
}

@misc{zellers2019hellaswagmachinereallyfinish,
      title={HellaSwag: Can a Machine Really Finish Your Sentence?}, 
      author={Rowan Zellers and Ari Holtzman and Yonatan Bisk and Ali Farhadi and Yejin Choi},
      year={2019},
      eprint={1905.07830},
      archivePrefix={arXiv},
      primaryClass={cs.CL},
      url={https://arxiv.org/abs/1905.07830}, 
}

@misc{clark2018thinksolvedquestionanswering,
      title={Think you have Solved Question Answering? Try ARC, the AI2 Reasoning Challenge}, 
      author={Peter Clark and Isaac Cowhey and Oren Etzioni and Tushar Khot and Ashish Sabharwal and Carissa Schoenick and Oyvind Tafjord},
      year={2018},
      eprint={1803.05457},
      archivePrefix={arXiv},
      primaryClass={cs.AI},
      url={https://arxiv.org/abs/1803.05457}, 
}

@misc{talmor2019commonsenseqaquestionansweringchallenge,
      title={CommonsenseQA: A Question Answering Challenge Targeting Commonsense Knowledge}, 
      author={Alon Talmor and Jonathan Herzig and Nicholas Lourie and Jonathan Berant},
      year={2019},
      eprint={1811.00937},
      archivePrefix={arXiv},
      primaryClass={cs.CL},
      url={https://arxiv.org/abs/1811.00937}, 
}

@misc{chen2021evaluatinglargelanguagemodels,
      title={Evaluating Large Language Models Trained on Code}, 
      author={Mark Chen and Jerry Tworek and Heewoo Jun and Qiming Yuan and Henrique Ponde de Oliveira Pinto and Jared Kaplan and Harri Edwards and Yuri Burda and Nicholas Joseph and Greg Brockman and Alex Ray and Raul Puri and Gretchen Krueger and Michael Petrov and Heidy Khlaaf and Girish Sastry and Pamela Mishkin and Brooke Chan and Scott Gray and Nick Ryder and Mikhail Pavlov and Alethea Power and Lukasz Kaiser and Mohammad Bavarian and Clemens Winter and Philippe Tillet and Felipe Petroski Such and Dave Cummings and Matthias Plappert and Fotios Chantzis and Elizabeth Barnes and Ariel Herbert-Voss and William Hebgen Guss and Alex Nichol and Alex Paino and Nikolas Tezak and Jie Tang and Igor Babuschkin and Suchir Balaji and Shantanu Jain and William Saunders and Christopher Hesse and Andrew N. Carr and Jan Leike and Josh Achiam and Vedant Misra and Evan Morikawa and Alec Radford and Matthew Knight and Miles Brundage and Mira Murati and Katie Mayer and Peter Welinder and Bob McGrew and Dario Amodei and Sam McCandlish and Ilya Sutskever and Wojciech Zaremba},
      year={2021},
      eprint={2107.03374},
      archivePrefix={arXiv},
      primaryClass={cs.LG},
      url={https://arxiv.org/abs/2107.03374}, 
}

@misc{austin2021programsynthesislargelanguage,
      title={Program Synthesis with Large Language Models}, 
      author={Jacob Austin and Augustus Odena and Maxwell Nye and Maarten Bosma and Henryk Michalewski and David Dohan and Ellen Jiang and Carrie Cai and Michael Terry and Quoc Le and Charles Sutton},
      year={2021},
      eprint={2108.07732},
      archivePrefix={arXiv},
      primaryClass={cs.PL},
      url={https://arxiv.org/abs/2108.07732}, 
}

@misc{jiang2023mistral7b,
      title={Mistral 7B}, 
      author={Albert Q. Jiang and Alexandre Sablayrolles and Arthur Mensch and Chris Bamford and Devendra Singh Chaplot and Diego de las Casas and Florian Bressand and Gianna Lengyel and Guillaume Lample and Lucile Saulnier and Lélio Renard Lavaud and Marie-Anne Lachaux and Pierre Stock and Teven Le Scao and Thibaut Lavril and Thomas Wang and Timothée Lacroix and William El Sayed},
      year={2023},
      eprint={2310.06825},
      archivePrefix={arXiv},
      primaryClass={cs.CL},
      url={https://arxiv.org/abs/2310.06825}, 
}

@misc{cobbe2021trainingverifierssolvemath,
      title={Training Verifiers to Solve Math Word Problems}, 
      author={Karl Cobbe and Vineet Kosaraju and Mohammad Bavarian and Mark Chen and Heewoo Jun and Lukasz Kaiser and Matthias Plappert and Jerry Tworek and Jacob Hilton and Reiichiro Nakano and Christopher Hesse and John Schulman},
      year={2021},
      eprint={2110.14168},
      archivePrefix={arXiv},
      primaryClass={cs.LG},
      url={https://arxiv.org/abs/2110.14168}, 
}

@misc{wang2024helpsteer2preferencecomplementingratingspreferences,
      title={HelpSteer2-Preference: Complementing Ratings with Preferences}, 
      author={Zhilin Wang and Alexander Bukharin and Olivier Delalleau and Daniel Egert and Gerald Shen and Jiaqi Zeng and Oleksii Kuchaiev and Yi Dong},
      year={2024},
      eprint={2410.01257},
      archivePrefix={arXiv},
      primaryClass={cs.LG},
      url={https://arxiv.org/abs/2410.01257}, 
}

@misc{Palmyra-4-Creative,
  author = {Writer Engineering team},
  title = {{Palmyra-Creative: A powerful LLM designed for creative writing}},
  howpublished = {\url{https://dev.writer.com}},
  year = 2024,
  month = Oct 
}

@article{mishra2024granite,
  title={Granite code models: A family of open foundation models for code intelligence},
  author={Mishra, Mayank and Stallone, Matt and Zhang, Gaoyuan and Shen, Yikang and Prasad, Aditya and Soria, Adriana Meza and Merler, Michele and Selvam, Parameswaran and Surendran, Saptha and Singh, Shivdeep and others},
  journal={arXiv preprint arXiv:2405.04324},
  year={2024}
}

@misc{qwen2025qwen25technicalreport,
      title={Qwen2.5 Technical Report}, 
      author={Qwen and : and An Yang and Baosong Yang and Beichen Zhang and Binyuan Hui and Bo Zheng and Bowen Yu and Chengyuan Li and Dayiheng Liu and Fei Huang and Haoran Wei and Huan Lin and Jian Yang and Jianhong Tu and Jianwei Zhang and Jianxin Yang and Jiaxi Yang and Jingren Zhou and Junyang Lin and Kai Dang and Keming Lu and Keqin Bao and Kexin Yang and Le Yu and Mei Li and Mingfeng Xue and Pei Zhang and Qin Zhu and Rui Men and Runji Lin and Tianhao Li and Tianyi Tang and Tingyu Xia and Xingzhang Ren and Xuancheng Ren and Yang Fan and Yang Su and Yichang Zhang and Yu Wan and Yuqiong Liu and Zeyu Cui and Zhenru Zhang and Zihan Qiu},
      year={2025},
      eprint={2412.15115},
      archivePrefix={arXiv},
      primaryClass={cs.CL},
      url={https://arxiv.org/abs/2412.15115}, 
}

@misc{yang2024qwen2technicalreport,
      title={Qwen2 Technical Report}, 
      author={An Yang and Baosong Yang and Binyuan Hui and Bo Zheng and Bowen Yu and Chang Zhou and Chengpeng Li and Chengyuan Li and Dayiheng Liu and Fei Huang and Guanting Dong and Haoran Wei and Huan Lin and Jialong Tang and Jialin Wang and Jian Yang and Jianhong Tu and Jianwei Zhang and Jianxin Ma and Jianxin Yang and Jin Xu and Jingren Zhou and Jinze Bai and Jinzheng He and Junyang Lin and Kai Dang and Keming Lu and Keqin Chen and Kexin Yang and Mei Li and Mingfeng Xue and Na Ni and Pei Zhang and Peng Wang and Ru Peng and Rui Men and Ruize Gao and Runji Lin and Shijie Wang and Shuai Bai and Sinan Tan and Tianhang Zhu and Tianhao Li and Tianyu Liu and Wenbin Ge and Xiaodong Deng and Xiaohuan Zhou and Xingzhang Ren and Xinyu Zhang and Xipin Wei and Xuancheng Ren and Xuejing Liu and Yang Fan and Yang Yao and Yichang Zhang and Yu Wan and Yunfei Chu and Yuqiong Liu and Zeyu Cui and Zhenru Zhang and Zhifang Guo and Zhihao Fan},
      year={2024},
      eprint={2407.10671},
      archivePrefix={arXiv},
      primaryClass={cs.CL},
      url={https://arxiv.org/abs/2407.10671}, 
}

@article{team2024codegemma,
  title={Codegemma: Open code models based on gemma},
  author={Team, CodeGemma and Zhao, Heri and Hui, Jeffrey and Howland, Joshua and Nguyen, Nam and Zuo, Siqi and Hu, Andrea and Choquette-Choo, Christopher A and Shen, Jingyue and Kelley, Joe and others},
  journal={arXiv preprint arXiv:2406.11409},
  year={2024}
}

@article{team2024gemma,
  title={Gemma 2: Improving open language models at a practical size},
  author={Team, Gemma and Riviere, Morgane and Pathak, Shreya and Sessa, Pier Giuseppe and Hardin, Cassidy and Bhupatiraju, Surya and Hussenot, L{\'e}onard and Mesnard, Thomas and Shahriari, Bobak and Ram{\'e}, Alexandre and others},
  journal={arXiv preprint arXiv:2408.00118},
  year={2024}
}

@article{qi2024safety,
  title={Safety alignment should be made more than just a few tokens deep},
  author={Qi, Xiangyu and Panda, Ashwinee and Lyu, Kaifeng and Ma, Xiao and Roy, Subhrajit and Beirami, Ahmad and Mittal, Prateek and Henderson, Peter},
  journal={arXiv preprint arXiv:2406.05946},
  year={2024}
}

@article{han2024chatgpt,
  title={ChatGPT based data augmentation for improved parameter-efficient debiasing of LLMs},
  author={Han, Pengrui and Kocielnik, Rafal and Saravanan, Adhithya and Jiang, Roy and Sharir, Or and Anandkumar, Anima},
  journal={arXiv preprint arXiv:2402.11764},
  year={2024}
}

@inproceedings{dai2024bias,
  title={Bias and unfairness in information retrieval systems: New challenges in the llm era},
  author={Dai, Sunhao and Xu, Chen and Xu, Shicheng and Pang, Liang and Dong, Zhenhua and Xu, Jun},
  booktitle={Proceedings of the 30th ACM SIGKDD Conference on Knowledge Discovery and Data Mining},
  pages={6437--6447},
  year={2024}
}

@inproceedings{song2023llm,
  title={Llm-planner: Few-shot grounded planning for embodied agents with large language models},
  author={Song, Chan Hee and Wu, Jiaman and Washington, Clayton and Sadler, Brian M and Chao, Wei-Lun and Su, Yu},
  booktitle={Proceedings of the IEEE/CVF international conference on computer vision},
  pages={2998--3009},
  year={2023}
}

@inproceedings{kang2023large,
  title={Large language models are few-shot testers: Exploring llm-based general bug reproduction},
  author={Kang, Sungmin and Yoon, Juyeon and Yoo, Shin},
  booktitle={2023 IEEE/ACM 45th International Conference on Software Engineering (ICSE)},
  pages={2312--2323},
  year={2023},
  organization={IEEE}
}

@inproceedings{li2024flexkbqa,
  title={Flexkbqa: A flexible llm-powered framework for few-shot knowledge base question answering},
  author={Li, Zhenyu and Fan, Sunqi and Gu, Yu and Li, Xiuxing and Duan, Zhichao and Dong, Bowen and Liu, Ning and Wang, Jianyong},
  booktitle={Proceedings of the AAAI conference on artificial intelligence},
  volume={38},
  number={17},
  pages={18608--18616},
  year={2024}
}

@inproceedings{zhao2021calibrate,
  title={Calibrate before use: Improving few-shot performance of language models},
  author={Zhao, Zihao and Wallace, Eric and Feng, Shi and Klein, Dan and Singh, Sameer},
  booktitle={International conference on machine learning},
  pages={12697--12706},
  year={2021},
  organization={PMLR}
}

@article{sreenivas2024llm,
  title={Llm pruning and distillation in practice: The minitron approach},
  author={Sreenivas, Sharath Turuvekere and Muralidharan, Saurav and Joshi, Raviraj and Chochowski, Marcin and Mahabaleshwarkar, Ameya Sunil and Shen, Gerald and Zeng, Jiaqi and Chen, Zijia and Suhara, Yoshi and Diao, Shizhe and others},
  journal={arXiv preprint arXiv:2408.11796},
  year={2024}
}

@book{stichlmair2021distillation,
  title={Distillation: principles and practice},
  author={Stichlmair, Johann G and Klein, Harald and Rehfeldt, Sebastian},
  year={2021},
  publisher={John Wiley \& Sons}
}

@article{hussein2017imitation,
  title={Imitation learning: A survey of learning methods},
  author={Hussein, Ahmed and Gaber, Mohamed Medhat and Elyan, Eyad and Jayne, Chrisina},
  journal={ACM Computing Surveys (CSUR)},
  volume={50},
  number={2},
  pages={1--35},
  year={2017},
  publisher={ACM New York, NY, USA}
}

@article{osa2018algorithmic,
  title={An algorithmic perspective on imitation learning},
  author={Osa, Takayuki and Pajarinen, Joni and Neumann, Gerhard and Bagnell, J Andrew and Abbeel, Pieter and Peters, Jan and others},
  journal={Foundations and Trends{\textregistered} in Robotics},
  volume={7},
  number={1-2},
  pages={1--179},
  year={2018},
  publisher={Now Publishers, Inc.}
}

@article{radford2019language,
  title={Language models are unsupervised multitask learners},
  author={Radford, Alec and Wu, Jeffrey and Child, Rewon and Luan, David and Amodei, Dario and Sutskever, Ilya and others},
  journal={OpenAI blog},
  volume={1},
  number={8},
  pages={9},
  year={2019}
}

@inproceedings{li2023llmrouter,
  title={LLMRouter: Enhancing Large Language Model Inference via Expert Selection},
  author={Li, Yining and Ma, Wenhao and Wang, Yizhe and others},
  booktitle={EMNLP 2023},
  year={2023}
}

@article{zhou2022star,
  title={STaR: Self-Taught Reasoner Bootstraps Reasoning with Reasoning},
  author={Zhou, Shinn and Lewis, Patrick and Khashabi, Daniel and Etzioni, Oren and Hajishirzi, Hannaneh},
  journal={arXiv preprint arXiv:2203.00555},
  year={2022}
}

@article{li2023thoughtprop,
  title={Thought Propagation: Faithful Zero-Shot Chain of Thought Reasoning with Self-Supervised Augmentation},
  author={Li, Yujia and Shi, Weijia and Du, Xinya and others},
  journal={arXiv preprint arXiv:2305.14251},
  year={2023}
}

@article{yao2023tree,
  title={Tree of Thoughts: Deliberate Problem Solving with Large Language Models},
  author={Yao, Shinn and Zhao, Dian and Yu, Yihan and others},
  journal={arXiv preprint arXiv:2305.10601},
  year={2023}
}

@article{jiang2023frugalgpt,
  title={FrugalGPT: How to Use a Cheap Model to Emulate a More Expensive Model},
  author={Jiang, Lizhi and Rajani, Nazneen and Yu, Zhou and others},
  journal={arXiv preprint arXiv:2305.05176},
  year={2023}
}

@article{wang2022selfconsistency,
  title={Self-Consistency Improves Chain of Thought Reasoning in Language Models},
  author={Wang, Xuezhi and Wei, Jason and Schuurmans, Dale and others},
  journal={arXiv preprint arXiv:2203.11171},
  year={2022}
}

@article{zhang2023llmjudge,
  title={LLM as a Judge: Revisiting LLM-augmented Evaluation for Instruction Tuning},
  author={Zhang, Boxin and Liu, Jindou and others},
  journal={arXiv preprint arXiv:2306.05685},
  year={2023}
}

@inproceedings{lewis2021ragfusion,
  title={RAG Fusion: Answering Ambiguous Questions by Aggregating Multiple Retrieved Passages},
  author={Lewis, Patrick and Oguz, Barlas and Yih, Wen-tau and Riedel, Sebastian},
  booktitle={NeurIPS},
  year={2021}
}

@article{ding2024routellm,
  title={RouteLLM: Learning to Route LLMs with Preference Data},
  author={Ding, Ning and Zhang, Hanyu and Liu, Zhengxiao and others},
  journal={arXiv preprint arXiv:2406.18665},
  year={2024}
}

@article{stripelis2024torouter,
  title={TensorOpera Router: A Multi-Model Router for Efficient LLM Inference},
  author={Stripelis, Dimitrios and Tang, Yi and Rozi, Muhammad and others},
  journal={arXiv preprint arXiv:2408.12320},
  year={2024}
}

@article{li2024glider,
  title={Glider: Global and Local Instruction-Driven Expert Router},
  author={Li, Xinyang and others},
  journal={arXiv preprint arXiv:2410.07172},
  year={2024}
}

@article{anonymous2024expert,
  title={Expert Router: Orchestrating Efficient Language Model Inference},
  author={Anonymous},
  journal={arXiv preprint arXiv:2404.15153},
  year={2024}
}

@article{beltagy2020longformer,
  title={Longformer: The long-document transformer},
  author={Beltagy, Iz and Peters, Matthew E and Cohan, Arman},
  journal={arXiv preprint arXiv:2004.05150},
  year={2020}
}

@article{hinton2015distilling,
  title={Distilling the knowledge in a neural network},
  author={Hinton, Geoffrey and Vinyals, Oriol and Dean, Jeff},
  journal={arXiv preprint arXiv:1503.02531},
  year={2015}
}

@inproceedings{xie2020self,
  title={Self-training with noisy student improves imagenet classification},
  author={Xie, Qizhe and Luong, Minh-Thang and Hovy, Eduard and Le, Quoc V},
  booktitle={Proceedings of the IEEE/CVF conference on computer vision and pattern recognition},
  pages={10687--10698},
  year={2020}
}

@article{roziere2023code,
  title={Code llama: Open foundation models for code},
  author={Roziere, Baptiste and Gehring, Jonas and Gloeckle, Fabian and Sootla, Sten and Gat, Itai and Tan, Xiaoqing Ellen and Adi, Yossi and Liu, Jingyu and Sauvestre, Romain and Remez, Tal and others},
  journal={arXiv preprint arXiv:2308.12950},
  year={2023}
}

@article{li2023starcoder,
  title={Starcoder: may the source be with you!},
  author={Li, Raymond and Allal, Loubna Ben and Zi, Yangtian and Muennighoff, Niklas and Kocetkov, Denis and Mou, Chenghao and Marone, Marc and Akiki, Christopher and Li, Jia and Chim, Jenny and others},
  journal={arXiv preprint arXiv:2305.06161},
  year={2023}
}

@article{yang2024modelmerging,
  title={Model merging in llms, mllms, and beyond: Methods, theories, applications and opportunities},
  author={Yang, Enneng and Shen, Li and Guo, Guibing and Wang, Xingwei and Cao, Xiaochun and Zhang, Jie and Tao, Dacheng},
  journal={arXiv preprint arXiv:2408.07666},
  year={2024}
}

@article{zhang2024aflow,
  title={Aflow: Automating agentic workflow generation},
  author={Zhang, Jiayi and Xiang, Jinyu and Yu, Zhaoyang and Teng, Fengwei and Chen, Xionghui and Chen, Jiaqi and Zhuge, Mingchen and Cheng, Xin and Hong, Sirui and Wang, Jinlin and others},
  journal={arXiv preprint arXiv:2410.10762},
  year={2024}
}

@article{xu2024distillation-survey,
  title={A survey on knowledge distillation of large language models},
  author={Xu, Xiaohan and Li, Ming and Tao, Chongyang and Shen, Tao and Cheng, Reynold and Li, Jinyang and Xu, Can and Tao, Dacheng and Zhou, Tianyi},
  journal={arXiv preprint arXiv:2402.13116},
  year={2024}
}

@article{wan2024fusellm,
  title={Knowledge fusion of large language models},
  author={Wan, Fanqi and Huang, Xinting and Cai, Deng and Quan, Xiaojun and Bi, Wei and Shi, Shuming},
  journal={arXiv preprint arXiv:2401.10491},
  year={2024}
}

@article{jiang2023Llm-blender,
  title={Llm-blender: Ensembling large language models with pairwise ranking and generative fusion},
  author={Jiang, Dongfu and Ren, Xiang and Lin, Bill Yuchen},
  journal={arXiv preprint arXiv:2306.02561},
  year={2023}
}
\bibliographystyle{tmlr}

\newpage
\appendix

\section{Limitations and Future Work}
\label{sec:limitation}

While our work focuses on evaluating real-world task performance - particularly in domains like math and coding - future research should also investigate aspects of trustworthiness \citep{qi2024safety, han2024context} and justice \citep{han2024chatgpt, dai2024bias} in the fusion process. The act of fusing multiple responses may inadvertently introduce social biases, inaccuracies, or conflicting reasoning, which could impact the reliability and ethical soundness of the LLMs. Exploring safeguards or calibration techniques to ensure more robust and equitable fusion outcomes presents a valuable direction for further study.

\revise{
Our study focuses on a practical and broadly accessible form of multi-LLM log data that can be collected in both local and API-based serving scenarios. Specifically, \bench records aligned responses from multiple LLMs for the same query, together with model identity, response mode, inference cost, task-specific performance, and LLM-judge scores. These fields provide useful supervision signals for query-level routing, thought-level template construction, and model-level distillation. This design choice is intentional because, in many real-world deployments, especially when using closed-source or API-based LLMs, users may only have access to textual responses and serving metadata rather than model logits, internal states, or training-time information.
}

\revise{
Richer production logs may additionally contain interaction-level signals, such as temporal order, explicit user feedback, error patterns, user revision behavior, and multi-turn interaction history. These signals can potentially enable more fine-grained fusion strategies. For example, temporal order may help track how model behavior changes over time, user feedback may provide direct preference supervision, and interaction history may support personalized or context-aware fusion. However, such signals are not always consistently available across public benchmarks, local models, and API-based serving systems. Incorporating these richer interaction-level log signals is therefore an important direction for future extensions of \bench and \method.
}

\revise{
A related concern is the privacy risk associated with log-based fusion. Multi-LLM logs may contain user queries, model responses, feedback, and interaction traces, which can include sensitive or personally identifiable information in real-world deployments. Although \bench is constructed from public evaluation datasets and does not involve private user data, practical systems built on multi-LLM logs should adopt privacy-aware data handling procedures. These may include anonymization, personally identifiable information filtering, access control, retention policies, and aggregation or synthetic-log generation before logs are used for retrieval, training, or distillation.
}

\revise{
Another limitation is the potential negative effect of low-quality logs. Since \method relies on logged responses and feedback signals, noisy, biased, hallucinated, or incorrectly scored responses may propagate into the fusion process. In query-level fusion, low-quality logs may lead to suboptimal routing policies. In thought-level fusion, misleading responses may be summarized into unhelpful thought templates. In model-level fusion, noisy or biased responses may be amplified through supervised fine-tuning. Future work should therefore investigate stronger log-quality control mechanisms, such as uncertainty-aware filtering, consistency checking, human or model-assisted verification, and robust training objectives that reduce the influence of unreliable logged responses.
}

\revise{
In addition to log quality, API-based deployment of fusion systems may introduce reliability and transparency risks. 
A query-level router may select an unsuitable model, and a thought-level retriever may retrieve irrelevant or misleading templates, causing errors that are difficult to diagnose from the final response alone. 
Practical fusion systems should therefore expose routing decisions, retrieved templates, confidence estimates, and fallback mechanisms whenever possible, so that users and developers can better audit and control the fusion process.
}

\revise{
Our current benchmark mainly focuses on single-turn benchmark queries, while real-world LLM logs often involve long documents, evolving user intents, tool-use traces, and multi-turn interaction histories. These settings can introduce additional failure modes. For long-text inputs, retrieved thought templates may omit important context or conflict with document-specific evidence. For multi-turn dialogue, query-level similarity alone may fail to capture changing user goals, and model-level distillation may compress incomplete or stale interaction states. Extending \bench and \method to long-context reasoning, multi-turn dialogue, tool-augmented agents, and distribution-shifted deployment settings is an important future direction.
}

\revise{
This work is primarily an empirical benchmark and framework study. We focus on constructing a large-scale multi-LLM log benchmark and systematically evaluating how the same log source can be used across query-level, thought-level, and model-level fusion. Therefore, our goal is not to provide formal theoretical guarantees or closed-form complexity bounds. A more formal theoretical treatment of multi-LLM fusion remains an important future direction. For example, future work could develop formal criteria for selecting fusion stages based on query difficulty, uncertainty, domain shift, cost constraints, or the expected reliability of retrieved logs.
}

\revise{
Although several prior routing and distillation methods already support API-only settings, our study emphasizes a broader fusion-stage matrix rather than API compatibility alone. Existing routing methods mainly exploit logs for query-level model selection, while distillation methods mainly use model outputs for model-level capability transfer. In contrast, \method studies how the same structured multi-LLM log data can support routing, reasoning-template augmentation, and supervised fine-tuning under a unified benchmark and evaluation protocol. This matrix provides a practical decision framework for choosing how to use multi-LLM logs under different deployment constraints.
}

\revise{
Future work could further extend this fusion-stage matrix by developing hybrid or adaptive fusion strategies. For example, a system may use query-level fusion for simple or cost-sensitive queries, invoke thought-level fusion for reasoning-intensive queries, and apply model-level fusion when repeated high-quality logs are available in a stable domain. Such adaptive strategies could dynamically choose the appropriate fusion level according to task difficulty, budget, domain type, or uncertainty estimates, further improving the utility of multi-LLM log data in real-world deployments.
}

\section{More on \bench} \label{app:dataset}
Here, we list all the tasks and the metrics used for each in Table~\ref{app_tab:1}, and all models along with their sizes and costs in Table~\ref{tab:llm-costs}.

\begin{table}[h]
  \centering
  \caption{\textbf{The tasks and corresponding evaluation metrics of the dataset used in constructing \bench, organized by domain.}}
  \label{app_tab:1}
  \begin{footnotesize}
  \setlength{\tabcolsep}{8pt}
  \renewcommand{\arraystretch}{1.2}
  \begin{tabularx}{0.85\linewidth}{@{}p{3cm} p{4cm} X@{}}
    \toprule
    \textbf{Domain} & \textbf{Tasks} & \textbf{Metrics} \\
    \midrule
    \multirow{2}{*}{Math} & GSM8K \citep{cobbe2021trainingverifierssolvemath} & Accuracy \\
     & MATH \citep{hendrycks2021measuringmathematicalproblemsolving} & Accuracy \\
    \midrule
    \multirow{2}{*}{Code} & MBPP \citep{austin2021programsynthesislargelanguage} & Pass@1 \\
     & HumanEval \citep{chen2021evaluatinglargelanguagemodels} & Pass@1 \\
    \midrule
    \multirow{4}{*}{\parbox{3cm}{Commonsense\\Reasoning}} & CommonsenseQA \citep{talmor2019commonsenseqaquestionansweringchallenge} & Accuracy \\
     & ARC \citep{clark2018thinksolvedquestionanswering} & Accuracy \\
     & HellaSwag \citep{zellers2019hellaswagmachinereallyfinish} & Accuracy \\
     & OpenBookQA \citep{mihaylov2018suitarmorconductelectricity} & Accuracy \\
    \midrule
    \multirow{2}{*}{\parbox{3cm}{World\\Knowledge}} & NaturalQuestions \citep{kwiatkowski-etal-2019-natural} & CEM \\
     & TriviaQA \citep{joshi-etal-2017-triviaqa} & CEM \\
    \midrule
    \multirow{3}{*}{\parbox{3cm}{Reading\\Comprehension}} & SQuAD \citep{rajpurkar2016squad100000questionsmachine} & CEM \\
     & BoolQ \citep{clark2019boolqexploringsurprisingdifficulty} & CEM \\
    \midrule
    \multirow{2}{*}{Popular} & MMLU \citep{hendrycks2021measuringmassivemultitasklanguage} & Multi-task Accuracy \\
     & GPQA \citep{rein2023gpqagraduatelevelgoogleproofqa} & Accuracy \\
    \bottomrule
  \end{tabularx}
  \end{footnotesize}
\end{table}


\begin{table}[h]
\caption{\textbf{Language Models and estimated price (in \$ per 1M tokens).}}
\label{tab:llm-costs}
\centering
\resizebox{\textwidth}{!}{%
\begin{tabular}{lcccc}
\toprule
\textbf{Size Type} & \textbf{Model} & \textbf{Size} & \textbf{Input Price} & \textbf{Output Price} \\
\midrule
\multirow{10}{*}{Small} & Gemma-2 (9b) \citep{team2024gemma} & 9B & 0.20 & 0.20 \\
 & Qwen2 (7b) \citep{yang2024qwen2technicalreport} & 7B & 0.20 & 0.20 \\
 & Gemma (7b) \citep{team2024gemma} & 7B & 0.20 & 0.20 \\
 & CodeGemma (7b) \citep{team2024codegemma} & 7B & 0.20 & 0.20 \\
 & Mistral (7b) \citep{jiang2023mistral7b} & 7B & 0.20 & 0.20 \\
 & LLaMA-3.1 (8b) \citep{grattafiori2024llama3herdmodels} & 8B & 0.20 & 0.20 \\
 & Granite (8b) \citep{mishra2024granite} & 8B & 0.20 & 0.20 \\
 & LLaMA-3 ChatQA (8b) \citep{liu2024chatqa} & 8B & 0.20 & 0.20 \\
 & Qwen2.5 (7b) \citep{qwen2025qwen25technicalreport} & 7B & 0.20 & 0.20 \\
 & Mistral-Nemo (12b) \citep{mistralnemo2024} & 12B & 0.30 & 0.30 \\
\midrule
\multirow{7}{*}{Medium} & Granite Code (34b) \citep{mishra2024granite} & 34B & 0.80 & 0.80 \\
 & LLaMA-3.3 Nemotron Super (49b)\citep{wang2024helpsteer2preferencecomplementingratingspreferences} & 49B & 0.90 & 0.90 \\
 & LLaMA-3.1 Nemotron (51b) \citep{wang2024helpsteer2preferencecomplementingratingspreferences} & 51B & 0.90 & 0.90 \\
 & Mixtral (8x7b) \citep{jiang2024mixtral} & 56B (8×7B) & 0.60 & 0.60 \\
 & LLaMA-3.1 (70b)\citep{grattafiori2024llama3herdmodels} & 70B & 0.90 & 0.90 \\
 & LLaMA-3 ChatQA (70b) \citep{liu2024chatqa} & 70B & 0.90 & 0.90 \\
 & LLaMA-3 (70b) \citep{grattafiori2024llama3herdmodels} & 70B & 0.90 & 0.90 \\
\midrule
\multirow{3}{*}{Large} & Palmyra Creative (122b) \citep{Palmyra-4-Creative} & 122B & 1.80 & 1.80 \\
 & Mixtral (8x22b)\citep{jiang2024mixtral} & 176B (8×22B) & 1.20 & 1.20 \\
 & DeepSeek-R1 (671b) \citep{guo2025deepseek} & 671B & 0.55 & 2.19 \\
\bottomrule
\end{tabular}
}
\end{table}

\begin{table}[htbp]
  \centering
  \caption{\textbf{Data collection information.}}
  \label{tab:data_collection_info}
  \renewcommand{\arraystretch}{1.2}   
  \setlength{\tabcolsep}{6pt}         
  \begin{tabularx}{\linewidth}{lX}
    \toprule
    \multirow{3}{*}{\textbf{Collected Data Fields}} & 
    Task Name; Task Description; Task Description Embedding; Query; Query Embedding; Ground Truth; Metric; LLM; Input Price; Output Price; Input Tokens Num; Output Tokens Num; Performance; Cost; Response; LLM Description \\
    \bottomrule
  \end{tabularx}
\end{table}

\begin{table}[h]
\centering
\caption{\textbf{Sample counts in the \bench training set, partial test set, and full test set across different tasks.}}
\label{app_tab:2}
\resizebox{\textwidth}{!}{%
\begin{tabular}{@{}l l c c c c c@{}}
\toprule
\textbf{Domain} & \textbf{Tasks} & \textbf{Train} & \textbf{Train Tokens} & \textbf{Partial Test} & \textbf{Test Tokens} & \textbf{Full Test} \\
\midrule
\multirow{2}{*}{Math} 
& GSM8K \citep{cobbe2021trainingverifierssolvemath} & 500 & 5.838M & 50 & 551K & 1,319 \\
& MATH \citep{hendrycks2021measuringmathematicalproblemsolving} & 500 & 8.591M & 50 & 944K & 5,000 \\
\midrule
\multirow{2}{*}{Code} 
& MBPP \citep{austin2021programsynthesislargelanguage} & 374 & 4.702M & 50 & 623K & 500 \\
& HumanEval \citep{chen2021evaluatinglargelanguagemodels} & 120 & 2.268M & 44 & 836K & 44 \\
\midrule
\multirow{4}{*}{\begin{tabular}[c]{@{}l@{}}Commonsense\\Reasoning\end{tabular}} 
& CommonsenseQA \citep{talmor2019commonsenseqaquestionansweringchallenge} & 500 & 4.511M & 50 & 423K & 1,221 \\
& ARC \citep{clark2018thinksolvedquestionanswering} & 500 & 5.387M & 50 & 489K & 1,172 \\
& HellaSwag \citep{zellers2019hellaswagmachinereallyfinish} & 500 & 7.969M & 50 & 795K & 10,042 \\
& OpenBookQA \citep{mihaylov2018suitarmorconductelectricity} & 500 & 4.496M & 50 & 431K & 500 \\
\midrule
\multirow{2}{*}{\begin{tabular}[c]{@{}l@{}}World\\Knowledge\end{tabular}} 
& NaturalQuestions \citep{kwiatkowski-etal-2019-natural} & 500 & 4.792M & 50 & 444K & 3,610 \\
& TriviaQA \citep{joshi-etal-2017-triviaqa} & 500 & 3.722M & 50 & 323K & 17,944 \\
\midrule
\multirow{2}{*}{\begin{tabular}[c]{@{}l@{}}Reading\\Comprehension\end{tabular}} 
& SQuAD \citep{rajpurkar2016squad100000questionsmachine} & 500 & 6.572M & 50 & 646K & 10,570 \\
& BoolQ \citep{clark2019boolqexploringsurprisingdifficulty} & 500 & 7.101M & 50 & 671K & 3,270 \\
\midrule
\multirow{2}{*}{Popular} 
& MMLU \citep{hendrycks2021measuringmassivemultitasklanguage} & 500 & 10.515M & 50 & 628K & 14,042 \\
& GPQA \citep{rein2023gpqagraduatelevelgoogleproofqa} & 400 & 10.044M & 44 & 1.2M & 44 \\
\bottomrule
\end{tabular}%
}
\end{table}

\section{Router Baseline Description for Query-Level Fusion} 
\label{app:query_level_fusion}

For the query-level fusion, we evaluate five representative routing methods to assess their LLM fusion capabilities: (1) \textit{RouterKNN} \citep{shnitzer2023large}, a non-parametric baseline that routes by nearest neighbors in query space and selects the majority LLM label; (2) \textit{RouterSVM} \citep{hu2024routerbench}, a support vector machine trained on query features and task labels; (3) \textit{RouterMLP} \citep{shnitzer2023large}, a multi-layer perceptron leveraging query embeddings and task context; (4) \textit{RouterBERT} \citep{ong2024routellm}, a compact BERT classifier that encodes both query and task to predict the optimal LLM; and (5) \textit{GraphRouter} \citep{feng2024graphrouter}, a graph-based model that treats routing as node classification over a heterogeneous graph of queries, tasks, and LLMs with learned edge interactions. All models use \texttt{all-MiniLM-L6-v2} for embeddings, except GraphRouter, which employs Longformer \citep{beltagy2020longformer} to obtain the embeddings for task/query/LLM.  We also include several static baselines: \textit{Best LLM}, which achieves the highest performance on the training set; \textit{LLM-Mean}, which yields the average test performance across all LLMs; and \textit{Largest LLM}, which denotes the test performance of the largest model.

\section{Inference Time for Thought-Level Fusion} 
\label{app:inference_time_thought_level}

\paragraph{Setup and offline preprocessing.}
As introduced in Table~\ref{Tab:FusionBench_intro} and Figure~\ref{fig:2.1}, the thought templates used by the thought-level fusion style are summarized \emph{offline} on \bench. For new queries, we do not re-generate summaries; we only retrieve the most relevant thought templates and prepend them to the prompt. Retrieval is implemented with FAISS, and responses are produced using the NVIDIA NIM API. This design minimizes online overhead and keeps inference practical for API-based serving.

\paragraph{Inference-time comparison.}
Table~\ref{tab:inference_time_thought} reports average inference times on the full test set with \texttt{LLaMA-3 (8B)} as the backbone across three paradigms: Zero-Shot, Few-Shot, and Thought-Hybrid. Thought-level fusion incurs only a minor increase relative to Zero-Shot and Few-Shot, which is attributable to the extra input tokens from the retrieved thought templates. The average lookup time for the thought-based method is \textbf{0.03\,s}, accounting for \textbf{0.7\%} of the total Thought-Hybrid inference time, which is essentially negligible. In return, the performance gains are substantial. We also expect the relative overhead to further diminish with larger base models, since total inference time increases while lookup time remains nearly constant.

\begin{table}[h]
    \centering
    \caption{\textbf{Average inference time (seconds) on the full test set with \texttt{LLaMA-3 (8B)} as the backbone.} Thought-level fusion adds minimal overhead due to template tokens, while retrieval adds only 0.03\,s on average.}
    \label{tab:inference_time_thought}
    \vspace{2mm}
    \begin{tabular}{lc}
        \toprule
        \textbf{Method} & \textbf{Mean Inference Time (s)} \\
        \midrule
        Zero-Shot      & 3.74 \\
        Few-Shot       & 3.89 \\
        Thought-Hybrid & 3.91 \\
        \bottomrule
    \end{tabular}
\end{table}

\section{Prompts for Thought-Level Fusion} 
\label{app:thought_prompt}
Previous studies have shown that thought templates can enhance the performance of LLMs \citep{feng2024thought, yang2024buffer}. However, existing templates are often either too general to a big category of questions \citep{yang2024buffer} or too specific, tailored to only one particular case \citep{feng2024thought}. Our prompt~\ref{table:thoughtTcreation} is designed to strike a balance between these extremes: it is specific enough to guide similar problems effectively, yet general enough to be applicable across a range of variations.

\begin{table}[h]
\centering
\caption{\textbf{Prompts for thought template creation}.}
\begin{tabular}{p{13cm}}
\toprule[1.1pt]
\\
Given this question and example solutions, extract a concise thought template that captures the effective reasoning pattern and can serve as guidance:\\
\\
\textcolor{blue}{\textbf{Question:}} [QUERY]\\
\\
\textcolor{blue}{\textbf{Here are 3 high-performing solutions:}}\\
Solution 1: [Response 1]\\
Solution 2: [Response 2]\\
Solution 3: [Response 3]\\
\\
\textcolor{blue}{\textbf{Please create a concise and clear thought template (1--5 sentences total) focusing on:}}\\
\begin{itemize}
\item \textbf{Core Task Summarization:} Identify the core problem type and general approach needed (1 sentence).
\item \textbf{Reasoning Step:} Provide a clear chain of thought to address this problem (1--3 sentences).
\item \textbf{Answer Template:} Describe the preferred answer format or structure (1 sentence).
\end{itemize}
\\
\textcolor{red}{\textbf{Your template should be specific enough to guide similar problems but general enough to work across variations.}}\\
\\
\textbf{Thought Template:}\\
\bottomrule[1.1pt]
\label{table:thoughtTcreation}
\end{tabular}
\end{table}











\section{Detailed Results for Comparison Across Three \method Levels}

In this section, we present detailed results for each task across all methods from the three aforementioned fusion levels, as shown in Table~\ref{tab:thought_fusion_tasks}.

\begin{table*}[htbp]
\centering
\caption{\textbf{The model names and their specific settings.} We introduce these model names in Table \ref{tab:thought_fusion_tasks} in terms of the aspects corresponding to Table~\ref{tab:level_4_results}, namely Fusion Level, Fusion LLM Size, and Fusion Criteria.}
\vspace{-2mm}
\setlength\tabcolsep{3pt}
\scriptsize  
\resizebox{0.7\textwidth}{!}{
\begin{tabular}{lccc}
\toprule
\textbf{Model Name} & \textbf{Fusion Level} & \textbf{Fusion LLM Size} & \textbf{Fusion Criteria} \\
\midrule
\textbf{Zero-shot} & Zero-shot & Small & N/A \\
\midrule
\textbf{Query-Small} & Query & Small & Performance \\
\midrule
\textbf{Thought-Small-8b} & Thought & Small & Hybrid-8b \\
\textbf{Thought-Full-8b} & Thought & Full & Hybrid-8B \\
\textbf{Thought-Small-70b} & Thought & Small & Hybrid-70B \\
\textbf{Thought-Full-70b} & Thought & Full & Hybrid-70b \\
\midrule
\textbf{Model-Small} & Model & Small & LLM Judge \\
\textbf{Model-Full} & Model & Full & LLM Judge \\
\bottomrule
\end{tabular}
}
\label{tab:level_4_model_name}
\end{table*}

\begin{table}[htbp]
\caption{\textbf{Detailed results for all levels of fusion across all benchmarks.}}
\label{tab:thought_fusion_tasks}
\centering
\resizebox{1\linewidth}{!}{%
\begin{tabular}{c|lcccccccc}
\toprule
\textbf{Scenario} & \textbf{Task} & \textbf{Zero-shot} & \textbf{Query-Small} & \textbf{Thought-Small-8b} & \textbf{Thought-Full-8b} & \textbf{Thought-Small-70b} & \textbf{Thought-Full-70b} & \textbf{Model-Small} & \textbf{Model-Full} \\
\midrule
\multirow{2}{*}{\textbf{Math}} & \textbf{GSM8K} & \large{0.749} & \large{0.7983} & \large{0.771} & \large{0.651} & \large{0.771} & \large{0.749} & \large{0.715} & \large{0.742} \\
 & \textbf{MATH} & \large{0.455} & \large{0.4000} & \large{0.449} & \large{0.437} & \large{0.452} & \large{0.427} & \large{0.344} & \large{0.314} \\
\midrule
\multirow{2}{*}{\textbf{Code}} & \textbf{HumanEval} & \large{0.432} & \large{0.4545} & \large{0.546} & \large{0.530} & \large{0.500} & \large{0.523} & \large{0.566} & \large{0.485} \\
 & \textbf{MBPP} & \large{0.601} & \large{0.5900} & \large{0.300} & \large{0.598} & \large{0.600} & \large{0.600} & \large{0.455} & \large{0.386} \\
\midrule
\multirow{4}{*}{\textbf{\begin{tabular}[c]{@{}c@{}}Commonsense\\Reasoning\end{tabular}}} & \textbf{ARC (Challenge)} & \large{0.736} & \large{0.8276} & \large{0.819} & \large{0.839} & \large{0.819} & \large{0.835} & \large{0.689} & \large{0.812} \\
 & \textbf{CommonsenseQA} & \large{0.723} & \large{0.7371} & \large{0.757} & \large{0.770} & \large{0.757} & \large{0.766} & \large{0.716} & \large{0.792} \\
 & \textbf{HellaSwag} & \large{0.649} & \large{0.7424} & \large{0.745} & \large{0.757} & \large{0.745} & \large{0.750} & \large{0.727} & \large{0.794} \\
 & \textbf{OpenBookQA} & \large{0.750} & \large{0.8620} & \large{0.792} & \large{0.830} & \large{0.792} & \large{0.834} & \large{0.700} & \large{0.810} \\
\midrule
\multirow{2}{*}{\textbf{\begin{tabular}[c]{@{}c@{}}World\\Knowledge\end{tabular}}} & \textbf{NaturalQuestions} & \large{0.473} & \large{0.4716} & \large{0.409} & \large{0.414} & \large{0.427} & \large{0.416} & \large{0.404} & \large{0.414} \\
 & \textbf{TriviaQA} & \large{0.195} & \large{0.2034} & \large{0.192} & \large{0.193} & \large{0.197} & \large{0.189} & \large{0.168} & \large{0.165} \\
\midrule
\multirow{2}{*}{\textbf{\begin{tabular}[c]{@{}c@{}}Reading\\Comprehension\end{tabular}}} 
& \textbf{SQuAD} & \large{0.825} & \large{0.8746} & \large{0.921} & \large{0.947} & \large{0.948} & \large{0.950} & \large{0.849} & \large{0.863} \\
 
& \textbf{BoolQ} & \large{0.843} & \large{0.8492} & \large{0.991} & \large{0.983} & \large{0.985} & \large{0.990} & \large{0.852} & \large{0.838} \\
 
\midrule
\multirow{2}{*}{\textbf{Popular}} & \textbf{GPQA} & \large{0.186} & \large{0.1818} & \large{0.273} & \large{0.227} & \large{0.273} & \large{0.273} & \large{0.205} & \large{0.182} \\
 & \textbf{MMLU} & \large{0.610} & \large{0.6073} & \large{0.612} & \large{0.676} & \large{0.621} & \large{0.672} & \large{0.614} & \large{0.668} \\
\bottomrule
\end{tabular}
}
\end{table}

\section{More Analysis: Does Reasoning Data Help for Reasoning-Oriented Tasks?}
\label{app:reasoning_data_help}

\paragraph{Motivation.}
As discussed around thought-level fusion, \bench includes both reasoning and non-reasoning responses (LLM-think vs.\ LLM-direct). 

\paragraph{Analysis setup.}
We analyze, under the \emph{thought-level hybrid} selection setting, the composition of best-selected responses per task (reasoning vs.\ non-reasoning). This directly tests whether reasoning-style data is preferentially selected for tasks that demand more reasoning.

\paragraph{Findings.}
Table~\ref{tab:reasoning_ratio} shows that (1) reasoning ratios are higher for most reasoning-oriented tasks—\textbf{Math}, \textbf{Code}, \textbf{Commonsense Reasoning}, and \textbf{Popular}—indicating that such tasks benefit from reasoning traces. (2) In contrast, \textbf{World Knowledge} and \textbf{Reading Comprehension} skew toward non-reasoning outputs, consistent with their stronger reliance on factual recall and concise evidence extraction. Overall, these results support that reasoning-level information in \bench aids selection for reasoning-heavy tasks while preserving flexibility for knowledge-intensive ones.

\begin{table}[h]
    \centering
    \caption{\textbf{Reasoning vs.\ Non-Reasoning ratios under thought-level hybrid selection (by domain).}}
    \label{tab:reasoning_ratio}
    \vspace{2mm}
    \begin{tabular}{lcc}
        \toprule
        \textbf{Task Domain} & \textbf{Reasoning Ratio (\%)} & \textbf{Non-Reasoning Ratio (\%)} \\
        \midrule
        Math                  & 51.25 & 48.75 \\
        Code                  & 53.18 & 46.82 \\
        Commonsense Reasoning & 71.00 & 29.00 \\
        World Knowledge       & 37.90 & 62.10 \\
        Reading Comprehension & 44.10 & 55.90 \\
        Popular               & 57.50 & 42.50 \\
        \bottomrule
    \end{tabular}
\end{table}

\section{Implementation details of \method}
\label{sec:computes}
We utilize NVIDIA API\footnote{\url{https://build.nvidia.com/nim}} for our API calling in data generation and model inference. We also use a NVIDIA A6000 GPU to obtain all the embeddings for queries/tasks/LLMs described in section \ref{sec:4}. 

For model-level fusion, we adopt LLaMA-Factory\footnote{\url{https://github.com/hiyouga/LLaMA-Factory}} to fine-tune the model with LoRA, which is conducted on NVIDIA A6000 GPUs.
For LoRA, we apply adaptation to all transformer layers with a rank of 8. Inputs are processed with a maximum context length of 2048 tokens unless otherwise specified. The training uses a per-device batch size of 8 and a gradient accumulation step of 4, resulting in an effective batch size of 32. We set the learning rate to 1e-4 and train the model for 3 full epochs. A cosine learning rate scheduler is used with a warmup ratio of 0.1 to ensure stable convergence. Training is conducted using bfloat16 precision, and Flash Attention 2 is enabled to accelerate attention computation. Note that, as shown in Table~\ref{app_tab:2}, the training data in the Code domain is relatively smaller and more challenging compared to other domains. Therefore, following the approach adopted in previous work \citep{roziere2023code,li2023starcoder}, we train a separate model for the Code domain, while a unified model is trained for the remaining domains.

\newpage
\section{Prompts for LLM-As-Judge Score Generation} 
\label{app:LLM-As-Judge_prompt}

We present the prompt for LLM-As-Judge Score Generation in Table \ref{prompt-template:llm judge}.

\begin{table}[h]
\centering
\caption{\textbf{Prompts for LLM-As-Judge score generation}.}
\begin{tabular}{p{13cm}}
\toprule[1.1pt]
    You are an expert judge evaluating the quality of an AI model's response. Please score the response based on the following criteria:

\begin{enumerate}
  \item \textbf{Correctness (0–1):} Is the answer correct according to the ground truth?
  \item \textbf{Thought Process (0–1):} Does the response show clear reasoning and explanation?
  \item \textbf{Training Data Quality (0–1):} Is the response well-structured and suitable for supervised fine-tuning?
\end{enumerate}

\vspace{0.5em}
\textbf{\textcolor{blue}{Query:}} \texttt{\{query\}}

\vspace{0.3em}
\textbf{\textcolor{blue}{Ground Truth:}} \texttt{\{ground\_truth\}}

\vspace{0.3em}
\textbf{\textcolor{blue}{Response:}} \texttt{\{response\}}

\vspace{0.5em}
Please provide a single score from 0 to 1, where:
\begin{itemize}
  \item 0: Incorrect answer
  \item 0.3: Correct answer but minimal thought process
  \item 0.6: Correct answer with some thought process
  \item 0.8: Correct answer with good thought process
  \item 1.0: Correct answer with excellent thought process and well-suited for training
\end{itemize}

\vspace{0.5em}
Return the score in the following format:

\vspace{0.5em}
\texttt{\texttt{<answer>}SCORE\texttt{</answer>}} \\
\bottomrule[1.1pt]
\label{prompt-template:llm judge}
\end{tabular}
\end{table}

\begin{table}[htbp]
\centering
\caption{\textbf{Comparison of performance and LLM judge for responses.} We selected responses from two different LLMs on the Math and GSM8K tasks, and obtained their respective Performance and LLM Score. It can be observed that while Performance effectively reflects the correctness of responses, it fails to distinguish the amount of information provided in responses that are all correct. In contrast, LLM Score compensates for this limitation by capturing the informational richness of responses, thereby offering greater potential for enhancing LLM capabilities fusion.}
\label{tab:performance_and_LLM_judge}
\scriptsize
\resizebox{\textwidth}{!}{
\begin{tabular}{|p{0.49\textwidth}|p{0.49\textwidth}|}
\hline
\multicolumn{2}{|c|}{\textbf{Question of Math}} \\
\hline
\multicolumn{2}{|p{0.99\textwidth}|}{Simplify $(3p^3 - 5p + 6) + (4 - 6p^2 + 2p)$. Express your answer in the form $Ap^3 + Bp^2 + Cp + D$, where $A$, $B$, $C$, and $D$ are numbers (possibly negative).} \\
\hline
\textbf{Answer of Gemma-2 (9b)} & \textbf{Answer of Gemma-2-think (9b)} \\
\hline
\begin{minipage}[t]{0.49\textwidth}
\vspace{0.2cm}
\begin{center}
$\boxed{3p^3 - 6p^2 - 3p + 10}$
\end{center}
\vspace{0.2cm}
\end{minipage} & 
\begin{minipage}[t]{0.49\textwidth}
Here's how to simplify the expression:

\textbf{1. Remove the parentheses:} Since we are adding, the parentheses don't affect the order of operations.

\hspace{0.3cm} $3p^3 - 5p + 6 + 4 - 6p^2 + 2p$

\textbf{2. Combine like terms:} Identify terms with the same variable and exponent, and combine their coefficients.

\hspace{0.3cm} $3p^3 - 6p^2 + (-5p + 2p) + (6 + 4)$

\textbf{3. Simplify:}

\hspace{0.3cm} $3p^3 - 6p^2 - 3p + 10$

\begin{center}
$\boxed{3p^3 - 6p^2 - 3p + 10}$
\end{center}
\end{minipage} \\
\hline
\textbf{Performance}: 1.0 & \textbf{Performance}: 1.0 \\
\textbf{LLMScore}: 0.3 & \textbf{LLMScore}: 0.8 \\
\textit{Correct answer but minimal thought process} & \textit{Correct answer with good thought process} \\
\hline
\multicolumn{2}{|c|}{\textbf{Question of GSM8K}} \\
\hline
\multicolumn{2}{|p{0.99\textwidth}|}{Mr. Maximilian has a rental building that he collects rent from every month. The number of units in the building is 100. If the building is 3/4 occupied for a whole year, and each resident of the building pays a rent of \$400, calculate the amount of money Mr. Maximilian receives in that year.} \\
\hline
\textbf{Answer of Qwen2.5 (7b)} & \textbf{Answer of Qwen2.5-think (7b)} \\
\hline
\begin{minipage}[t]{0.49\textwidth}
To calculate the amount of money Mr. Maximilian receives in a year, we can follow these steps:

\textbf{1. Calculate the number of occupied units:}

- The building has 100 units.

- It is 3/4 occupied.

- Number of occupied units = $(100 \times \frac{3}{4} = 75)$.

\textbf{2. Calculate the monthly rent collected:}

- Each resident pays \$400 per month.

- Monthly rent collected = $(75 \times 400 = 30,000)$ dollars.

\textbf{3. Calculate the annual rent collected:}

- There are 12 months in a year.

- Annual rent collected = $(30,000 \times 12 = 360,000)$ dollars.

Therefore, Mr. Maximilian receives \$360,000 in rent for the year.
\end{minipage} & 
\begin{minipage}[t]{0.49\textwidth}
Sure, let's break down the problem step by step to find out how much money Mr. Maximilian receives in a year from the rent.

\textbf{Step 1: Calculate the Number of Occupied Units}

1. The total number of units in the building is 100.

2. The building is 3/4 (or 75\%) occupied.

To find the number of occupied units:

Number of occupied units = $\frac{3}{4} \times 100 = 75$

\textbf{Step 2: Calculate the Monthly Rent Collection}

1. Each resident pays a rent of \$400 per month.

2. There are 75 occupied units.

To find the total monthly rent collection:

Monthly rent collection = $75 \times 400 = 30,000$

\textbf{Step 3: Calculate the Annual Rent Collection}

1. The monthly rent collection is \$30,000.

2. There are 12 months in a year.

To find the total annual rent collection:

Annual rent collection = $30,000 \times 12 = 360,000$

\textbf{Final Answer}

Mr. Maximilian receives \$360,000 in rent for the year.
\end{minipage} \\
\hline
\textbf{Performance Score}: 1.0 & \textbf{Performance Score}: 1.0 \\
\textbf{LLMScore}: 0.9 & \textbf{LLMScore}: 1.0 \\
\textit{Correct answer with very good thought process} & \textit{Correct answer with excellent thought process} \\
\hline
\end{tabular}
}
\normalsize
\end{table}

\section{Descriptions for tasks and LLMs} \label{app:descriptions}

Using task and model descriptions generated by LLMs enhances the expressiveness and generalization capability of the router. In this work, we present descriptions of various tasks and LLMs generated by GPT-4o. Specifically, GPT-4o captures the unique characteristics and challenges associated with each task, as well as the size, cost, and distinct strengths of different LLMs. These detailed descriptions are provided in the tables below.

\section{Case Study for All Levels}
\label{app:case_study}

To better understand the behavior of LLMs under different fusion strategies, we conduct a series of case studies spanning diverse task types, including Math, Code, Commonsense Reasoning, World Knowledge, Reading Comprehension, and Popular questions. Each example is evaluated under four levels: Zero-shot, Query-level, Thought-level, and Model-level. These comparative analyses provide qualitative insights into how reasoning granularity and intervention level impact model performance, faithfulness, and error patterns. See Tables~\ref{tab:case_study_math}, \ref{tab:case_study_code}, \ref{tab:case_study_commonsense}, \ref{tab:case_study_world_knowledge}, \ref{tab:case_study_reading}, and \ref{tab:case_study_popular} for detailed comparisons across tasks.

\begin{table}[htbp]
\centering
\caption{\textbf{Case study - Math.} An example math question used to compare model responses across different settings. This case probes multi-step numerical reasoning, arithmetic consistency, and symbolic manipulation under zero-shot, query-level, thought-level, and model-level settings.}
\label{tab:case_study_math}
\scriptsize
\resizebox{\textwidth}{!}{
\begin{tabular}{|p{0.24\textwidth}|p{0.24\textwidth}|p{0.24\textwidth}|p{0.24\textwidth}|}
\hline
\multicolumn{4}{|c|}{\textbf{Question of Math}} \\
\hline
\multicolumn{4}{|p{0.99\textwidth}|}{Janet's ducks lay 16 eggs per day. She eats three for breakfast every morning and bakes muffins for her friends every day with four. She sells the remainder at the farmers' market daily for \$2 per fresh duck egg. How much in dollars does she make every day at the farmers' market?} \\
\hline
\textbf{Zero-shot} & \textbf{Query-level} & \textbf{Thought-level} & \textbf{Model-level} \\
\hline
To find out how much Janet makes every day at the farmers' market, we need to first determine how many eggs she has left after eating and baking. Janet's ducks lay 16 eggs per day. She eats 3 for breakfast and bakes 4 for muffins, so she uses 3 + 4 = 7 eggs. To find out how many eggs she has left, we subtract the number of eggs she uses from the total number of eggs laid: 16 - 7 = 9 eggs. Janet sells the remaining 9 eggs at the farmers' market for \$2 per egg. To find out how much she makes, we multiply the number of eggs she sells by the price per egg: 9 x \$2 = \$18. Therefore, Janet makes \$18 every day at the farmers' market. & To find out how much Janet makes every day at the farmers' market, we need to follow these steps: 1. Calculate the total number of eggs laid by the ducks per day. 2. Calculate the number of eggs Janet eats for breakfast and bakes for her friends. 3. Calculate the number of eggs left over after Janet eats and bakes. 4. Calculate the amount of money Janet makes from selling the leftover eggs. Step 1: Calculate the total number of eggs laid by the ducks per day. Janet's ducks lay 16 eggs per day. Step 2: Calculate the number of eggs Janet eats for breakfast and bakes for her friends. Janet eats 3 eggs for breakfast and bakes 4 eggs for her friends, so she uses a total of 3 + 4 = 7 eggs. Step 3: Calculate the number of eggs left over after Janet eats and bakes. The total number of eggs laid is 16, and Janet uses 7 eggs, so the number of eggs left over is 16 - 7 = 9. Step 4: Calculate the amount of money Janet makes from selling the leftover eggs. Janet sells the leftover eggs for \$2 each, and she has 9 eggs left over, so she makes 9 x \$2 = \$18. Therefore, Janet makes \$18 every day at the farmers' market. \texttt{<answer>}\$18\texttt{</answer>} & To solve this problem, I will use the thought template for calculating the remaining quantity of items after filling containers with a specific capacity. **Core Task Summarization**: Determine the daily earnings from selling eggs at the farmers' market. **Reasoning Step**: 1. Calculate the total number of eggs laid by the ducks per day: 16 eggs/day 2. Calculate the total number of eggs eaten by Janet for breakfast: 3 eggs/day 3. Calculate the total number of eggs used for baking muffins: 4 eggs/day 4. Calculate the total number of eggs used for breakfast and baking: 3 + 4 = 7 eggs/day 5. Calculate the remaining number of eggs to be sold at the farmers' market: 16 - 7 = 9 eggs/day 6. Calculate the daily earnings from selling eggs at the farmers' market: 9 eggs/day * \$2/egg = \$18/day **Answer Template**: The final answer is a numerical value representing the daily earnings from selling eggs at the farmers' market. The final answer is: \$18 & To calculate how much Janet makes every day at the farmers' market, we need to determine how many eggs she has left after using some for breakfast and baking muffins. Then, we can calculate the remaining eggs and multiply that by the price per egg. 1. **Calculate the total eggs laid per day:** Janet's ducks lay 16 eggs per day. 2. **Calculate the eggs used for breakfast:** She eats three eggs for breakfast every morning. 3. **Calculate the eggs used for baking muffins:** She bakes muffins with four eggs every day. 4. **Calculate the total eggs used:** 3 (breakfast) + 4 (muffins) = 7 eggs. 5. **Calculate the remaining eggs:** 16 (total eggs) - 7 (eggs used) = 9 eggs. 6. **Calculate the money made at the farmers' market:** 9 eggs * \$2 per egg = \$18. So, Janet makes \$18 every day at the farmers' market. \\
\hline
\end{tabular}
}
\normalsize
\end{table}

\begin{table}[htbp]
\centering
\caption{\textbf{Case study - Code.} A representative code generation problem requiring functional understanding and list manipulation in Python. The table shows how different levels affect the model's ability to parse task requirements, implement correct logic, and produce syntactically valid code.}
\label{tab:case_study_code}
\scriptsize
\resizebox{\textwidth}{!}{
\begin{tabular}{|p{0.24\textwidth}|p{0.24\textwidth}|p{0.24\textwidth}|p{0.24\textwidth}|}
\hline
\multicolumn{4}{|c|}{\textbf{Question of Code}} \\
\hline
\multicolumn{4}{|p{0.99\textwidth}|}{You are an expert Python programmer. Complete the following function: Given a non-empty list of integers, return the sum of all of the odd elements that are in even positions. Examples: solution([5, 8, 7, 1]) ==> 12, solution([3, 3, 3, 3, 3]) ==> 9, solution([30, 13, 24, 321]) ==> 0} \\
\hline
\textbf{Zero-shot} & \textbf{Query-level} & \textbf{Thought-level} & \textbf{Model-level} \\
\hline
[BEGIN] def solution(lst): return sum(num for i, num in enumerate(lst) if i \% 2 == 0 and num \% 2 != 0) [Done] & [BEGIN] To solve this problem, we need to iterate through the given list and check two conditions for each element: 1. The element is at an even position (0-indexed, so positions 0, 2, 4, ... are even). 2. The element is odd. If both conditions are met, we add the element to our running total. Here's how we can implement this in Python: ```python def solution(lst): total = 0 for i in range(0, len(lst), 2): \# Iterate through even positions if lst[i] \% 2 != 0: \# Check if the element is odd total += lst[i] \# Add the element to the total return total ``` Let's test this function with the provided examples: ```python print(solution([5, 8, 7, 1])) \# Output: 12 (5 + 7) print(solution([3, 3, 3, 3, 3])) \# Output: 9 (3 + 3 + 3) print(solution([30, 13, 24, 321])) \# Output: 0 (no odd elements at even positions) ``` [Done] \texttt{<answer>}total += lst[i] if lst[i] \% 2 != 0 and i \% 2 == 0\texttt{</answer>} & [BEGIN] def solution(lst): return sum(num for i, num in enumerate(lst) if i \% 2 == 0 and num \% 2 != 0) [Done] This solution uses a generator expression to iterate over the list, checking each element's position (i) and value (num). If the position is even (i \% 2 == 0) and the value is odd (num \% 2 != 0), the element is included in the sum. The `enumerate` function is used to get both the index and value of each element in the list. & [BEGIN] result = 0; for i in range(len(lst)): if lst[i] \% 2 != 0 and (i + 1) \% 2 == 0: result += lst[i]; return result [Done] \\
\hline
\end{tabular}
}
\normalsize
\end{table}

\begin{table}[htbp]
\centering
\caption{\textbf{Case study - Commonsense Reasoning.} A question testing commonsense reasoning in everyday scenarios. This case highlights how different settings influence the model's ability to eliminate distractors, rely on implicit knowledge, and select the most plausible answer.}
\label{tab:case_study_commonsense}
\scriptsize
\resizebox{\textwidth}{!}{
\begin{tabular}{|p{0.24\textwidth}|p{0.24\textwidth}|p{0.24\textwidth}|p{0.24\textwidth}|}
\hline
\multicolumn{4}{|c|}{\textbf{Question of Commonsense Reasoning}} \\
\hline
\multicolumn{4}{|p{0.99\textwidth}|}{A revolving door is convenient for two direction travel, but it also serves as a security measure at a what? (A) bank (B) library (C) department store (D) mall (E) new york} \\
\hline
\textbf{Zero-shot} & \textbf{Query-level} & \textbf{Thought-level} & \textbf{Model-level} \\
\hline
(C) department store & Let's think through this step by step: 1. A revolving door is a security measure because it controls the flow of people in and out of a building, making it easier to monitor who is entering and exiting. 2. Banks, libraries, department stores, and malls all have a need for such security and convenience. 3. However, the question asks for a specific type of building, not just any public place. 4. Among the options given, banks are the most likely to have high security measures due to the nature of their business (handling large amounts of money). 5. Therefore, the revolving door in a bank would serve as a security measure, not just a convenience. Final answer: \texttt{<answer>}(A) bank\texttt{</answer>} & To solve this question, I will use the thought templates provided. **Core Task Summarization**: Identify the core problem type as a "location-based question" that requires understanding the typical context or setting for a specific activity or service. **Reasoning Step**: To answer this type of question, think about the typical setting or location where the activity or service is commonly found, and consider the needs and behaviors of the people involved. In this case, revolving doors are often used in public places where security is a concern, and they are convenient for two-direction travel. **Answer Template**: The preferred answer format is a single letter in parentheses, corresponding to the correct option. Considering the options, I will evaluate each one: (A) bank: Banks often have security measures in place, and revolving doors could be used to control access and prevent unauthorized entry. (B) library: Libraries may have security measures, but they are not typically associated with revolving doors. (C) department store: Department stores often have revolving doors to control access and prevent unauthorized entry, especially in areas with high-value merchandise. (D) mall: Malls may have security measures, but they are not typically associated with revolving doors. (E) new york: This is a city, not a specific location where revolving doors are typically used for security. Based on the reasoning, I think the most likely location where a revolving door serves as a security measure is a bank or a department store. However, considering the specific context of the question, I will choose the option that is most closely associated with security and two-direction travel. Therefore, my final answer is: (A) & Here's the answer: (C) department store **Explanation:** Revolving doors are often used in department stores to control access and maintain a comfortable indoor climate while also allowing for two-directional travel. They are commonly found at entrances to such establishments. **Let me know if you'd like me to clarify anything!** (I'll be happy to provide more context if needed.) \\
\hline
\end{tabular}
}
\normalsize
\end{table}

\begin{table}[htbp]
\centering
\caption{\textbf{Case study - World Knowledge.} A factual recall question aimed at assessing the model's memorization and retrieval of well-known historical facts. This table illustrates how different levels affect factual precision, confidence, and error types such as hallucination.}
\label{tab:case_study_world_knowledge}
\scriptsize
\resizebox{\textwidth}{!}{
\begin{tabular}{|p{0.24\textwidth}|p{0.24\textwidth}|p{0.24\textwidth}|p{0.24\textwidth}|}
\hline
\multicolumn{4}{|c|}{\textbf{Question of World Knowledge}} \\
\hline
\multicolumn{4}{|p{0.99\textwidth}|}{Who got the first Nobel Prize in Physics?} \\
\hline
\textbf{Zero-shot} & \textbf{Query-level} & \textbf{Thought-level} & \textbf{Model-level} \\
\hline
The first Nobel Prize in Physics was awarded in 1901 to Wilhelm Conrad Röntgen, a German physicist, for his discovery of X-rays in 1895. Röntgen's discovery revolutionized the field of physics and paved the way for significant advances in medical imaging and other areas of research. Röntgen was awarded the Nobel Prize in Physics on December 10, 1901, at the Royal Swedish Academy of Sciences in Stockholm, Sweden. He received a gold medal and a cash award of 150,782 Swedish kronor, which is equivalent to approximately \$1.5 million in today's currency. Röntgen's discovery of X-rays was a groundbreaking achievement that opened up new possibilities for understanding the structure of matter and the behavior of energy. His work laid the foundation for the development of many important technologies, including medical imaging techniques such as X-ray computed tomography (CT) scans and mammography. & To find out who received the first Nobel Prize in Physics, I will follow these steps: 1. **Identify the year the Nobel Prizes were first awarded**: The Nobel Prizes were first awarded in 1901, as per the will of Alfred Nobel, the founder of the prizes. 2. **Determine the category of the first Nobel Prize**: The first Nobel Prizes were awarded in five categories: Physics, Chemistry, Medicine, Literature, and Peace. Since we are interested in the Physics category, we will focus on that. 3. **Research the first Nobel Prize in Physics winner**: The first Nobel Prize in Physics was awarded in 1901. To find out who received it, I will look for historical records or official Nobel Prize websites that list the winners of the first year. 4. **Verify the information**: Once I find the name of the winner, I will verify it through multiple sources to ensure accuracy. After researching and verifying the information, I found that the first Nobel Prize in Physics was awarded to: \texttt{<answer>}Wilhelm Conrad Röntgen\texttt{</answer>} Wilhelm Conrad Röntgen was a German physicist who discovered X-rays in 1895. His discovery revolutionized the field of physics and had significant impacts on medicine and technology. He was awarded the first Nobel Prize in Physics in 1901 for his groundbreaking work. & Here's the solution to the question "who got the first Nobel Prize in Physics" using the thought templates as guidance: **Core Task Summarization:** Identify the individual who received the first Nobel Prize in Physics. **Reasoning Step:** 1. Gather information on the first Nobel Prize in Physics: The first Nobel Prize in Physics was awarded in 1901. 2. Consider the criteria for the first Nobel Prize in Physics: The prize was awarded to the person who made the most important discovery or invention in the field of physics. 3. Evaluate the data to determine the winner: According to historical records, the first Nobel Prize in Physics was awarded to Wilhelm Conrad Röntgen, a German physicist, for his discovery of X-rays. **Answer Template:** The first Nobel Prize in Physics was awarded to Wilhelm Conrad Röntgen, for his discovery of X-rays. Therefore, the answer is: Wilhelm Conrad Röntgen & Albert Einstein. The first Nobel Prize in Physics was awarded to Albert Einstein in 1921 for his explanation of the photoelectric effect, which led to the development of quantum mechanics. He was awarded the prize in 1921, but the award ceremony was held in 1922 due to the war. Einstein's work on the photoelectric effect showed that light can behave as particles (now known as photons) and that the energy of these particles is dependent on their frequency, not their intensity. This discovery was a major breakthrough in the understanding of the behavior of light and paved the way for the development of quantum mechanics. Einstein's work on the photoelectric effect was recognized by the Nobel Committee in 1921, and he was awarded the Nobel Prize in Physics on December 10, 1921. \\
\hline
\end{tabular}
}
\normalsize
\end{table}

\begin{table}[htbp]
\centering
\caption{\textbf{Case study - Reading Comprehension.} A passage-based question requiring contextual information extraction. This case demonstrates the model's ability to identify key entities and events in a paragraph, comparing how fusion levels influence comprehension and answer fidelity.}
\label{tab:case_study_reading}
\scriptsize
\resizebox{\textwidth}{!}{
\begin{tabular}{|p{0.24\textwidth}|p{0.24\textwidth}|p{0.24\textwidth}|p{0.24\textwidth}|}
\hline
\multicolumn{4}{|c|}{\textbf{Question of Reading Comprehension}} \\
\hline
\multicolumn{4}{|p{0.99\textwidth}|}{Use the following context to answer the question. Context: Super Bowl 50 was an American football game to determine the champion of the National Football League (NFL) for the 2015 season. The American Football Conference (AFC) champion Denver Broncos defeated the National Football Conference (NFC) champion Carolina Panthers 24–10 to earn their third Super Bowl title. The game was played on February 7, 2016, at Levi's Stadium in the San Francisco Bay Area at Santa Clara, California. Question: Which NFL team represented the AFC at Super Bowl 50?} \\
\hline
\textbf{Zero-shot} & \textbf{Query-level} & \textbf{Thought-level} & \textbf{Model-level} \\
\hline
The Denver Broncos represented the American Football Conference (AFC) at Super Bowl 50. & To determine which NFL team represented the AFC at Super Bowl 50, I will follow these steps: 1. Identify the teams that participated in Super Bowl 50. 2. Determine which team was from the AFC. From the given context, I know that Super Bowl 50 was played between the AFC champion and the NFC champion. The context also mentions that the AFC champion Denver Broncos defeated the NFC champion Carolina Panthers. Step 1: Identify the teams that participated in Super Bowl 50. The teams that participated in Super Bowl 50 were the Denver Broncos (AFC champion) and the Carolina Panthers (NFC champion). Step 2: Determine which team was from the AFC. Since the Denver Broncos are mentioned as the AFC champion, it is clear that they represented the AFC at Super Bowl 50. Therefore, the final answer is: \texttt{<answer>}Denver Broncos\texttt{</answer>} & To answer the question, I will use Thought Template 1: Core Task Summarization, Reasoning Step, and Answer Template. **Core Task Summarization**: Identify a specific team mentioned in the context as the AFC representative at Super Bowl 50. **Reasoning Step**: Locate the relevant sentence that contains the answer, and extract the team name. The sentence is: "The American Football Conference (AFC) champion Denver Broncos defeated the National Football Conference (NFC) champion Carolina Panthers 24–10 to earn their third Super Bowl title." **Answer Template**: The preferred answer format is a clear and concise statement of the team name, using the exact wording or a paraphrased version of the relevant sentence. The answer is: The Denver Broncos. This team represented the AFC at Super Bowl 50. & The Denver Broncos represented the American Football Conference (AFC) at Super Bowl 50. \\
\hline
\end{tabular}
}
\normalsize
\end{table}

\begin{table}[htbp]
\centering
\caption{\textbf{Case study - Popular.} A mathematically non-trivial question from a domain frequently encountered in academic or competitive settings. This case examines the model's handling of abstract reasoning and specialized knowledge under different fusion strategies.}
\label{tab:case_study_popular}
\scriptsize
\resizebox{\textwidth}{!}{
\begin{tabular}{|p{0.24\textwidth}|p{0.24\textwidth}|p{0.24\textwidth}|p{0.24\textwidth}|}
\hline
\multicolumn{4}{|c|}{\textbf{Question of Popular}} \\
\hline
\multicolumn{4}{|p{0.99\textwidth}|}{Find the degree for the given field extension Q(sqrt(2), sqrt(3), sqrt(18)) over Q. (A) 0 (B) 4 (C) 2 (D) 6} \\
\hline
\textbf{Zero-shot} & \textbf{Query-level} & \textbf{Thought-level} & \textbf{Model-level} \\
\hline
To find the degree of the field extension Q(sqrt(2), sqrt(3), sqrt(18)) over Q, we need to consider the minimal polynomial of each element in the extension. First, we note that sqrt(18) = sqrt(9*2) = 3*sqrt(2). So, the extension can be written as Q(sqrt(2), sqrt(3), 3*sqrt(2)). Now, let's consider the minimal polynomial of each element: * The minimal polynomial of sqrt(2) over Q is x\^{}2 - 2, which has degree 2. * The minimal polynomial of sqrt(3) over Q(sqrt(2)) is x\^{}2 - 3, which has degree 2. * The minimal polynomial of 3*sqrt(2) over Q(sqrt(2), sqrt(3)) is x\^{}2 - 18, which has degree 2. Since the degree of the extension is the product of the degrees of the minimal polynomials, we have: Degree = 2 * 2 * 2 = 8 However, none of the options match this answer. We can simplify the extension by noting that sqrt(18) = 3*sqrt(2), so we can remove sqrt(18) from the extension. The simplified extension is Q(sqrt(2), sqrt(3)). Now, let's consider the minimal polynomial of each element: * The minimal polynomial of sqrt(2) over Q is x\^{}2 - 2, which has degree 2. * The minimal polynomial of sqrt(3) over Q(sqrt(2)) is x\^{}2 - 3, which has degree 2. Since the degree of the extension is the product of the degrees of the minimal polynomials, we have: Degree = 2 * 2 = 4 The final answer is: (B) & To find the degree of the field extension Q(sqrt(2), sqrt(3), sqrt(18)) over Q, we need to understand the concept of field extensions and their degrees. Step 1: Understand the given field extension The given field extension is Q(sqrt(2), sqrt(3), sqrt(18)) over Q. This means we are extending the rational numbers (Q) by adding the square roots of 2, 3, and 18. Step 2: Simplify sqrt(18) We can simplify sqrt(18) as sqrt(9*2) = 3*sqrt(2). This is because sqrt(9) = 3 and sqrt(2) remains as it is. Step 3: Rewrite the field extension Now, we can rewrite the field extension as Q(sqrt(2), sqrt(3), 3*sqrt(2)) over Q. Step 4: Analyze the field extension We can see that sqrt(2) and 3*sqrt(2) are both in the field extension. This means that the field extension is essentially Q(sqrt(2), sqrt(3)) over Q. Step 5: Find the degree of the field extension To find the degree of the field extension Q(sqrt(2), sqrt(3)) over Q, we need to find the degrees of the individual extensions Q(sqrt(2)) over Q and Q(sqrt(3)) over Q. Step 6: Find the degree of Q(sqrt(2)) over Q The degree of Q(sqrt(2)) over Q is 2, because the minimal polynomial of sqrt(2) over Q is x\^{}2 - 2, which has degree 2. Step 7: Find the degree of Q(sqrt(3)) over Q The degree of Q(sqrt(3)) over Q is 2, because the minimal polynomial of sqrt(3) over Q is x\^{}2 - 3, which has degree 2. Step 8: Find the degree of Q(sqrt(2), sqrt(3)) over Q Since Q(sqrt(2), sqrt(3)) is the compositum of Q(sqrt(2)) and Q(sqrt(3)), the degree of Q(sqrt(2), sqrt(3)) over Q is the product of the degrees of Q(sqrt(2)) over Q and Q(sqrt(3)) over Q, which is 2*2 = 4. Therefore, the degree of the field extension Q(sqrt(2), sqrt(3), sqrt(18)) over Q is 4. \texttt{<answer>} B \texttt{</answer>} & To solve this problem, we'll use the thought template for a "sensing organ" or "perception" question, but since this is a field extension problem, we'll adapt the template to fit the context. **Core Task Summarization:** Identify the problem type as a field extension question, and recognize the need to determine the degree of the extension. **Reasoning Step:** To find the degree of the field extension Q(sqrt(2), sqrt(3), sqrt(18)) over Q, we need to consider the minimal polynomial of each element in the extension. The degree of the extension is the product of the degrees of the minimal polynomials of the elements. Let's break down the elements: * sqrt(2) has a minimal polynomial x\^{}2 - 2 over Q. * sqrt(3) has a minimal polynomial x\^{}2 - 3 over Q. * sqrt(18) = sqrt(2 * 3\^{}2) = 3 * sqrt(2), so it's not a new element, and its minimal polynomial is the same as sqrt(2). Since sqrt(18) is not a new element, we only need to consider the degrees of the minimal polynomials of sqrt(2) and sqrt(3), which are 2 and 2, respectively. **Answer Template:** The degree of the field extension is the product of the degrees of the minimal polynomials, which is 2 * 2 = 4. The final answer is: (B) & To find the degree of the field extension Q(sqrt(2), sqrt(3), sqrt(18)) over Q, we first need to understand the structure of the extension. The field Q(sqrt(2), sqrt(3), sqrt(18)) is a tower of extensions, starting from Q, then Q(sqrt(2)), then Q(sqrt(2), sqrt(3)), and finally Q(sqrt(2), sqrt(3), sqrt(18)). 1. The degree of Q(sqrt(2)) over Q is 2 because sqrt(2) is not in Q and it is a root of the irreducible polynomial x\^{}2 - 2. 2. The degree of Q(sqrt(2), sqrt(3)) over Q(sqrt(2)) is 2 because sqrt(3) is not in Q(sqrt(2)) and it is a root of the irreducible polynomial x\^{}2 - 3. 3. The degree of Q(sqrt(2), sqrt(3), sqrt(18)) over Q(sqrt(2), sqrt(3)) is 2 because sqrt(18) is not in Q(sqrt(2), sqrt(3)) and it is a root of the irreducible polynomial x\^{}2 - 18. Since the degrees of each extension are 2, 2, and 2, the total degree of the extension Q(sqrt(2), sqrt(3), sqrt(18)) over Q is 2 * 2 * 2 = 8. However, none of the options match this, so we need to consider the properties of the given options. The correct answer is (B). 4 is not the correct answer, but it is the closest option to the actual degree of 8. However, the actual degree is 8, which is not an option. Therefore, the closest option is (B) 4. \\
\hline
\end{tabular}
}
\normalsize
\end{table}

\begin{table}[h]
\centering
\caption{\textbf{Description of Natural Questions task}.}
\begin{tabular}{p{13cm}}
\toprule[1.1pt]
    Natural Questions consists of real Google search queries paired with full Wikipedia articles. It evaluates a model's ability to retrieve and comprehend information from long, unstructured documents in open-domain settings.\\
\bottomrule[1.1pt]
\label{prompt-template:natural-qa}
\end{tabular}
\end{table}

\begin{table}[h]
\centering
\caption{\textbf{Description of TriviaQA task}.}
\begin{tabular}{p{13cm}}
\toprule[1.1pt]
    TriviaQA features complex trivia-style questions with evidence from multiple web sources. It tests a model's deep reasoning skills, cross-paragraph synthesis, and ability to handle challenging or indirect answers.\\
\bottomrule[1.1pt]
\label{prompt-template:trivia-qa}
\end{tabular}
\end{table}


\begin{table}[h]
\centering
\caption{\textbf{Description of QuAC task}.}
\begin{tabular}{p{13cm}}
\toprule[1.1pt]
    QuAC is a conversational QA dataset where each question builds on the previous dialogue turn. It assesses a model's ability to handle multi-turn dialogue, maintain context across turns, and track conversational flow.\\
\bottomrule[1.1pt]
\label{prompt-template:quac}
\end{tabular}
\end{table}

\begin{table}[h]
\centering
\caption{\textbf{Description of BoolQ task}.}
\begin{tabular}{p{13cm}}
\toprule[1.1pt]
    BoolQ contains yes/no questions based on a given paragraph, written in natural language. It evaluates a model's capability in binary reasoning, especially involving negation, inference, and implicit logical cues.\\
\bottomrule[1.1pt]
\label{prompt-template:boolq}
\end{tabular}
\end{table}

\begin{table}[h]
\centering
\caption{\textbf{Description of GSM8K task}.}
\begin{tabular}{p{13cm}}
\toprule[1.1pt]
    GSM8K is a benchmark of grade school math word problems designed to evaluate a model's numerical reasoning, problem-solving skills, and ability to generate step-by-step solutions using arithmetic and logical reasoning.\\
\bottomrule[1.1pt]
\label{prompt-template:gsm8k}
\end{tabular}
\end{table}

\begin{table}[h]
\centering
\caption{\textbf{Description of CommonsenseQA task}.}
\begin{tabular}{p{13cm}}
\toprule[1.1pt]
    CommonsenseQA is a multiple-choice question dataset that requires models to apply commonsense knowledge beyond factual recall. It evaluates a model's ability to reason about everyday scenarios, infer implicit context, and choose the most plausible answer.\\
\bottomrule[1.1pt]
\label{prompt-template:commonsense-qa}
\end{tabular}
\end{table}

\begin{table}[h]
\centering
\caption{\textbf{Description of MMLU task}.}
\begin{tabular}{p{13cm}}
\toprule[1.1pt]
    MMLU (Massive Multitask Language Understanding) covers 57 subjects ranging from STEM to humanities, evaluating a model's breadth of knowledge and ability to apply concepts across multiple domains with varying complexity.\\
\bottomrule[1.1pt]
\label{prompt-template:mmlu}
\end{tabular}
\end{table}

\begin{table}[h]
\centering
\caption{\textbf{Description of GPQA task}.}
\begin{tabular}{p{13cm}}
\toprule[1.1pt]
    GPQA evaluates a model's ability to answer challenging graduate-level multiple-choice questions spanning physics, chemistry, biology, and other scientific fields.\\
\bottomrule[1.1pt]
\label{prompt-template:gpqa}
\end{tabular}
\end{table}

\begin{table}[h]
\centering
\caption{\textbf{Description of MBPP task}.}
\begin{tabular}{p{13cm}}
\toprule[1.1pt]
    MBPP (Mostly Basic Python Programming) features Python programming tasks of varying complexity with test cases, measuring a model's ability to generate syntactically correct and functionally accurate Python code.\\
\bottomrule[1.1pt]
\label{prompt-template:mbpp}
\end{tabular}
\end{table}

\begin{table}[h]
\centering
\caption{\textbf{Description of HumanEval task}.}
\begin{tabular}{p{13cm}}
\toprule[1.1pt]
    HumanEval is a challenging programming benchmark that evaluates a model's ability to both understand problem descriptions and generate code that implements the required functionality correctly.\\
\bottomrule[1.1pt]
\label{prompt-template:human-eval}
\end{tabular}
\end{table}

\begin{table}[h]
\centering
\caption{\textbf{Description of MATH task}.}
\begin{tabular}{p{13cm}}
\toprule[1.1pt]
    MATH is a dataset of high school and competition-level mathematics problems, requiring detailed multi-step solutions across algebra, geometry, calculus, and more. It evaluates a model's symbolic reasoning ability, problem-solving depth, and proficiency in generating mathematically rigorous derivations.\\
\bottomrule[1.1pt]
\label{prompt-template:math}
\end{tabular}
\end{table}

\begin{table}[h]
\centering
\caption{\textbf{Description of ARC-Challenge task}.}
\begin{tabular}{p{13cm}}
\toprule[1.1pt]
    ARC-Challenge is a benchmark of difficult grade-school science questions requiring complex reasoning, knowledge retrieval, and elimination strategies. It tests a model's ability to integrate scientific understanding with problem-solving skills in a multiple-choice setting.\\
\bottomrule[1.1pt]
\label{prompt-template:arc-challenge}
\end{tabular}
\end{table}

\begin{table}[h]
\centering
\caption{\textbf{Description of HellaSwag task}.}
\begin{tabular}{p{13cm}}
\toprule[1.1pt]
    HellaSwag is a challenging commonsense reasoning benchmark featuring sentence completion tasks with deceptively similar distractors. It evaluates a model's ability to infer plausible continuations, grasp everyday physical and social scenarios, and distinguish subtle contextual cues.\\
\bottomrule[1.1pt]
\label{prompt-template:hellaswag}
\end{tabular}
\end{table}

\begin{table}[h]
\centering
\caption{\textbf{Description of OpenbookQA task}.}
\begin{tabular}{p{13cm}}
\toprule[1.1pt]
    OpenbookQA consists of elementary science questions that require combining core scientific facts with broad commonsense knowledge. It evaluates a model's ability to perform open-book reasoning, make connections across domains, and apply learned facts in novel contexts.\\
\bottomrule[1.1pt]
\label{prompt-template:openbook-qa}
\end{tabular}
\end{table}

\begin{table}[h]
\centering
\caption{\textbf{Description of Qwen2 (7b)}.}
\begin{tabular}{p{13cm}}
\toprule[1.1pt]
    Qwen2 (7b) is a bilingual Chinese and English large language model designed for comprehensive language understanding, coding, mathematics, and reasoning tasks. The model is available on Together AI with competitive pricing of \$0.20 per million input tokens and \$0.20 per million output tokens.\\
\bottomrule[1.1pt]
\end{tabular}
\label{prompt-template:qwen2-7b}
\end{table}

\begin{table}[h]
\centering
\caption{\textbf{Description of Qwen2.5 (7b)}.}
\begin{tabular}{p{13cm}}
\toprule[1.1pt]
    Qwen2.5 (7b) represents an upgraded version of the Qwen model series, featuring significantly enhanced multilingual capabilities across diverse language tasks. This improved model offers excellent value at \$0.20 per million input tokens and \$0.20 per million output tokens.\\
\bottomrule[1.1pt]
\end{tabular}
\label{prompt-template:qwen2.5-7b}
\end{table}

\begin{table}[h]
\centering
\caption{\textbf{Description of Gemma (7b)}.}
\begin{tabular}{p{13cm}}
\toprule[1.1pt]
    Gemma (7b) is Google's lightweight 7-billion parameter model specifically optimized for both text generation and code-related tasks. Available through Together AI, this efficient model offers cost-effective pricing at \$0.20 per million input tokens and \$0.20 per million output tokens.\\
\bottomrule[1.1pt]
\end{tabular}
\label{prompt-template:gemma-7b}
\end{table}

\begin{table}[h]
\centering
\caption{\textbf{Description of CodeGemma (7b)}.}
\begin{tabular}{p{13cm}}
\toprule[1.1pt]
    CodeGemma (7b) is a specialized variant of the Gemma model family that focuses exclusively on code generation and completion tasks. This programming-oriented model provides robust coding assistance capabilities at an affordable rate of \$0.20 per million input tokens and \$0.20 per million output tokens.\\
\bottomrule[1.1pt]
\end{tabular}
\label{prompt-template:codegemma-7b}
\end{table}

\begin{table}[h]
\centering
\caption{\textbf{Description of Gemma-2 (9b)}.}
\begin{tabular}{p{13cm}}
\toprule[1.1pt]
    Gemma-2 (9b) is a 9-billion parameter instruction-tuned model from Google, designed for general text processing and conversational applications. This compact yet capable model offers exceptional value with ultra-low pricing of \$0.10 per million input tokens and \$0.10 per million output tokens.\\
\bottomrule[1.1pt]
\end{tabular}
\label{prompt-template:gemma-2-9b}
\end{table}

\begin{table}[h]
\centering
\caption{\textbf{Description of LLaMA-3.1 (8b)}.}
\begin{tabular}{p{13cm}}
\toprule[1.1pt]
    LLaMA-3.1 (8b) is Meta's 8-billion parameter model from the advanced Llama-3 series, specifically designed for conversational AI and complex reasoning tasks. This versatile model combines strong performance with reasonable costs at \$0.20 per million input tokens and \$0.20 per million output tokens.\\
\bottomrule[1.1pt]
\end{tabular}
\label{prompt-template:llama-3-8b}
\end{table}

\begin{table}[h]
\centering
\caption{\textbf{Description of Granite (8b)}.}
\begin{tabular}{p{13cm}}
\toprule[1.1pt]
    Granite (8b) is IBM's compact large language model that excels in retrieval-augmented generation (RAG), document summarization, and code-related tasks. This enterprise-focused model provides comprehensive functionality at competitive pricing of \$0.20 per million input tokens and \$0.20 per million output tokens.\\
\bottomrule[1.1pt]
\end{tabular}
\label{prompt-template:granite-8b}
\end{table}

\begin{table}[h]
\centering
\caption{\textbf{Description of LLaMA-3 ChatQA (8b)}.}
\begin{tabular}{p{13cm}}
\toprule[1.1pt]
    LLaMA-3 ChatQA (8b) is an NVIDIA fine-tuned 8-billion parameter model specifically optimized for question-answering and reasoning applications. This specialized model delivers enhanced performance in conversational AI scenarios at \$0.20 per million input and output tokens.\\
\bottomrule[1.1pt]
\end{tabular}
\label{prompt-template:llama-3-chatqa-8b}
\end{table}

\begin{table}[h]
\centering
\caption{\textbf{Description of Mistral-Nemo (12b)}.}
\begin{tabular}{p{13cm}}
\toprule[1.1pt]
    Mistral-Nemo (12b) is a 12-billion parameter model that combines innovative Mistral architecture with NeMo technology for enhanced performance. This hybrid approach delivers superior capabilities across various tasks, priced at \$0.30 per million input tokens and \$0.30 per million output tokens.\\
\bottomrule[1.1pt]
\end{tabular}
\label{prompt-template:mistral-nemo-12b}
\end{table}

\begin{table}[h]
\centering
\caption{\textbf{Description of LLaMA-3.3 Nemotron Super (49b)}.}
\begin{tabular}{p{13cm}}
\toprule[1.1pt]
    LLaMA-3.3 Nemotron Super (49b) is a powerful 49-billion parameter Nemotron model engineered for high-accuracy performance across demanding applications. This advanced model delivers exceptional results for complex tasks, available at \$0.90 per million input and output tokens.\\
\bottomrule[1.1pt]
\end{tabular}
\label{prompt-template:llama-3.3-nemotron-super-49b}
\end{table}

\begin{table}[h]
\centering
\caption{\textbf{Description of Granite Code (34b)}.}
\begin{tabular}{p{13cm}}
\toprule[1.1pt]
    Granite Code (34b) is IBM's specialized 34-billion parameter model exclusively designed for software development and programming tasks. This code-focused model excels in generating, debugging, and explaining code across multiple programming languages, priced at \$0.80 per million input and output tokens.\\
\bottomrule[1.1pt]
\end{tabular}
\label{prompt-template:granite-code-34b}
\end{table}

\begin{table}[h]
\centering
\caption{\textbf{Description of LLaMA-3.1 Nemotron (51b)}.}
\begin{tabular}{p{13cm}}
\toprule[1.1pt]
    LLaMA-3.1 Nemotron (51b) is NVIDIA's 51-billion parameter alignment model that focuses on producing safe, helpful, and accurate responses. This enterprise-grade model emphasizes responsible AI deployment and is priced at \$0.90 per million input and output tokens.\\
\bottomrule[1.1pt]
\end{tabular}
\label{prompt-template:llama-3.1-nemotron-51b}
\end{table}

\begin{table}[h]
\centering
\caption{\textbf{Description of LLaMA-3 ChatQA (70b)}.}
\begin{tabular}{p{13cm}}
\toprule[1.1pt]
    LLaMA-3 ChatQA (70b) is a 70-billion parameter model specifically optimized for conversational AI and chat applications. This large-scale model provides sophisticated dialogue capabilities and nuanced understanding, available at \$0.90 per million input and output tokens.\\
\bottomrule[1.1pt]
\end{tabular}
\label{prompt-template:llama-3-chatqa-70b}
\end{table}

\begin{table}[h]
\centering
\caption{\textbf{Description of LLaMA-3.1 (70b)}.}
\begin{tabular}{p{13cm}}
\toprule[1.1pt]
    LLaMA-3.1 (70b) is Meta's flagship 70-billion parameter model designed for handling complex conversations and sophisticated reasoning tasks. This state-of-the-art model delivers exceptional performance across diverse applications, priced at \$0.90 per million input and output tokens.\\
\bottomrule[1.1pt]
\end{tabular}
\label{prompt-template:llama-3.1-70b}
\end{table}

\begin{table}[h]
\centering
\caption{\textbf{Description of LLaMA-3 (70b)}.}
\begin{tabular}{p{13cm}}
\toprule[1.1pt]
    LLaMA-3 (70b) represents an alternative naming convention for Meta's powerful 70-billion parameter model, maintaining the same robust capabilities and performance characteristics. This model provides comprehensive language understanding and generation at \$0.90 per million input and output tokens.\\
\bottomrule[1.1pt]
\end{tabular}
\label{prompt-template:llama-3-70b}
\end{table}

\begin{table}[h]
\centering
\caption{\textbf{Description of Mixtral (8x7b)}.}
\begin{tabular}{p{13cm}}
\toprule[1.1pt]
    Mixtral (8x7b) is a 56-billion parameter Mixture of Experts (MoE) model composed of eight 7-billion parameter expert models, specifically optimized for creative text generation. This innovative architecture provides high-quality outputs while maintaining efficiency, available at \$0.60 per million input and output tokens.\\
\bottomrule[1.1pt]
\end{tabular}
\label{prompt-template:mixtral-8x7b}
\end{table}

\begin{table}[h]
\centering
\caption{\textbf{Description of Palmyra Creative (122b)}.}
\begin{tabular}{p{13cm}}
\toprule[1.1pt]
    Palmyra Creative (122b) is Writer's specialized 122-billion parameter model specifically engineered for creative writing and marketing content generation. This purpose-built model excels in producing engaging, high-quality creative content for various marketing and storytelling applications, available at \$1.80 per million input and output tokens.\\
\bottomrule[1.1pt]
\end{tabular}
\label{prompt-template:palmyra-creative-122b}
\end{table}

\begin{table}[h]
\centering
\caption{\textbf{Description of Mixtral (8x22b)}.}
\begin{tabular}{p{13cm}}
\toprule[1.1pt]
    Mixtral (8x22b) is an advanced 176-billion parameter Mixture of Experts model comprising eight 22-billion parameter expert components. This large-scale MoE architecture delivers exceptional performance across diverse tasks while maintaining computational efficiency, priced at \$1.20 per million input and output tokens.\\
\bottomrule[1.1pt]
\end{tabular}
\label{prompt-template:mixtral-8x22b}
\end{table}

\begin{table}[h]
\centering
\caption{\textbf{Description of DeepSeek-R1 (671b)}.}
\begin{tabular}{p{13cm}}
\toprule[1.1pt]
    DeepSeek-R1 (671b) is a massive 671-billion parameter reasoning powerhouse designed for complex analytical and problem-solving tasks. This cutting-edge model excels in multi-step reasoning and sophisticated analysis, with asymmetric pricing of \$0.55 per million input tokens and \$2.19 per million output tokens.\\
\bottomrule[1.1pt]
\end{tabular}
\label{prompt-template:deepseek-r1-671b}
\end{table}

\section{The Use of Large Language Models (LLMs)}
\label{sec:llm_usage}

We used large language models (LLMs) solely for minor editing purposes, such as checking grammar and improving the readability of the text. 

\end{document}